\PassOptionsToPackage{hyphens}{url}
\documentclass[11pt]{article}
\usepackage[T1]{fontenc}
\usepackage[utf8]{inputenc}
\usepackage{lmodern}
\usepackage[margin=1in]{geometry}
\usepackage{amsmath,amssymb}
\usepackage{booktabs,array,tabularx}
\usepackage{graphicx}
\usepackage{xcolor}
\usepackage{enumitem}
\usepackage{placeins}
\usepackage{needspace}
\usepackage{float}
\usepackage[font=small,labelfont=bf,labelsep=period,skip=4pt]{caption}
\usepackage{microtype}
\usepackage{url}
\usepackage[colorlinks=true,linkcolor={blue!50!black},urlcolor={blue!50!black},citecolor={blue!50!black}]{hyperref}
\hypersetup{pdftitle={RidgeFill: Training-Free Late-Layer Cache Fill for Faster Prefill},pdfauthor={Dylan Johnston and Robert P Smith}}
\title{RidgeFill: Training-Free Late-Layer Cache Fill for Faster Prefill}
\author{Dylan Johnston\\\normalsize Independent researcher \and Robert P Smith\\\normalsize Independent researcher}
\date{Preprint, 2026-11-02}
\begin{document}
\maketitle
\begin{abstract}
RidgeFill shortens prefill by not computing most of a model's upper layers for the prompt. The lower layers run exactly up to a split; from the split layer's state a per-layer map predicts every later layer's input, and each late layer writes its cache from that prediction. The design follows YOCO and DeepSeek-V4.1-Flash, where the projection is trained into the model. RidgeFill retrofits it onto models that were never trained for it: the map is a ridge regression fitted once in closed form from one pass over calibration text, the layer's own projections run unchanged, and a chosen set of rows, the last few thousand tokens, and every decoding step run the full model. Nothing is trained and no token is dropped.

Most of the work was in measurement, finding a repeatable methodology. Loss averaged over all tokens hides RidgeFill's error. Needle retrieval in the filled region, loss on tokens that depend on distant context, and splice tests that swap filled keys or values for exact ones show which component carries it: the keys on Llama-3.1-8B, half of it in their scale, and the values and the recurrent state on Qwen3.5-9B.

On Qwen3.8-27B in a llama.cpp fork on an A100, time to first token is 1.25 to 1.31$\times$ faster at 16K to 120K tokens with the maps in VRAM; RULER changes by $-$1.2 and $-$1.4 points and LongBench v2 by $-$1.4; the 95~\% intervals include zero but do not rule out losses of a few points. On Qwen3.8-Flash-Next in vLLM it is 1.50 to 1.60$\times$ faster than exact at 9K to 32K tokens, with RULER's needle tasks at 99.8 against 100.0 and perplexity on novel tokens 2.65~\% higher [$-$0.02~\%, +4.93~\%]. On three 8 to 9B reference models, at the same split and tail as the identity fill, the fitted map retrieves 49.1 and 49.6 more filled-region RULER points on the two dense models (at that split Llama's loss is worse), and its long-document loss is within 0.03 nats of the released SwiftKV checkpoint's.

\end{abstract}

\section{Introduction}\label{sec:1}
For a long prompt, prefill is the wait before a response, and the user sees nothing until it completes. Before the first output token, a decoder runs every prompt token through every layer. At 120,000 tokens on Qwen3.8-27B that is 128 seconds on an A100. Decoding the answer afterwards is comparatively fast.

\begin{figure}[!htbp]
\centering
\includegraphics[width=\linewidth]{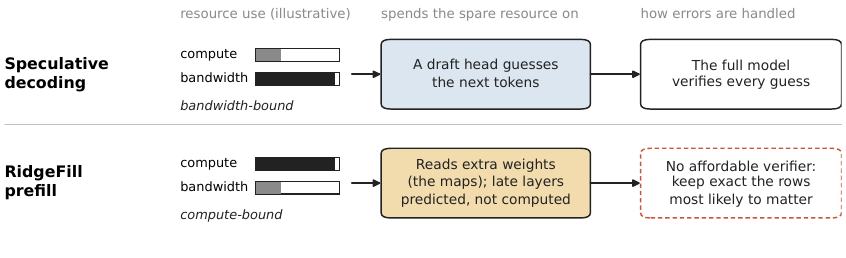}
\caption{Speculative decoding and RidgeFill each spend the resource their phase has spare. Decoding is bandwidth-bound, so a draft head spends the spare compute guessing tokens and the full model verifies every guess. Prefill reuses each weight read across hundreds of prompt rows, so it is limited by arithmetic. RidgeFill spends the spare bandwidth on extra reads, the fill maps from VRAM or host memory, and in return skips most of the late layers' arithmetic. A filled cache has no affordable verifier, so the rows most likely to matter are kept exact instead (\S\ref{sec:2.4}).}\label{fig:1}
\end{figure}

RidgeFill shortens the wait by not fully computing most of the upper layers. It runs the lower layers exactly up to a split layer, whose depth is one of two settings a deployment chooses. From the split layer's hidden state, RidgeFill predicts the input to every later layer. Each late layer then computes only its key and value projections from that prediction, or in a Gated DeltaNet layer its recurrent state update, and skips the rest of the layer, MLP included, for most prompt tokens. Seen this way, the method is speculative decoding (Leviathan et al., 2023; Chen et al., 2023) run backwards (Figure~\ref{fig:1}). Decoding is bandwidth-bound and has compute to spare, so a draft model or head (DeepSeek-AI, 2024) spends that compute guessing tokens the model then verifies. Prefill is the opposite. Each weight read is reused across hundreds of prompt rows, so the GPU is limited by arithmetic, not by memory bandwidth. RidgeFill spends the spare bandwidth on extra reads: one map for each late layer, held in VRAM or in host memory. Each map predicts its layer's input. For most prompt rows, the layer writes its cache from that prediction and skips the rest of its computation.

\begin{figure}[!htbp]
\centering
\includegraphics[width=\linewidth]{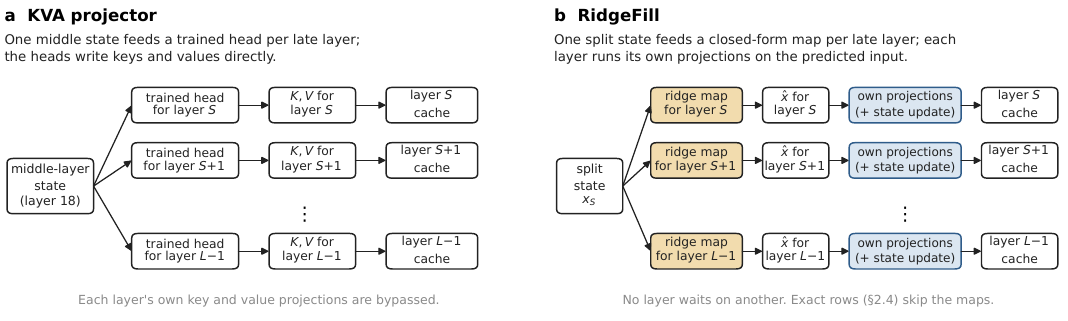}
\caption{Two ways to fill a late layer's cache from a middle layer's state. (a) kishida's KVA projector: trained heads on the layer-18 state write each late layer's keys and values directly, bypassing the layer's own key and value projections. (b) RidgeFill: a closed-form ridge map predicts each late layer's input; the layer's own projections run unchanged, and on a Gated DeltaNet layer so does its state update. Exact rows (\S\ref{sec:2.4}) skip the map and run the layer from its exact input.}\label{fig:2}
\end{figure}

The idea comes from DeepSeek-V4.1-Flash (DeepSeek-AI, 2026), which builds on YOCO (Sun et al., 2024). In that model the upper layers take their caches from a projection of a middle layer's state, and the whole model is trained to absorb the error. The day it came out, Hugging Face user ``kishida'' retrofitted the design onto Qwen3-8B as the KVA projector (kishida, 2026): trained per-layer heads that write the late layers' keys and values directly from the layer-18 state. RidgeFill differs in three ways (Figure~\ref{fig:2}). First, it predicts each late layer's input rather than its keys and values, so the layer's own projections run unchanged, and on a hybrid model so does its recurrent state update. Second, the map is a ridge regression, fitted once in closed form from one pass over calibration text. The process requires no training steps. Third, it does not approximate every token the same way. Some rows, meaning a token's entries in the late layers' caches, are computed exactly, and choosing which ones turned out to be most of the challenge.

Initial efforts focused on finding the method's ceiling. An approximation has random error in it, and the problem resembled digital-to-analog conversion: a lossy stage whose output has to land inside the next stage's acceptance band. So we reached for the tools that solve that problem in other domains, where they work. Jensen's inequality for the softmax, Gaussian closures for the keys, Nyquist's argument for what could be interpolated between exact samples. None of them beat what we already had. The Jensen and Gaussian corrections left the attention error they target unchanged or made it worse, by up to 80~\%. The best of the other corrections lowered that error by 8.6~\% and left the loss where it was. The fill's error is close to white noise from one row to the next, so there is nothing between exact samples to interpolate (Appendix~\ref{sec:I.1}). The analogy fails on its premise. A converter normalizes its input and knows its target. A prompt is not known until it is being processed, it is different every time, and the only definition of the correct output is the full model run on it. Those tools also treat the error as independent of what will be asked. It isn't.

The underlying structure that we spent most of our time in pursuit of showed up initially in needle tests. The first results were not inspiring: on needles inside the filled region, the model returned the right number, save for the last digit, which was wrong (Appendix~\ref{sec:I.2}). The fallout from this illuminated our primary concepts of how to improve accuracy. We had been reading the needle test as a score, but it is also an error instrument.

It says where the fill fails, and it failed on strings the model has to copy exactly: numbers, identifiers, names. Rather than trying to project those, we stopped filling them and ran the rows exactly. Running them exactly resulted in the digit needles recovering. Exact digit rows did little or nothing for needles that were not numbers (Appendix~\ref{sec:I.2}). Using this methodology, we applied similar concepts elsewhere and generalized the rules; the selectors that shipped choose rows by attention or by token class, not by digit (\S\ref{sec:2.4}).

That is also precisely where we discovered our initial means of measurement to be wrong. We had been measuring how far the fill erred on each token. That error is nearly flat across kinds of tokens, which is reassuring and tells you nothing. What is not flat is what an error costs in relation to specific tokens. The damage concentrates in a small share of the prompt's tokens (Appendix~\ref{sec:I.3}), and which tokens those are depends on what will be asked. The useful measurements are which part of the cache carries the error and which way it leans. Splice tests answer the first: swap the filled keys for exact keys and see how much loss goes away, then the values, then both. On Llama-3.1-8B the error is in the keys, half of it in their scale. On Qwen3.5-9B it is in the values and the recurrent state. On Qwen3-8B keys and values have to be fixed together. The corrections in the method answer the second. The value rescale and, in Flash-Next's quality mode, the terminal-state correction each target an average error of the kind the splices exposed, pushing the filled cache back toward the mean; neither was ablated on its own in the final runs. The exact rows remove the error outright where it costs the most.

Unlike speculative decoding, RidgeFill cannot verify (Figure~\ref{fig:1}). The only judge of a filled cache is the full model, and running the full model on every row is exact prefill at exact prefill's cost. We have tried this. It works. Cheaper checks only say where to look: one that scores every filled row at each decoding step added 8 to 18~\% to decoding time, and one that checked each row every few steps missed rows that needed repair in 15 to 98~\% of steps (Appendix~\ref{sec:I.4}). So the goal inverted. Not to make the fill right as often as possible, which ends in training it, but to find what it gets wrong the most and compute those rows exactly.

The rest followed from asking what else predicts where the damage lands. Attention does. Put the question at the end of a long prompt and the lower layers' attention from those last tokens falls on the rows that answer it. Running those rows exactly restores retrieval on two of the three reference models (\S\ref{sec:6.5}). Hiding the question from the selector costs 16 and 14 points, more than half of what choosing the rows adds over random rows of the same count. An exact tail of the most recent tokens does the same for recent context; its length is the second setting a deployment chooses. On Qwen3.8-Flash-Next, an optional quality mode also keeps exact the rarest tokens of the classes that carried the most damage in development. Each of these is a rule a server can apply with no training. The attention rule rests on a property of softmax attention: a token's influence on the output is bounded in proportion to the attention it receives (\S\ref{sec:2.4}), so the rows that receive the most attention from the queries that will be asked are the rows to keep.

One tradeoff does not move. In an architecture that prefills every token through every layer, a method that skips part of that work cannot be both lossless on every prompt and faster than exact (Table~\ref{tab:H.1}). One that manages both is a better kernel, and that is someone else's paper. Exact rows buy quality at exact cost. Filled rows buy speed at some loss. The split depth and the tail length pick the point on that curve.

This paper measures the curve rather than claiming a point on it. For orientation, at split 16 of 64 on Qwen3.8-27B, with the maps in VRAM, prefill is 1.25 to 1.31$\times$ faster at 16K to 120K tokens. The cost is 1.2 and 1.4 points on RULER at 16K and 32K and 1.4 points on LongBench v2. The intervals include zero but do not rule out losses of a few points. We measured on official protocols where they exist, and where they do not, on the community's own instruments: perplexity and KL against the unmodified model, which on the 27B we call stock. Paired bootstrap intervals on the quality measures, except where a table notes otherwise. Baselines at matched depth and matched compute on the reference models. On the 27B, a byte-identical check that the serving engine with the fill switched off equals the unmodified upstream build, so that every difference reported on it is the fill's. The results that cut against the method stay in: Llama-3.1-8B gets worse at shallow splits, the rows that help two models hurt the third, and common-word extraction drops 10 to 14 points on the 27B.

\needspace{6\baselineskip}\noindent \textbf{Contributions.}\par\nopagebreak

\begin{enumerate}[leftmargin=*,itemsep=2pt,topsep=4pt]
\item RidgeFill fills the late layers' caches without training. One capture pass over calibration text gives one ridge map per late layer, and the model's weights are unchanged. An exact tail and the row selectors of \S\ref{sec:2.4} decide which rows bypass the fill.
\item It runs in two serving engines. On Qwen3.8-27B, a 64-layer hybrid of 48 Gated DeltaNet and 16 attention layers, our llama.cpp fork prefills 1.19 to 1.31$\times$ faster than stock on one A100 at 16K to 120K tokens, depending on whether the maps are held in VRAM or streamed from host memory. On RULER and LongBench v2 the point estimates are 1.2 to 1.4 points below stock; the intervals include zero but do not rule out losses of a few points. On Qwen3.8-Flash-Next, a 48-layer hybrid mixture of experts, our vLLM fork on two R9700 GPUs prefills 1.50 to 1.60$\times$ faster than exact prefill at 9K to 32K tokens and scores 99.8 against 100.0 on RULER's needle tasks.
\item A reference study on three 8 to 9B models isolates the fill's effect. It uses needle items whose needle falls inside the filled region, tokens that depend on distant context, question-blind and random-row controls, and splices that locate the error.
\item We compare against POP and SwiftKV as published and at matched split and tail, including needle retrieval for POP's published configuration, which its paper does not report.
\end{enumerate}

\textbf{Other ways to skip late layers.} Prior methods differ in what happens to a skipped token's cache and in whether anything is trained (Table~\ref{tab:1}). POP (He et al., 2026) fills the cache with the identity and trains nothing. SwiftKV (Qiao et al., 2025) fills it with distilled projections, and KV Prediction (Horton et al., 2024) with a trained auxiliary model. TokenGate (Patel et al., 2026) drops tokens above a learned gate and fine-tunes the backbone, and SPEED (Oh et al., 2026) drops prompt tokens from the upper layers and fine-tunes for it, while FastKV (Jo et al., 2026), GemFilter (Shi et al., 2024) and LazyLLM (Fu et al., 2024) drop tokens by attention-based rules. RidgeFill fills the cache with a closed-form map and trains nothing.

At the same split and tail, the fitted map scores 39.7 and 49.6 points higher than the identity fill on filled-region needle retrieval for Llama-3.1-8B (8K to 64K) and Qwen3-8B (8K to 32K). On the hybrid Qwen3.5-9B the gap is 3.3 points, with an interval through zero. RidgeFill stays non-inferior to POP on retrieval, within a 3-point margin, even when it skips more layers than POP's published configuration, though at more multiply-accumulates than POP below about 46K tokens on Llama and 59K on Qwen3-8B (\S\ref{sec:6.4}).

SwiftKV is the trained version of the same idea. Its authors released it for three Llama models; we test the 8B, Llama-3.1-8B, whose distillation run took about 680 million tokens. As released, at its own split, it ties RidgeFill: +0.010 nats, with an interval through zero. Placed inside RidgeFill's pipeline with the same exact tail, SwiftKV's trained projections beat the ridge map by 0.024 nats. That is what the training buys on that model, with our tail.

We do not compare against TokenGate at a matched budget. It trains and drops tokens, which puts it in a different cell of Table~\ref{tab:1}, and it reports larger speedups for that cost.

Of the fill methods in Table~\ref{tab:1}, RidgeFill is the only one that improves on the identity fill without training: on filled-region retrieval, by a wide margin on both dense models, although at Llama's primary split its loss is worse.

\begin{table}[!htbp]
\centering
\caption{Methods that skip late layers for prompt tokens, grouped by what they do with a skipped token's cache and whether they train. The reference study compares within the top-left cell (fitted fill against identity fill at the same split and tail) and down the left column (the ridge map against SwiftKV's trained projections).}\label{tab:1}
\footnotesize
\setlength{\tabcolsep}{4pt}
\begin{tabularx}{\linewidth}{l>{\raggedright\arraybackslash\hsize=1.157\hsize}X>{\raggedright\arraybackslash\hsize=0.843\hsize}X}
\toprule
 & Fill (every token keeps a cache entry) & Drop (pruned tokens leave the cache) \\
\midrule
Training-free & \textbf{RidgeFill} (closed-form map); POP (identity) & FastKV, GemFilter, LazyLLM, Speculative Prefill \\
Trained & SwiftKV (distilled projections); KV Prediction (trained auxiliary model); KVA projector (trained heads); YOCO and DeepSeek-V4.1-Flash (pretrained in) & TokenGate (learned gate + LoRA); SPEED (fine-tuned) \\
\bottomrule
\end{tabularx}
\end{table}

\section{Method}\label{sec:2}
Table~\ref{tab:2} lists the notation used in this section and the rest of the paper.

\begin{table}[!htbp]
\centering
\caption{Notation.}\label{tab:2}
\small
\setlength{\tabcolsep}{4pt}
\begin{tabularx}{\linewidth}{l>{\raggedright\arraybackslash\hsize=1.000\hsize}X}
\toprule
Symbol & Meaning \\
\midrule
$L$, $N$ & number of layers; number of prompt tokens \\
$S$, $T$ & split layer; tail length in tokens \\
$\ell$, $t$ & a layer; a token \\
$d$ & hidden size \\
$x_\ell$, $\hat{x}_\ell$ & true and predicted input to layer $\ell$ \\
$A_\ell$, $b_\ell$ & the map's matrix and offset for layer $\ell$ \\
$d_\text{in}$, $d_\text{out}$ & width of a map's input and output (both $d$ except on Flash-Next) \\
$\alpha$, $\lambda$ & ridge penalty; its scale-free setting \\
$c$ & tail rows used by the value rescale \\
$\Delta M_\ell$, $\rho_\ell$ & terminal-state offset; filled rows' share of the state at the end of the prefix \\
$a_{ij}$, $V$ & attention from query $i$ to row $j$; largest value norm among the rows query $i$ attends to \\
$r$ & share of exact prefill compute that remains \\
\bottomrule
\end{tabularx}
\end{table}

\subsection{Split, tail and prefix}\label{sec:2.1}
A model has $L$ layers, numbered from 0, and we write $\ell$ for any one of them. A prompt has $N$ tokens. We call each prompt token's position a row. A deployment chooses two numbers: the split $S$, which is a layer number, and the tail length $T$, which is a number of tokens. Layers $0$ to $S-1$ run exactly for every row. The last $T$ prompt tokens, the tail, run every layer exactly, as does every generated token. The remaining $N - T$ prompt tokens are the prefix.

In the late layers, from $S$ up, a prefix row is either exact or filled. An exact row runs every layer normally, like the tail; \S\ref{sec:2.4} describes how exact rows are chosen. Filled rows are where the savings come from.

In a normal forward pass, each layer's input is the previous layer's output, so reaching layer 40 means running layers 16 through 39 first. RidgeFill breaks that chain for filled rows. Write $x_\ell$ for the hidden state entering layer $\ell$. RidgeFill predicts each late layer's input, $\hat{x}_\ell$, directly from a single state, $x_S$, the hidden state entering the split layer. On the 27B, with the split at 16, a filled row's input to layer 40 comes straight from its layer-16 state, so layer 40 doesn't wait on layers 16 to 39.

From $\hat{x}_\ell$, each late layer computes only what later tokens will read from that row: keys and values in an attention layer (and, on Flash-Next, the sparse-attention indexer's keys, \S\ref{sec:5.1}), and the recurrent state update in a Gated DeltaNet layer. Everything else in the layer, including the MLP, is skipped for that row.

The tail reads the filled rows, so the tail's outputs can differ from exact even though the tail itself is computed exactly.

\begin{figure}[!htbp]
\centering
\includegraphics[width=\linewidth]{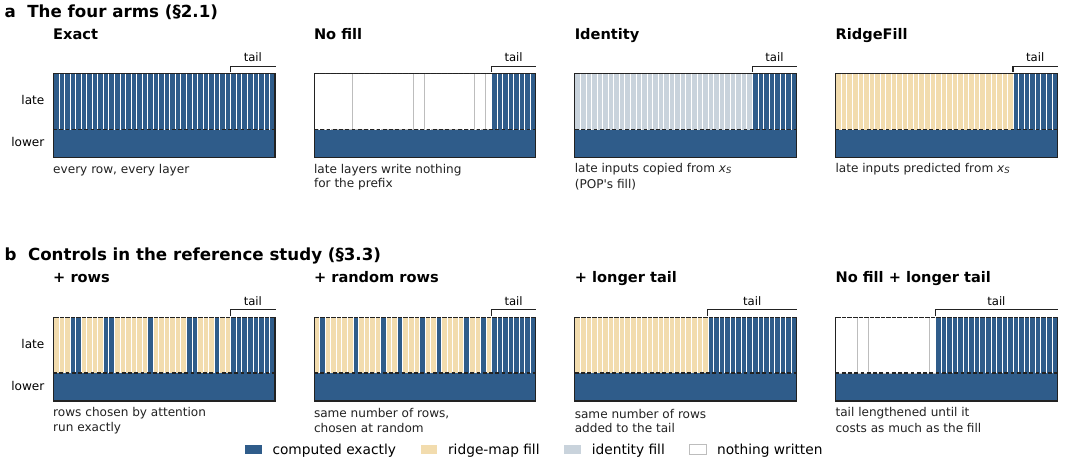}
\caption{The arms. (a) The four fills of \S\ref{sec:2.1}, shown for the layers above the split: exact, no fill, identity (POP's fill) and RidgeFill; the lower layers and the tail run exactly in every arm. (b) The reference study's controls (\S\ref{sec:3.3}): attention-selected exact rows, the same number of random rows, the same number of rows added to the tail, and no fill with the tail lengthened to the fill's cost. Row positions are illustrative.}\label{fig:3}
\end{figure}

Four ways of handling the prefix in the late layers define the experimental arms used throughout the paper (Table~\ref{tab:3}, Figure~\ref{fig:3}a).

\begin{table}[!htbp]
\centering
\caption{The four arms. $A_\ell$ and $b_\ell$ are a matrix and a vector fitted for each late layer (\S\ref{sec:2.2}).}\label{tab:3}
\small
\setlength{\tabcolsep}{4pt}
\begin{tabularx}{\linewidth}{l>{\raggedright\arraybackslash\hsize=1.000\hsize}X}
\toprule
Arm & Input to layer $\ell \ge S$ for prefix rows \\
\midrule
Exact & the true $x_\ell$; nothing is skipped \\
No fill & none; the late layers write nothing for the prefix, so there the tail attends only to itself \\
Identity & $\hat{x}_\ell = x_S$ (POP's fill) \\
RidgeFill & $\hat{x}_\ell = A_\ell x_S + b_\ell$ \\
\bottomrule
\end{tabularx}
\end{table}

\subsection{The ridge map}\label{sec:2.2}
For each filled layer $\ell$, RidgeFill predicts the layer's input by multiplying the split state by a matrix and adding a vector:

\[\hat{x}_\ell = A_\ell x_S + b_\ell .\]

Here $x_S$ and $x_\ell$ are vectors of length $d$, the model's hidden size (4,096 on the reference models). $A_\ell$ is a $d \times d$ matrix and $b_\ell$ is a vector of length $d$, and each filled layer has its own pair. On Flash-Next the input is wider than the output; see ``Size and cost'' below.

The pair is fitted on calibration text. Run the unmodified model over it, record $x_S$ and $x_\ell$ for every token $t$, and choose the $A_\ell$ and $b_\ell$ whose predictions come closest to the true inputs. Ridge regression measures ``closest'' as squared error, plus a penalty that keeps the entries of $A_\ell$ small so the map doesn't overfit the calibration text:

\[\min_{A_\ell,\, b_\ell} \sum_t \lVert x_{\ell,t} - (A_\ell x_{S,t} + b_\ell) \rVert^2 + \alpha \lVert A_\ell \rVert_F^2 .\]

The first term adds up the squared prediction error over the calibration tokens. The second is the penalty, and $\alpha$ sets its strength.

This problem has an exact solution, so nothing is trained iteratively. Stack the calibration states as rows of two matrices, $X$ for the split states and $Y_\ell$ for layer $\ell$'s inputs, after subtracting each one's mean ($\mu_S$ and $\mu_\ell$). Then

\[A_\ell = Y_\ell^\top X \,(X^\top X + \alpha I)^{-1}, \qquad b_\ell = \mu_\ell - A_\ell \mu_S .\]

The offset $b_\ell$ makes the map right on average.

At inference the predicted input goes through the layer's own normalization and projections, as a true input would. In an attention layer the keys also get their rotary embedding as usual.

\textbf{Fitting.} One capture pass runs the unmodified model over the calibration text. For each split chosen in advance, it keeps running sums of $X^\top X$, $X^\top Y_\ell$ and the two means, one document at a time, with fp32 products summed in fp64. The result is the same as solving on all rows at once. One CPU solve per filled layer finishes the fit. A new $\lambda$ needs only new solves from the saved sums; a new split needs a new pass. On the 27B, one pass for splits 16, 24 and 32, plus their solves, took 26 minutes on one RTX 5090. Appendix~\ref{app:G} gives the calibration text, row counts and fit times for each model.

\textbf{The penalty.} The penalty is set as $\alpha = \lambda\,\mathrm{tr}(X^\top X)/d_\text{in}$. This ties the penalty's strength to the size of the inputs, so the same $\lambda$ means the same thing on every model. One $\lambda$ is shared by every filled layer at a split. On the reference models and the 27B, $\lambda$ was chosen by the same search: a grid from $10^{-4}$ to 1, scored on held-out documents by the cosine similarity between what each layer writes to its cache from the predicted input and from the true input. It picked 0.03 on Llama-3.1-8B, Qwen3.5-9B and Qwen3.8-27B, and 0.01 on Qwen3-8B (0.003 at its split 18); on the reference models the next-best value scored within about 0.0004. Flash-Next's $\lambda$ was chosen on ten held-out captures by the cosine similarity between the predicted and the true layer input, and also came out at 0.03.

\textbf{Size and cost.} A map has $d_\text{out}$ rows and $d_\text{in}$ columns, where $d_\text{in}$ is the width of the split state and $d_\text{out}$ the width of the layer input. On the reference models both are 4,096, so a map is 32 MiB per filled layer in bf16. On the 27B both are 5,120, and the maps are stored as int8 with one scale per 128 weights of each row: 1.2 GiB for all 48 late layers. Flash-Next is different because it uses Qwen's Gated Residual, a form of hyper-connections (Zhu et al., 2025). Its residual stream is four parallel streams of width 2,560, and each layer reads a gated, normalized mix of them. Its maps take all four streams entering layer 24 as input ($d_\text{in} = 10{,}240$) and predict each late layer's input ($d_\text{out} = 2{,}560$); stored as int8 in the same format, its 24 maps take 0.60 GiB. Applying a map costs $d_\text{in} d_\text{out}$ multiply-accumulates per filled row per layer, on top of the layer's own projections. That cost is included in the compute-matched controls (\S\ref{sec:3.3}), in the POP comparison (\S\ref{sec:2.5}, \S\ref{sec:6.4}), and in the 27B's multiply-accumulate shares (\S\ref{sec:4.2}).

\subsection{Hybrid layers and two corrections}\label{sec:2.3}
Three of the five models are hybrids. Qwen3.5-9B, Qwen3.8-27B and Qwen3.8-Flash-Next interleave Gated DeltaNet layers (Yang et al., 2025b) with attention layers. A Gated DeltaNet layer keeps one recurrent state instead of a key-value cache. At every token, it multiplies its state by a gate between 0 and 1 and then adds the new token's update, so older tokens fade.

That difference decides where the fill's error ends up. In an attention layer, a filled row's error stays in that row's keys and values. In a Gated DeltaNet layer, every filled row updates the shared state from its predicted input, so all of their errors are folded into the state the tail inherits.

Two corrections push the filled cache back toward the exact one. Each removes an average error rather than predicting any one token's error, and neither is trained.

\textbf{Value rescale.} In each late attention layer, the first $c$ tail rows go through the layer's value projection twice: once from their true input and once from the map's prediction for the same rows. For each key-value head, the ratio of the exact value norm to the predicted one, averaged over the $c$ rows, gives a scale. The filled prefix rows' values are multiplied by that scale before the tail reads them. The scale is measured on every request, from that request's own tokens.

The correction targets values because the Qwen models normalize their keys, which fixes each key's size and leaves a key rescale little to correct (\S\ref{sec:6.6}), while values have no such normalization. On Llama-3.1-8B, which does not normalize its keys, a key rescale halves the fill's added loss (the added loss on novel tokens, \S\ref{sec:3.1}, falls from +1.006 to +0.502 nats at split 12) and a value rescale makes it worse (\S\ref{sec:6.6}). Of the three deployments, only the 27B uses the value rescale, with $c = 512$ (Table~\ref{tab:4}).

\textbf{Terminal-state correction.} Flash-Next's quality mode (\S\ref{sec:5}) corrects the recurrent state directly. At the end of the prefix, it adds $\rho_\ell\,\Delta M_\ell$ once to the state of each filled Gated DeltaNet layer.

$\Delta M_\ell$ is the average difference between the exact and the filled state, per layer and value head. It is measured every 2,048 tokens (62 states per layer) on 13 calibration prompts kept apart from the maps' calibration text and from every evaluation set.

\begin{figure}[!htbp]
\centering
\includegraphics[width=\linewidth]{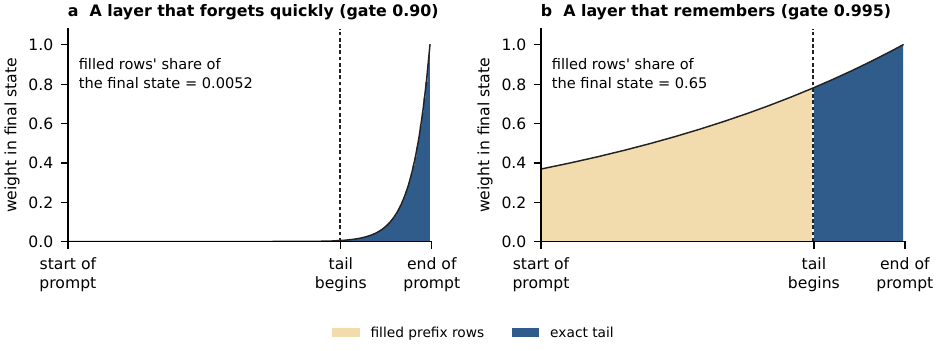}
\caption{Why the correction matters only on layers that remember. A prompt row's weight in a Gated DeltaNet layer's final state is the product of the gates after it; tan is the filled prefix and blue the exact tail. (a) With gates of 0.90, the filled rows' share of the final state is about 0.005. (b) With gates of 0.995 it is 0.65. Illustrative: 200 rows, the last 50 exact. The share shown is the share of the state the tail ends with, not $\rho$ at the end of the prefix (\S\ref{sec:2.3}).}\label{fig:4}
\end{figure}

$\rho_\ell$ is the share of the state that still comes from filled rows. Each token's update is shrunk by every gate that comes after it, so a token's weight in the state is the product of those later gates. $\rho_\ell$ is the filled rows' total weight divided by all rows' total weight, read per request from the gates the fill computes anyway. Scaling by $\rho_\ell$ makes the correction match how much of the state is actually filled: Appendix~\ref{app:H} shows that, in a simplified model without the delta rule's erase term, if the fill shifted every filled row's update to the state by the same amount, this scale removes the shift exactly and any other scale leaves part of it. Behind an exact tail, the filled rows' share of the final state is tiny on layers whose gates are small, because they have faded by the time the tail ends; the error and the correction fade together (Figure~\ref{fig:4}). Only layers whose gates stay near 1 need the correction (Appendix~\ref{app:H} bounds this).

The correction covers all 18 Gated DeltaNet layers above Flash-Next's split. The 27B and the reference study run without it.

\subsection{Exact prefix rows}\label{sec:2.4}
A chosen subset of prefix rows runs every layer exactly, in addition to the tail. These rows cost full-model compute, so the selector's job is to spend them where the fill's error would do the most damage. The paper uses three selectors, one per deployment. None has a trained component, and each works from quantities available at prefill time.

\textbf{Attention-selected rows.} The reference study (Llama-3.1-8B, Qwen3.5-9B and Qwen3-8B) keeps exact, in its ``+ rows'' arms, the 10~\% of prompt tokens that receive the most attention from the last 64 prompt tokens. Attention is summed over the heads of every attention layer among layers 3, 7, 11 and 15 that lies below the split. When a question ends the prompt, these are the tokens the question attends to, so the selector is query-aware. In the long-document loss measurements, the 64-token window is the stretch just before the scored continuation, which is not a question. Every selection is compared with two controls of equal cost: the same number of randomly chosen rows, and the same number of rows added to the exact tail.

\textbf{Why attention.} The rule rests on a bound on how much a row's error can change what a later token reads from it. Take a query $i$, a token near the end of the prompt, and let $a_{ij}$ be the attention it pays to row $j$. Let $V$ be the largest value norm among the rows query $i$ attends to. If row $j$'s value vector changes by $\delta$, query $i$'s attention output changes by exactly $a_{ij}\delta$. If row $j$'s attention score (the logit before the softmax) changes by $\varepsilon$, the output changes by at most $2V a_{ij}(e^{|\varepsilon|} - 1)$. Both bounds scale with $a_{ij}$. So if the fill's error were the same size at every row, keeping exact the rows with the largest summed attention would minimize the total of these bounds over the queries. It is not the same size everywhere, which is why the other two selectors also protect tokens by what they are. The reference selector measures attention in the lower layers, as a proxy for the late layers that read the filled rows. The 27B probe below measures it in the late layers directly.

\textbf{Protected-token rows.} The quality mode of \S\ref{sec:5}, Qwen3.8-Flash-Next, protects tokens by what they are rather than by what the question attends to. Four token classes are eligible: capitalized word starts, capitalized word pieces, lower-case word pieces, and mixed tokens (not all letters, not all punctuation, and not digits with punctuation). On a 300-document development corpus, measured on Llama-3.1-8B and Qwen3.5-9B at split 12, these classes carried 1.2 to 1.7 times their share of the fill's first-order damage (Appendix~\ref{sec:I.3}). Lower-case word starts carried 0.8 to 0.9 times their share and plain numbers 0.8 to 1.05 times theirs, so neither class is eligible. Of the eligible tokens, the rarest 25~\%, by add-one unigram frequency over the calibration documents, run exactly. On the Flash-Next run this came to 3 to 10~\% of prefix positions. The selector does not read the question.

\begin{figure}[!htbp]
\centering
\includegraphics[width=\linewidth]{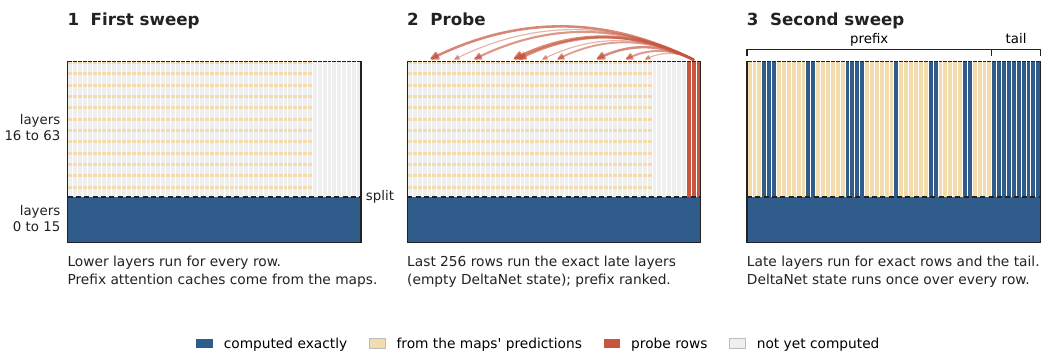}
\caption{The two-sweep prefill on Qwen3.8-27B, split at layer 16. (1) The first sweep runs layers 0 to 15 for every row and writes the prefix rows' late attention caches from the maps; the tail and the late Gated DeltaNet layers are not yet computed. (2) The probe: the last 256 rows run the exact late layers over the filled cache, with the Gated DeltaNet layers starting from an empty state, and every prefix row is ranked by the attention it receives from them. (3) The second sweep runs the late layers for the chosen rows and the tail, and carries the Gated DeltaNet state once over every prefix row; the other rows keep their filled attention caches.}\label{fig:5}
\end{figure}

\textbf{The 27B selector.} Qwen3.8-27B (\S\ref{sec:4}) keeps exact the rows chosen by a probe, plus special tokens and lexical-net rows, on top of the tail. Figure~\ref{fig:5} shows the order in which these run.

The probe runs once the first sweep (Figure~\ref{fig:5}) has run the lower layers for every row and written the prefix rows' late attention caches from the maps. The last 256 prompt rows then run the exact late layers over that filled cache, and every prefix row is ranked by the attention it receives from those 256 rows, summed over the late attention layers and their heads. The highest-ranked rows are kept exact. Because the probe reads the real late layers' attention on the real filled cache, it measures what the question attends to in the layers that will answer it.

\begin{figure}[!htbp]
\centering
\includegraphics[width=\linewidth]{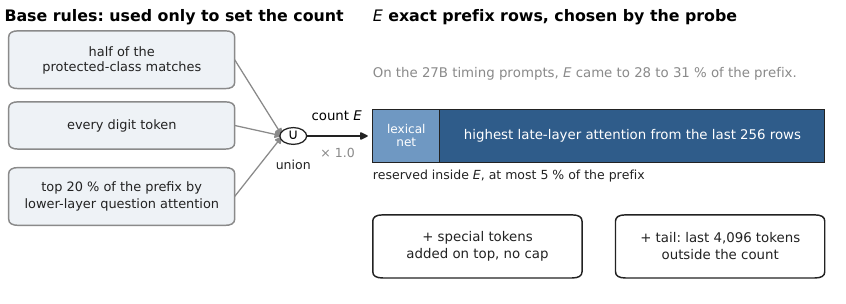}
\caption{How the probe's budget is set. The base rules, the rarer half of the protected-class matches, every digit token, and the top 20~\% of the prefix by lower-layer question attention, are used only to count $E$, the number of exact prefix rows. The probe fills $E$ with the rows that receive the most late-layer attention from the last 256 rows; lexical-net rows are reserved inside $E$, at most 5~\% of the prefix. Special tokens are added on top, and the 4,096-token tail is outside the count. On the 27B timing prompts $E$ came to 28 to 31~\% of the prefix.}\label{fig:6}
\end{figure}

The number of rows the probe keeps is a count, not a share (Figure~\ref{fig:6}). It equals the number of rows the base rules would have selected on the same prompt: the union of the rarer half of the protected-class matches, every digit token, and 20~\% of the prefix ranked by lower-layer question attention, multiplied by a budget factor that is 1.0 in this paper. The probe's ranking replaces the rows those rules would have chosen, and only their count carries over. Digit tokens therefore set the size of the budget but are not kept exact unless the probe ranks them highly. On the 27B timing prompts the count came to 28 to 31~\% of the prefix. The probe itself costs 256 rows through the late layers, about 1.6~\% of a 16K prompt's late-layer rows, plus their attention over the prefix.

Special tokens (control, unknown and user-defined tokens) also run exactly, on top of the probe's count. A prompt contains few of them, and there is no cap.

Lexical-net rows catch the identifier the question names. A sentence of the prefix is kept exact if it shares a run of four or more words with the last 64 content tokens of the prompt, and that run contains a word occurring no more than twice in the whole prefix. Content tokens exclude control and chat-template tokens, so in a chat prompt the window is the end of the user's last turn. At most the first 80 rows of a sentence are kept, and at most 5~\% of the prefix in total. Unlike special tokens, lexical-net rows count against the probe's budget. Appendix~\ref{app:H} gives the word and sentence normalization rules.

The tail is separate from all of this. It runs exactly in every deployment and never counts against the selector's budget. Its length is 4,096 tokens everywhere except Flash-Next, which uses 2,048 (Table~\ref{tab:4}).

\textbf{Two sweeps.} The 27B prefill therefore runs in two sweeps. The first sweep runs the lower layers for every row and writes the prefix rows' late attention keys and values from the maps; the tail and the late Gated DeltaNet layers wait. The probe then runs the last 256 rows through the exact late layers, with the Gated DeltaNet layers starting from an empty state, and ranks the prefix. The second sweep runs the late layers for the exact rows and the tail, reusing the filled keys and values for every other row. In the Gated DeltaNet layers the state has to be carried through the prompt in order, so the second sweep computes it once, over every prefix row: from the true input on exact rows and from the predicted input on the rest. Section~\ref{sec:4} calls this the two-sweep prefill (Figure~\ref{fig:5}).

Each selector is a rule the server applies at prefill time. The probe is the training-free counterpart of a learned gate: instead of a scorer trained to predict which rows the late layers will attend to, it reads that attention directly.

\subsection{Compute}\label{sec:2.5}
A simple count gives a ceiling on the speedup. If every layer had the same cost per token and the maps were free, the share of exact prefill compute that remains would be

\[r = \frac{S(N-T) + LT}{LN} = \frac{S}{L} + \left(1 - \frac{S}{L}\right)\frac{T}{N}.\]

The numerator counts layer passes: each of the $N - T$ prefix tokens runs $S$ layers, and each of the $T$ tail tokens runs all $L$. The denominator is exact prefill, all $N$ tokens through all $L$ layers. The speedup can be at most $1/r$. On Qwen3.8-Flash-Next ($L = 48$, $S = 24$, $T = 2{,}048$) that ceiling is 1.64$\times$ at 9K tokens and 1.88$\times$ at 32K, against measured speedups of 1.50$\times$ and 1.60$\times$ (\S\ref{sec:5}).

\begin{figure}[!htbp]
\centering
\includegraphics[width=\linewidth]{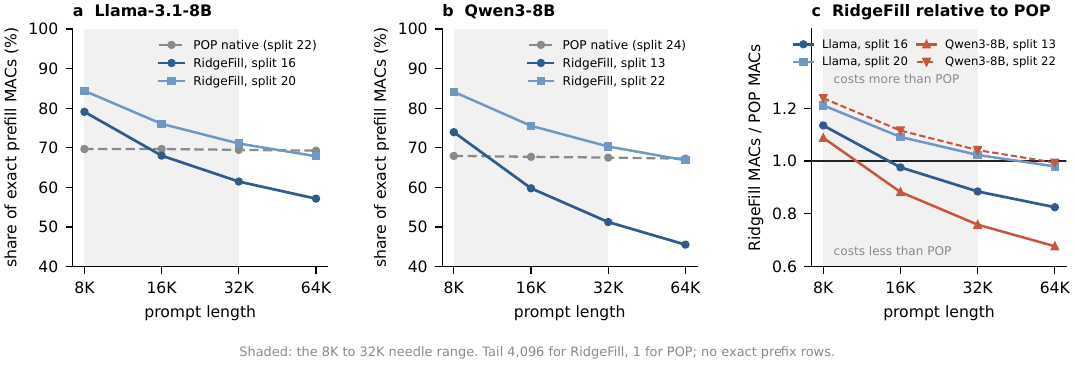}
\caption{Compute against POP, from the multiply-accumulate model of \S\ref{sec:2.5}. (a, b) Share of exact prefill compute for POP at its own split and for plain RidgeFill with a 4,096-token tail and no exact prefix rows, on Llama-3.1-8B and Qwen3-8B. (c) RidgeFill's cost relative to POP; above 1, RidgeFill costs more. Shaded: the 8K to 32K needle range.}\label{fig:7}
\end{figure}

This count leaves out the maps, which matters when a comparison is close. The comparison with POP in \S\ref{sec:6.4} therefore counts multiply-accumulates (Figure~\ref{fig:7}). The cost model counts every layer each row runs, attention over the context included, plus the maps, and it also sets the size of the compute-matched controls (\S\ref{sec:3.3}). POP fills the last third of the layers, from layer $L - \lfloor L/3 \rfloor$, with a one-token tail. The comparison uses plain RidgeFill with a 4,096-token tail and no exact prefix rows. On the 32-layer Llama-3.1-8B, RidgeFill costs less than POP above about 14K tokens at half depth and above about 46K at 5/8 depth. At 5/8 depth it therefore costs more than POP across the whole 8K to 32K needle range. On the 36-layer Qwen3-8B, split 13 costs less than POP's split 24 above about 10.6K tokens, and 24~\% less at 32K. At 5/8 depth, split 22 costs more than POP's split 24 below about 59K tokens: 1.04 times POP's cost at 32K and 0.99 times at 64K.

Three things move the real number away from $r$: exact prefix rows, the maps' own cost, and layers that do not cost the same. On the 27B, after exact rows and the tail, 54~\% of prompt tokens were filled at 16K and 67~\% at 120K, and a full count, which charges attention, the Gated DeltaNet state work that filled rows still do, the maps, the probe and the value-rescale measurement, put the share of multiply-accumulates skipped at 30~\% and 42~\% (\S\ref{sec:4.2}). The measured speedups are the numbers to use; the analytic count is context.

\subsection{The three deployments}\label{sec:2.6}
Table~\ref{tab:4} collects the settings of each deployment in one place.

\begin{table}[!htbp]
\centering
\caption{The three deployments.}\label{tab:4}
\footnotesize
\setlength{\tabcolsep}{4pt}
\begin{tabularx}{\linewidth}{>{\raggedright\arraybackslash\hsize=0.707\hsize}X>{\raggedright\arraybackslash\hsize=1.333\hsize}X>{\raggedright\arraybackslash\hsize=0.925\hsize}X>{\raggedright\arraybackslash\hsize=1.034\hsize}X}
\toprule
 & Reference study & Qwen3.8-27B & Qwen3.8-Flash-Next \\
\midrule
Models & Llama-3.1-8B, Qwen3.5-9B, Qwen3-8B & Qwen3.8-27B & Qwen3.8-Flash-Next \\
Split $S$ /\allowbreak{} layers $L$ & 12/\allowbreak{}32, 12/\allowbreak{}32, 13/\allowbreak{}36 & 16/\allowbreak{}64 & 24/\allowbreak{}48 \\
Tail $T$ & 4,096 & 4,096 & 2,048 \\
Exact prefix rows & none; attention-selected 10~\% in ``+ rows'' arms & probe, special tokens, lexical net & none; protected tokens in quality mode \\
$\lambda$ & 0.03, 0.03, 0.01 & 0.03 & 0.03 \\
Value rescale & off & on, $c = 512$ & off \\
Terminal-state correction & off & off & quality mode only \\
\bottomrule
\end{tabularx}
\end{table}

\section{Evaluation design}\label{sec:3}
\begin{figure}[!htbp]
\centering
\includegraphics[width=\linewidth]{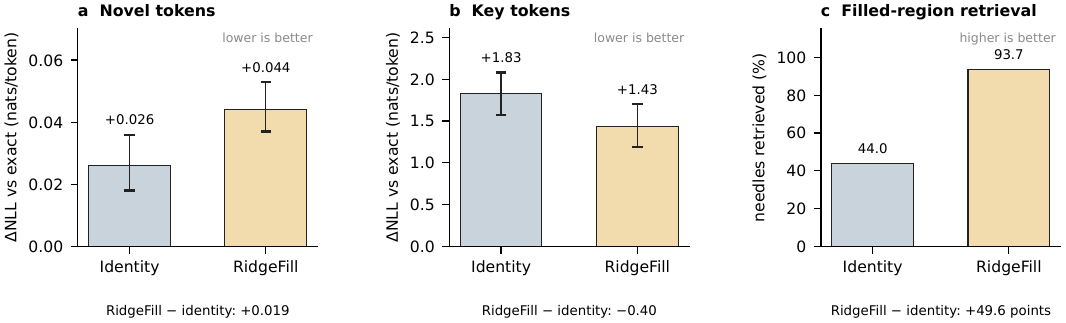}
\caption{Qwen3-8B at the primary split: loss on 90 documents (63 for key tokens) and filled-region retrieval on 138 RULER samples. On novel-token loss RidgeFill looks slightly worse than the identity fill. On the tokens that need the prefix it is better, and on retrieval from the filled region it is far better.}\label{fig:8}
\end{figure}

Any result can masquerade as noise if it is measured with the wrong stick. On Qwen3-8B at the primary split, the fitted fill and the identity fill differ by 0.019 nats of loss on novel tokens, the ones the model cannot copy from earlier in the document, a number small enough to read as nothing or as a slight cost. On the same runs the identity fill retrieves 44.0~\% of the needles that lie in the filled region and the fitted fill retrieves 93.7~\% (Figure~\ref{fig:8}). As it turns out, loss averaged over tokens was the wrong stick. Most scored tokens do not depend on the prefix, so the fill's error is diluted by tokens it cannot affect.

Every measure in this section is chosen to avoid that. A needle in the exact tail is never filled and says nothing about the fill, so filled-region retrieval excludes it. A token the model predicts as well from the last 4,096 tokens as from the whole prompt does not test the prefix cache, so key-token loss excludes it. Splice tests replace one component of the filled cache with its exact counterpart, so the loss that remains names the component. And a mean with a heavy tail is reported next to its median: on the 27B, the mean KL at 32K is driven by five copy-heavy documents, while the median is 0.004.

The line between fixing what a benchmark shows and fitting to the benchmark is thin. We drew it in three places. A fix had to be a general rule, not a pattern matched to the test: token classes, attention received, and rarity. The tests that count are official suites we did not build, with count-matched random controls beside every selected-row arm in the reference study. And once our own needle suites passed a configuration that failed RULER's multi-key tasks, we stopped using them for any decision (Appendix~\ref{sec:I.5}).

Each deployment's settings are in Table~\ref{tab:4} (\S\ref{sec:2.6}).

\subsection{What we measure}\label{sec:3.1}
\begin{figure}[!htbp]
\centering
\includegraphics[width=\linewidth]{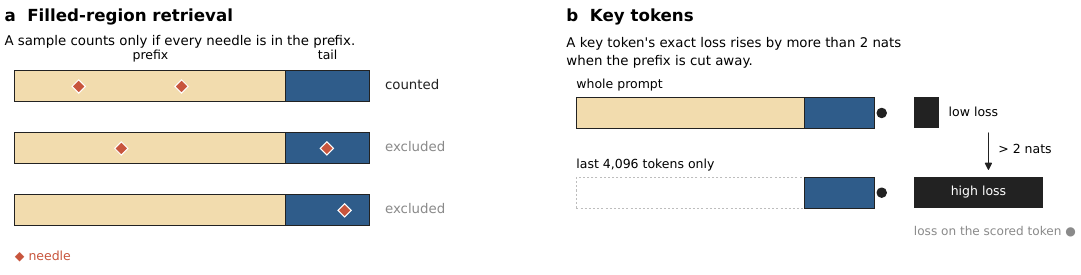}
\caption{(a) Filled-region retrieval counts a sample only if every needle lies in the prefix. (b) A key token is one whose exact loss rises by more than 2 nats when the context is cut to its last 4,096 tokens.}\label{fig:9}
\end{figure}

\textbf{Filled-region retrieval.} This scores RULER's needle tasks only on the samples whose needles all lie in the prefix, before the exact tail (Figure~\ref{fig:9}a). The set of samples is fixed per model and shared by every arm and split; this works because the tail has the same length at every split.

\textbf{Loss on tokens that need the prefix.} On long documents we report $\Delta$NLL, an arm's loss minus the exact arm's, in nats per token. We report it on two subsets of the scored tokens, because the average over all of them hides the fill's effect. Novel tokens are those not inside any 8-token run that also appears earlier in the prompt, so the model cannot simply copy them. Key tokens are those whose exact loss rises by more than 2 nats when the context is cut to its last 4,096 tokens, that is, when the prefix is removed (Figure~\ref{fig:9}b). These are the tokens the prefix actually helps, so $\Delta$NLL on key tokens measures how much of the prefix's contribution the fill keeps.

\textbf{Agreement with the full model.} On the reference models, we take the exact arm's greedy 512-token continuation of 20 documents and feed it to each arm one token at a time. The 20 are a seeded draw from the 90 long documents up to 32K (\S\ref{sec:3.2}), balanced across lengths of 8,192, 16,384 and 32,768 tokens (7, 7 and 6 documents) but not across domains; no legal document was drawn. Agreement is the share of steps where the arm's most likely next token matches the exact arm's. KL is the mean per-step divergence KL(exact $\Vert$ arm). On the 27B, both arms instead score the 512 real tokens that follow the prompt, and we report KL(stock $\Vert$ RidgeFill) with its median (\S\ref{sec:4.7}).

\textbf{Which component carries the error.} Splices replace the filled keys, the filled values, or both with their exact counterparts, in every filled attention layer at once. The share of the fill's added loss that goes away names the component (\S\ref{sec:6.6}).

\textbf{Official suites.} The 27B runs RULER's 13 tasks at 16K and 32K and LongBench v2 (Bai et al., 2025) in full (\S\ref{sec:4}). Flash-Next is scored on RULER's needle tasks and by its perplexity gap to exact on the novel tokens of 90 long documents (\S\ref{sec:5}). Llama-3.1-8B also runs eight tasks from LongBench v1 (Bai et al., 2024), 50 samples each.

\textbf{Speed.} Time to first token is measured only in the serving engines (\S\ref{sec:4.4}, \S\ref{sec:5}). The reference implementation is a measuring tool and is not timed; its compute is reported as the multiply-accumulate count of \S\ref{sec:2.5}.

\subsection{On what}\label{sec:3.2}
\textbf{Reference models.} The three reference models are the post-trained releases of Llama-3.1-8B (Grattafiori et al., 2024), Qwen3.5-9B (Qwen Team, 2026a) and Qwen3-8B (Yang et al., 2025a); we write Llama-3.1-8B for Llama-3.1-8B-Instruct, and Appendix~\ref{app:F} gives the revisions. Llama-3.1-8B has 32 attention layers and does not normalize its queries and keys before attention. Qwen3.5-9B has 32 layers, three Gated DeltaNet layers for every attention layer, and normalizes queries and keys (QK-norm); it also accepts images, and we use only its text path. Qwen3-8B has 36 attention layers with QK-norm, and runs to 32K tokens, its native context, without YaRN context extension (Peng et al., 2024). All reference results come from a layer-by-layer PyTorch implementation in bf16.

The primary split is at 3/8 of the depth (layer 12 of 32, and 13 of 36 on Qwen3-8B), so about 5/8 of the layers are filled. That is more than SwiftKV (half) or POP (a third) skips. The split curve of \S\ref{sec:6.1} covers splits from 1/4 to 5/8 of the depth.

\textbf{Long documents.} The loss measurements use 126 documents in six domains (prose, papers, chat, legal, encyclopedia, source code), from 8K to about 128K tokens, frozen before any model was run on them. They come from PG-19 (Rae et al., 2020), arXiv papers, UltraChat (Ding et al., 2023), EU legislation through LexGLUE (Chalkidis et al., 2022), English Wikipedia, and four GitHub repositories (bootstrap, django, react and regex). Qwen3-8B uses the 90 documents up to 32K, 15 per domain, because it runs only to 32K; Flash-Next uses the same 90. Each document is scored on the same stretch of text for every model: its last 512 tokens under the Qwen3.8 tokenizer. Other tokenizers split that text into slightly different numbers of tokens (Table~\ref{tab:5}). Key tokens are found once, from the exact arm, and every arm is scored on the same ones.

\textbf{RULER.} The reference study runs four needle tasks, niah\_single\_1, niah\_multikey\_1, niah\_multivalue and niah\_multiquery, with 25 samples per task and length and the official scorer (Hsieh et al., 2024). Llama and Qwen3.5 run from 8K to 64K; Qwen3-8B stops at 32K, the furthest it goes without YaRN. The filled-region subsets therefore cover different length ranges, but at the lengths all three share they are nearly the same size (Table~\ref{tab:5}). The only empty task-length cell is multiquery at 8K, where each of the 25 samples has at least one of its four needles in the tail. Qwen3-8B also runs eight RULER tasks: the four needle tasks plus cwe, fwe, qa\_1 and vt. Appendix~\ref{app:D} describes a two-turn variant, in which a second question is asked of a cache filled once.

\begin{table}[!htbp]
\centering
\caption{Evaluation sets in the reference study.}\label{tab:5}
\small
\setlength{\tabcolsep}{4pt}
\begin{tabular}{llll}
\toprule
 & Llama-3.1-8B & Qwen3.5-9B & Qwen3-8B \\
\midrule
RULER lengths & 8K--64K & 8K--64K & 8K--32K \\
Filled-region samples, all lengths & 216 & 219 & 138 \\
Filled-region samples, 8K--32K & 136 & 139 & 138 \\
Long documents & 126 & 126 & 90 \\
Scored tokens per document & 431--539 & 512 & 454--523 \\
Key tokens (documents) & 467 (80) & 605 (76) & 344 (63) \\
\bottomrule
\end{tabular}
\end{table}

\textbf{Serving deployments.} The 27B and Flash-Next are measured in their serving engines on official protocols, described with their results in \S\ref{sec:4} and \S\ref{sec:5}. Both declare the same deviations from those protocols: the question and the answer format go in a single user turn, no answer prefix is prefilled (RULER's own scripts prefill one), and thinking is off. The two engines word the instruction differently; Appendix~\ref{app:E} gives both prompts.

\subsection{Against what}\label{sec:3.3}
Every arm is compared with an exact run on the same inputs. In the reference implementation that is the exact arm of \S\ref{sec:2.1}, run in the same session. On the 27B it is stock, the unmodified upstream build, which is byte-identical to the fork with the fill switched off (\S\ref{sec:4.3}). On Flash-Next it is the fork's own exact path: on its own server start for the quality results, and alternating within the same start for the timings (\S\ref{sec:5.2}). That path was never checked against upstream, so Flash-Next's baseline is called exact, not stock. The serving deployments compare RidgeFill with their baseline only; the identity fill was not run there.

Besides the four arms of \S\ref{sec:2.1}, the reference study uses the baselines and controls in Table~\ref{tab:6} (Figure~\ref{fig:3}b).

\begin{table}[!htbp]
\centering
\caption{Baselines and controls in the reference study.}\label{tab:6}
\footnotesize
\setlength{\tabcolsep}{4pt}
\begin{tabularx}{\linewidth}{>{\raggedright\arraybackslash\hsize=0.526\hsize}X>{\raggedright\arraybackslash\hsize=1.474\hsize}X}
\toprule
Arm & Definition \\
\midrule
POP native & identity fill at POP's own split, $L - \lfloor L/3 \rfloor$, with a one-token tail \\
SwiftKV as released & the released Llama-3.1-8B-Instruct SwiftKV checkpoint (Snowflake, 2024) at its split 16, with its own one-token tail \\
SwiftKV maps + exact tail & SwiftKV's trained projections with our 4,096-token tail \\
+ rows & attention-selected exact rows (\S\ref{sec:2.4}) \\
+ random rows & the same number of rows, chosen uniformly from the prefix with a per-sample seed; the identity and RidgeFill random arms use the same rows \\
+ longer tail & the same number of rows added to the exact tail instead \\
No fill + longer tail & no fill, with the tail lengthened until it costs as much as the fill would \\
Question-blind rows & rows chosen by the attention they receive from the 64 tokens before the tail, which never include the question \\
\bottomrule
\end{tabularx}
\end{table}

The random-row control matches the selection's count, not just its cost. Suppose each exact row adds its own fixed benefit, regardless of which other rows are exact. Then a random set of rows gains, on average, the same share of the total benefit as its share of the rows: 10~\% of the rows, 10~\% of the benefit. A selection's margin over a random set of the same size therefore measures only how well it chose. A control of a different size would add a term that depends on the count alone (Table~\ref{tab:H.1}).

\subsection{With what intervals}\label{sec:3.4}
All comparisons are paired, per sample or per document, and every interval is 95~\%. Loss averages weight each document equally; accuracy weights each sample equally. Differences are computed before rounding, so a reported difference can be 0.1 away from the difference of the rounded values. The interval method depends on the deployment and the measure (Table~\ref{tab:7}). Resampling within each task or domain keeps the mix of tasks the same in every resample.

\begin{table}[!htbp]
\centering
\caption{Interval methods.}\label{tab:7}
\footnotesize
\setlength{\tabcolsep}{4pt}
\begin{tabularx}{\linewidth}{>{\raggedright\arraybackslash\hsize=0.698\hsize}X>{\raggedright\arraybackslash\hsize=0.491\hsize}X>{\raggedright\arraybackslash\hsize=1.810\hsize}X}
\toprule
Deployment & Measure & Interval \\
\midrule
Reference study, Flash-Next & all & bias-corrected and accelerated bootstrap (Efron, 1987), 5,000 resamples within task or domain \\
Qwen3.8-27B & RULER & paired item bootstrap, 10,000 resamples within task \\
Qwen3.8-27B & LongBench v2 & paired percentile bootstrap over items, 10,000 resamples, computed by our report script (data repository); sign test on discordant items \\
Qwen3.8-27B & agreement (\S\ref{sec:4.7}) & bootstrap over documents \\
\bottomrule
\end{tabularx}
\end{table}

Each non-inferiority claim compares a one-sided 95~\% lower bound against a margin fixed in advance, 2 points against exact and 3 against POP. The bound is the more conservative of two numbers: the bootstrap bound, and a binomial floor of $-100(1 - 0.05^{1/n})$ points for $n$ samples. The floor matters when both arms get nearly every sample right. The bootstrap interval then shrinks toward zero, but $n$ samples cannot rule out a failure rate of up to about $3/n$, and the floor keeps the bound from claiming more precision than the sample size allows. On Llama and Qwen3.5 the floor is the binding limit: their bounds against exact, $-$1.38 and $-$1.36 points, are exactly their floors. On Qwen3-8B the floor alone is $-$2.15 points at $n = 138$, so a 2-point test cannot pass there, and the paper makes no such claim. Its selected-rows result is reported as a drop.

Time to first token is the median of repeated requests on one server start, reported without intervals: three requests on the 27B, measured on the server, and five on Flash-Next, measured at the client. Flash-Next's 64K request time is the one timing taken across separate server starts (\S\ref{sec:5}).

\section{Qwen3.8-27B in llama.cpp}\label{sec:4}
On a 27B hybrid model in a production engine, prefill runs 1.19 to 1.31$\times$ faster with RidgeFill, depending on the prompt length and on whether the fill maps are held in VRAM or streamed from host memory. The two long-context benchmarks move by about a point; their intervals include zero but do not rule out losses of a few points.

\subsection{Model, engine, hardware}\label{sec:4.1}
Qwen3.8-27B (Qwen Team, 2026c) has 64 layers and a hidden size of 5,120. Every fourth layer is attention and the rest are Gated DeltaNet, so of the 48 layers above the split, 12 are attention and 36 are Gated DeltaNet. We serve unsloth's Q4\_K\_XL GGUF at revision 4ca72078 (unsloth, 2026).

Both arms run llama.cpp. Stock is unmodified upstream at commit 1242be66f. RidgeFill runs in our fork at commit b0439f7bb. The token-level measurements use a scorer build at commit b70b0c5b4, which differs from the runtime only in its tools and tests. Every flag is the same in both arms except RidgeFill itself: batch 2048, micro-batch 512, f16 KV cache, flash attention on, cache reuse off, thinking off. RULER decodes at temperature 0 with top-k 1. LongBench decodes at temperature 0.1 with a fixed seed per item, shared by both arms.

The machine has four A100 PCIe 40 GB GPUs, and the model fits on one. Every timing and multi-user measurement ran on a single GPU, with both arms on the same device. RULER and LongBench spread their items across the four GPUs, and LongBench ran on a second machine of the same type on the same day.

\subsection{Configuration}\label{sec:4.2}
The split is at layer 16. Layers 0 to 15 run exactly for every row. In layers 16 to 63, most prefix rows are filled: their inputs come from 48 per-layer maps, one per late layer, stored as int8 with one scale per 128 weights of each row (1.2 GiB in total) and served as the fork's Q8\_0 conversion of that release (Appendix~\ref{app:F}). The remaining prefix rows run exactly: the rows the probe ranks highest, special tokens, and lexical-net rows (\S\ref{sec:2.4}). The last 4,096 prompt tokens also run exactly, as the tail.

On the timing prompts, exact rows made up 28 to 31~\% of the prefix. The probe chose all of them; special-token and lexical-net rows contributed none on these prompts. Counting the tail, 54~\% of prompt tokens were filled at 16K and 67~\% at 120K (Figure~\ref{fig:10}a). The analytic share of multiply-accumulates skipped, net of the cost of the maps and the probe, was 30~\% at 16K and 42~\% at 120K.

Of the method's corrections (\S\ref{sec:2.3}), the value rescale is on: it is measured on the first 512 tail rows and applied to the filled values. The Gated DeltaNet terminal-state correction is off.

The maps either stay in VRAM (resident) or stream from host memory one layer at a time (streamed). The two modes produce identical outputs.

On this model RidgeFill is compared against stock only. The identity fill, which is POP's fill at the same split and tail (\S\ref{sec:6}), was not run.

\subsection{Equivalence checks}\label{sec:4.3}
Three checks tie the measurements to the baseline. First, with RidgeFill switched off, the fork's logits are byte-identical to upstream's in seven comparisons: prompts at 8K, 16K and 32K, two LongBench prompts at about 32K and 120K, and the scorer's loss on two documents. Second, resident and streamed modes produce identical 128-token outputs and identical scorer loss at 16K and 64K. Third, the builds used for measurement and the instrumented builds used for these checks produce identical 128-token outputs and top-10 probabilities.

\subsection{Time to first token and memory}\label{sec:4.4}
Time to first token (TTFT) here is the warm prompt time: the server's prompt-processing time for a request sent after one warm-up request on the same server, taking the median of three such requests (Table~\ref{tab:8}). Stock and RidgeFill run on the same physical GPU. Client-side time to first token differs from this by 8 to 63 ms.

\begin{table}[!htbp]
\centering
\caption{Qwen3.8-27B time to first token and peak VRAM over stock, one A100 PCIe 40 GB.}\label{tab:8}
\footnotesize
\setlength{\tabcolsep}{4pt}
\begin{tabular}{lllllll}
\toprule
\begin{tabular}[b]{@{}l@{}}Prompt \\ tokens\end{tabular} & Stock s & Resident s & Speedup & Streamed s & Speedup & \begin{tabular}[b]{@{}l@{}}VRAM over stock, \\ resident /\allowbreak{} streamed\end{tabular} \\
\midrule
16,384 & 12.97 & 10.38 & 1.249$\times$ & 10.88 & 1.192$\times$ & +1,592 /\allowbreak{} +268 MiB \\
32,768 & 27.49 & 21.11 & 1.302$\times$ & 21.69 & 1.267$\times$ & +1,802 /\allowbreak{} +478 MiB \\
65,536 & 60.50 & 46.28 & 1.307$\times$ & 46.55 & 1.300$\times$ & +2,270 /\allowbreak{} +944 MiB \\
120,000 & 128.10 & 100.01 & 1.281$\times$ & 99.06 & 1.293$\times$ & +3,014 /\allowbreak{} +1,732 MiB \\
\bottomrule
\end{tabular}
\end{table}

\begin{figure}[!htbp]
\centering
\includegraphics[width=\linewidth]{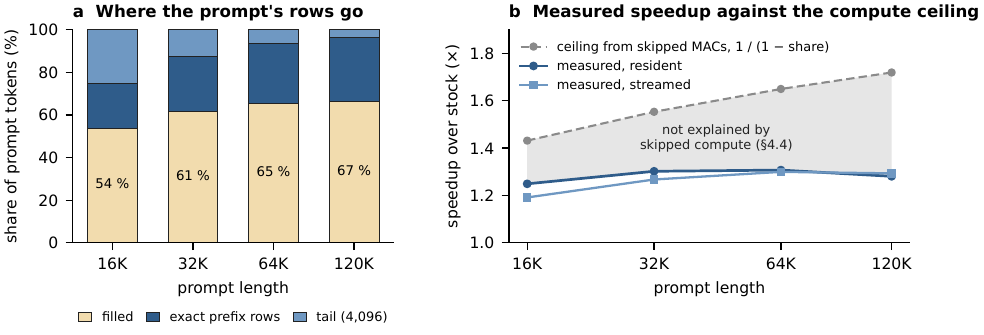}
\caption{Qwen3.8-27B on one A100. (a) Where the prompt's rows go: filled, exact prefix rows, and the 4,096-token tail, as shares of prompt tokens. (b) Measured speedup over stock, resident and streamed, against the ceiling if time were proportional to multiply-accumulates, 1/(1 $-$ skipped share). The shaded gap is discussed in \S\ref{sec:4.4}.}\label{fig:10}
\end{figure}

Streaming the maps costs 0.5 s at 16K and nothing at 120K, and it cuts the extra VRAM by 1.3 GiB. On the LongBench workload itself, run as a benchmark in which the two arms ran on different GPUs for 462 of the 503 items, prefill was 1.180$\times$ faster. That figure is supporting evidence only; Table~\ref{tab:8} follows the timing protocol.

The measured speedup is smaller than the skipped compute suggests (Figure~\ref{fig:10}b). If prefill time were proportional to multiply-accumulates, skipping 30~\% at 16K and 42~\% at 120K would allow 1.43$\times$ and 1.72$\times$. The measured speedups are 1.25$\times$ and 1.28$\times$, and above 32K they level off at about 1.3$\times$ even as the skipped share keeps rising. The analytic share counts every multiply-accumulate alike, and time does not. Three costs are the likely causes; we have not profiled how the difference divides among them. First, the exact rows and the tail still attend over the whole context in the 12 late attention layers; that attention is counted in the skipped share, but it runs at a lower rate than the dense layers and grows with the square of the prompt length. Second, each micro-batch of the second sweep reads the late layers' weights in full whether it holds 512 exact rows or 50, so scattered exact rows pay stock's weight traffic for a fraction of stock's work. Third, the probe runs 256 rows through the late layers over the filled cache before the exact rows run.

\subsection{RULER}\label{sec:4.5}
RULER at 16K and 32K (Tables~\ref{tab:9} and~\ref{tab:10}): 13 tasks, 20 items each, with the official answer budgets and the question given as an instruction (\S\ref{sec:3.2}).

\begin{table}[!htbp]
\centering
\caption{RULER overall. Intervals are paired item bootstraps, 10,000 resamples, resampled within task.}\label{tab:9}
\small
\setlength{\tabcolsep}{4pt}
\begin{tabular}{lllll}
\toprule
Context & n & Stock~\% & RidgeFill~\% & $\Delta$ points [95~\% CI] \\
\midrule
16,384 & 260 & 85.69 [83.15, 88.04] & 84.51 [81.87, 87.00] & $-$1.19 [$-$3.13, +0.79] \\
32,768 & 260 & 87.85 [85.46, 90.04] & 86.42 [83.88, 88.81] & $-$1.42 [$-$3.21, +0.35] \\
\bottomrule
\end{tabular}
\end{table}

\begin{table}[!htbp]
\centering
\caption{RULER by task.}\label{tab:10}
\small
\setlength{\tabcolsep}{4pt}
\begin{tabular}{lllll}
\toprule
Task (n = 20) & 16K stock & \begin{tabular}[b]{@{}l@{}}16K \\ RidgeFill\end{tabular} & 32K stock & \begin{tabular}[b]{@{}l@{}}32K \\ RidgeFill\end{tabular} \\
\midrule
cwe & 64.00 & 54.00 & 82.00 & 68.50 \\
fwe & 100.00 & 98.33 & 100.00 & 100.00 \\
niah\_multikey\_1 /\allowbreak{} 2 /\allowbreak{} 3 & 100.00 & 100.00 & 100.00 & 100.00 \\
niah\_multiquery & 100.00 & 98.75 & 100.00 & 98.75 \\
niah\_multivalue & 100.00 & 97.50 & 100.00 & 96.25 \\
niah\_single\_1 /\allowbreak{} 2 /\allowbreak{} 3 & 100.00 & 100.00 & 100.00 & 100.00 \\
qa\_1 & 90.00 & 85.00 & 85.00 & 85.00 \\
qa\_2 & 60.00 & 65.00 & 75.00 & 75.00 \\
vt & 0.00 & 0.00 & 0.00 & 0.00 \\
\bottomrule
\end{tabular}
\end{table}

Every needle task scores at least 96~\% in both arms, and no task other than common-word extraction (cwe) moves by more than 5 points. cwe changes by $-$10.0 points at 16K [$-$32.0, +12.0] and $-$13.5 at 32K [$-$36.0, +9.0], on 20 items each. RidgeFill's cwe answers also hit the length cap more often, 12 against 8 at 16K and 7 against 4 at 32K. Variable tracking (vt) scores zero in both arms: without RULER's prefilled answer prefix (\S\ref{sec:3.2}), every answer spends the official 30-token budget before reaching the value. The cell carries no information; it is kept so that the 13-task mean covers the official task list.

\subsection{LongBench v2}\label{sec:4.6}
LongBench v2 (Bai et al., 2025), run with the official prediction script in chain-of-thought mode at full length: a 1,024-token first pass, a 128-token extraction pass, and the official middle cut at 120,000 tokens, over all 503 items.

\begin{table}[!htbp]
\centering
\caption{LongBench v2, official CoT protocol, full length. Intervals are paired percentile bootstraps over items, 10,000 resamples, computed by our report script.}\label{tab:11}
\small
\setlength{\tabcolsep}{4pt}
\begin{tabular}{lllll}
\toprule
Group & n & Stock~\% & RidgeFill~\% & $\Delta$ points [95~\% CI] \\
\midrule
Overall & 503 & 38.57 & 37.18 & $-$1.39 [$-$4.97, +2.19] \\
Easy & 192 & 42.71 & 40.62 & $-$2.08 [$-$8.85, +4.69] \\
Hard & 311 & 36.01 & 35.05 & $-$0.96 [$-$5.14, +3.54] \\
Short & 180 & 46.11 & 44.44 & $-$1.67 [$-$7.22, +4.44] \\
Medium & 215 & 38.14 & 36.74 & $-$1.40 [$-$7.44, +4.65] \\
Long & 108 & 26.85 & 25.93 & $-$0.93 [$-$7.41, +6.48] \\
Single-document QA & 175 & 44.00 & 42.29 & $-$1.71 [$-$8.00, +4.57] \\
Multi-document QA & 125 & 30.40 & 34.40 & +4.00 [$-$2.40, +10.40] \\
Long in-context learning & 81 & 38.27 & 30.86 & $-$7.41 [$-$18.52, +3.70] \\
Long-dialogue history & 39 & 48.72 & 41.03 & $-$7.69 [$-$23.08, +7.69] \\
Code repository & 50 & 34.00 & 34.00 & 0.00 [$-$10.00, +12.00] \\
Long structured data & 33 & 36.36 & 36.36 & 0.00 [$-$12.12, +12.12] \\
\bottomrule
\end{tabular}
\end{table}

No interval excludes zero, overall or in any group (Table~\ref{tab:11}). The two largest drops, long in-context learning and long-dialogue history, come to 6 items of 81 and 3 of 39. Across all items, stock answered 48 that RidgeFill missed and RidgeFill answered 41 that stock missed; a sign test on these discordant items gives p = 0.525.

The protocol runs at its limits on this model. The first pass hit its 1,024-token cap on 301 items for stock and 302 for RidgeFill. About 185 items per arm produced no extractable answer (189 stock, 181 RidgeFill), and overall accuracy is 38~\% on a four-choice benchmark. The comparison is still fair, since both arms meet the same limits on the same items with the same seeds, but the headroom is small.

\subsection{Agreement with the unmodified model}\label{sec:4.7}
For 100 LongBench documents at each context length $N$, both arms read the first $N + 512$ tokens, and we score the last 512 under each. KL is KL(stock $\Vert$ RidgeFill) over the full vocabulary at each position. $\Delta$NLL is RidgeFill's loss minus stock's, so a negative value means RidgeFill gives the actual next token more probability.

\begin{table}[!htbp]
\centering
\caption{Post-prompt KL and $\Delta$NLL, 100 documents per length, 51,200 positions.}\label{tab:12}
\footnotesize
\setlength{\tabcolsep}{4pt}
\begin{tabular}{lllllll}
\toprule
Context & Mean KL [CI] & Median KL & p99 KL & \begin{tabular}[b]{@{}l@{}}Top-1 \\ agreement\end{tabular} & Mean $\Delta$NLL [CI] & \begin{tabular}[b]{@{}l@{}}Median \\ $\Delta$NLL\end{tabular} \\
\midrule
16,384 & 0.0894 [0.0399, 0.1556] & 0.0040 & 0.965 & 94.64~\% & $-$0.0330 [$-$0.0875, +0.0059] & +0.0044 \\
32,768 & 0.2656 [0.0599, 0.6017] & 0.0042 & 12.77 & 93.60~\% & $-$0.2142 [$-$0.5629, $-$0.0025] & +0.0071 \\
\bottomrule
\end{tabular}
\end{table}

At the median position the two arms nearly agree (Table~\ref{tab:12}), and the top-1 token matches at 94~\% of positions. The means are carried by a few documents. Splitting the documents by how much of the scored text is a verbatim copy of earlier text (the token and the 8 tokens before it already occur earlier in the window, a slightly stricter rule than \S\ref{sec:3.1}'s novel-token mask) shows which ones (Table~\ref{tab:13}):

\begin{table}[!htbp]
\centering
\caption{Post-prompt KL and $\Delta$NLL by copyable share.}\label{tab:13}
\small
\setlength{\tabcolsep}{4pt}
\begin{tabular}{lllll}
\toprule
Context & \begin{tabular}[b]{@{}l@{}}Copyable \\ share\end{tabular} & Docs & Mean KL & Mean $\Delta$NLL \\
\midrule
16,384 & \textless{} 0.2 & 86 & 0.0461 & $-$0.0055 \\
16,384 & 0.2 to 0.5 & 7 & 0.2949 & $-$0.1540 \\
16,384 & $\geq$ 0.5 & 7 & 0.4157 & $-$0.2488 \\
32,768 & \textless{} 0.2 & 82 & 0.0352 & +0.0069 \\
32,768 & 0.2 to 0.5 & 13 & 0.3074 & $-$0.0849 \\
32,768 & $\geq$ 0.5 & 5 & 3.9351 & $-$4.1765 \\
\bottomrule
\end{tabular}
\end{table}

On the copy-heavy documents, RidgeFill's loss is far below stock's. That is not RidgeFill being right. We ran the full-precision BF16 model on 25 documents at 32K chosen by disagreement: the ten with the largest KL where RidgeFill's loss was lower, the ten with the largest KL where it was higher, and five near the median. On 22 of the 25, the Q4 stock model is closer to BF16 than RidgeFill is.

The extreme case is a code-repository item whose scored tail at 32K repeats earlier text verbatim (copyable share 1.00). Stock's loss there is 14.37 nats per token and BF16's is 15.87, both above the 12.4 nats of a uniform guess over the 248,320-token vocabulary. RidgeFill's is 0.005. The model itself fails to copy there, confidently, and RidgeFill copies. The large-KL documents are RidgeFill departing from the model, in both directions.

RULER and the agreement study stop at 32K. At the lengths where the speedup is largest, the quality evidence is LongBench v2's long group: 108 items, $-$0.9 points, interval through zero.

\subsection{Serving several users}\label{sec:4.8}
One 16K request arrives while C $-$ 1 other users are decoding, so C counts every user, the new request included. RidgeFill runs in streamed mode, on one GPU, with two repetitions per cell. In the unified schedule all users share one KV cache; in the separate-KV schedule each user slot has its own. The other users' throughput is the tokens delivered to them divided by the time from the first request to the last token, so it includes the stall while the new request prefills. ITL is their inter-token latency, from client timestamps (Table~\ref{tab:14}).

\begin{table}[!htbp]
\centering
\caption{Mixed serving, two repetitions per cell. These runs are separate from those of Table~\ref{tab:8}.}\label{tab:14}
\footnotesize
\setlength{\tabcolsep}{4pt}
\begin{tabular}{llllll}
\toprule
Schedule & \begin{tabular}[b]{@{}l@{}}New request \\ TTFT s, stock /\allowbreak{} \\ RidgeFill\end{tabular} & \begin{tabular}[b]{@{}l@{}}Other users \\ tok/\allowbreak{}s, stock /\allowbreak{} \\ RidgeFill\end{tabular} & \begin{tabular}[b]{@{}l@{}}Others' ITL \\ mean ms, stock \\ /\allowbreak{} RidgeFill\end{tabular} & \begin{tabular}[b]{@{}l@{}}Others' ITL p99 \\ ms, stock /\allowbreak{} \\ RidgeFill\end{tabular} & \begin{tabular}[b]{@{}l@{}}Others' ITL max ms, \\ stock /\allowbreak{} RidgeFill\end{tabular} \\
\midrule
Unified C1 & 13.06 /\allowbreak{} 11.11 & n/\allowbreak{}a & n/\allowbreak{}a & n/\allowbreak{}a & n/\allowbreak{}a \\
Unified C2 & 13.53 /\allowbreak{} 11.70 & 19.92 /\allowbreak{} 21.60 & 49.33 /\allowbreak{} 45.50 & 1,605.7 /\allowbreak{} 863.0 & 1,707.8 /\allowbreak{} 1,317.1 \\
Unified C4 & 14.18 /\allowbreak{} 12.21 & 60.06 /\allowbreak{} 62.96 & 46.47 /\allowbreak{} 44.20 & 148.0 /\allowbreak{} 669.0 & 1,785.0 /\allowbreak{} 1,323.7 \\
Unified C8 & 15.32 /\allowbreak{} 13.53 & 86.90 /\allowbreak{} 89.48 & 71.58 /\allowbreak{} 69.35 & 621.7 /\allowbreak{} 774.9 & 1,919.6 /\allowbreak{} 1,876.1 \\
Separate KV C2 & 13.48 /\allowbreak{} 11.34 & 19.91 /\allowbreak{} 21.87 & 49.38 /\allowbreak{} 44.90 & 1,592.2 /\allowbreak{} 858.3 & 1,703.4 /\allowbreak{} 1,310.7 \\
Separate KV C8 & 13.93 /\allowbreak{} 11.86 & 88.88 /\allowbreak{} 91.98 & 69.84 /\allowbreak{} 67.46 & 584.7 /\allowbreak{} 654.3 & 1,821.1 /\allowbreak{} 1,809.6 \\
\bottomrule
\end{tabular}
\end{table}

With RidgeFill, the new request's time to first token is 12 to 16~\% shorter, and the other users' throughput rises 3 to 10~\%, because the shorter prefill blocks them for less time. Their worst stall is shorter in every schedule. Their 99th-percentile stall is shorter at two users and longer at four and eight, where RidgeFill's two-sweep prefill (\S\ref{sec:2.4}, Figure~\ref{fig:5}) yields more often in smaller steps. The stock value at four users (148 ms, against 1,606 and 622 ms at two and eight) comes from two repetitions and should be read with that in mind. All runs completed with no engine errors.

\section{Qwen3.8-Flash-Next in vLLM}\label{sec:5}
On a 48-layer hybrid mixture-of-experts model in vLLM, prefill runs 1.50 to 1.60$\times$ faster with RidgeFill at 9K to 32K tokens, and a 64K request finishes in 47.5 s instead of 82.6 s. RULER's needle tasks score 99.8 against exact's 100.0. Perplexity on the novel tokens of long documents rises 2.65~\%, with an interval from $-$0.02~\% to +4.93~\% that does not exclude zero.

This deployment differs from the 27B in engine, architecture and hardware, and it reads most routed-expert weights over PCIe during prefill.

\subsection{Model, engine, hardware}\label{sec:5.1}
Qwen3.8-Flash-Next (Qwen Team, 2026b) has 48 layers. Every fourth layer, 12 in all, is a sparse-attention layer, and the other 36 are Gated DeltaNet. Each sparse-attention layer has a small indexer, four 128-wide query heads sharing one key head, that picks which blocks of keys each query attends to. The feed-forward blocks are mixtures of experts, and the residual stream is four Gated Residual streams (\S\ref{sec:2.2}).

We serve an MXFP4/FP8 quantization of the model (Smith, 2026a) in a vLLM fork (Smith, 2026b). One file in the fork is replaced to make the indexer's block selection deterministic, which costs 5 to 7~\% of decode throughput; every arm runs with it (Appendix~\ref{app:F} lists the builds).

The model runs tensor-parallel on two AMD Radeon AI PRO R9700 GPUs. About two thirds of the routed experts, 16,042 of the 24,576 on each GPU, live in pinned host memory and are read over PCIe during prefill. Every arm uses the same placement.

\subsection{Configuration}\label{sec:5.2}
The split is at layer 24 of 48, which leaves 6 sparse-attention and 18 Gated DeltaNet layers above it, and the exact tail is 2,048 tokens. The maps are the public release \texttt{TheCacheIsALie/\allowbreak{}RidgeFill-\allowbreak{}Projector-\allowbreak{}Qwen3.\allowbreak{}8-\allowbreak{}Flash-\allowbreak{}Next-\allowbreak{}i8}: one int8 map per late layer, fitted with $\lambda$ = 0.03 on activations of the quantized model as served, captured through the engine from 151 calibration documents (\S\ref{sec:2.2}, Appendix~\ref{app:G}). In the sparse-attention layers, filled rows write the indexer's keys as well as the attention keys and values. Plain RidgeFill keeps no exact prefix rows and uses neither correction. Quality mode adds the terminal-state correction and protected-token rows (\S\ref{sec:2.3}, \S\ref{sec:2.4}). Table~\ref{tab:4} (\S\ref{sec:2.6}) lists both configurations.

Exact, plain RidgeFill and quality mode each ran their quality measurements on their own server start, meaning their own launch of the serving process; exact also runs inside the other two starts for the timings. The perplexity and RULER results are paired per document or per sample against exact's start; only the timings alternate with exact inside a single start. Two checks support comparing across starts, and one is missing. Exact's two perplexity passes on its start are bitwise identical on all 90 documents. With the fill switched off, the quality-mode start reproduces exact bitwise on 6 of 6 documents. The plain RidgeFill start ran no such check, though it used the same container image, replaced file and arguments.

\subsection{Results}\label{sec:5.3}
\textbf{How each number is measured.} Table~\ref{tab:15} gives quality and Table~\ref{tab:16} speed. The perplexity gap is the relative increase in perplexity over exact on novel tokens, $\exp(\Delta\mathrm{NLL}) - 1$. It is computed per document over the 90 long documents of \S\ref{sec:3.2}, averaged within each domain, and then averaged across domains; with 15 documents per domain, that equals the plain average over documents. RULER is the four needle tasks of \S\ref{sec:3.2} at 8K to 32K plus niah\_single\_1 at 64K: 325 samples, 221 of them in the filled region.

Time to first token is measured at the client, not on the server as on the 27B (\S\ref{sec:4.4}). The prompts are the first 9,216, 16,384 and 32,768 tokens of one frozen chat document. Each length gets one warm-up and five timed requests, alternating with exact on the same server start, and the speedup is the ratio of the two medians.

The 64K figure is measured differently. It is the median time of a whole request, prefill plus a short answer, over the 25 RULER samples at 64K, and it is taken across separate server starts. Exact's 64K figure comes from its own start only; its time to first token at 9K to 32K, measured inside all three starts, agrees across them to within 1.4~\%.

\begin{table}[!htbp]
\centering
\caption{Quality on Qwen3.8-Flash-Next. The perplexity gap is measured on novel tokens. RULER has no interval: RidgeFill fell short of exact on two of 221 filled-region samples (\S\ref{sec:5.3}).}\label{tab:15}
\small
\setlength{\tabcolsep}{4pt}
\begin{tabular}{llll}
\toprule
Arm & Perplexity gap [95~\% CI] & RULER, all & \begin{tabular}[b]{@{}l@{}}RULER, \\ filled \\ region\end{tabular} \\
\midrule
Exact & n/\allowbreak{}a & 100.0 & 100.0 \\
RidgeFill & +2.65~\% [$-$0.02~\%, +4.93~\%] & 99.8 & 99.8 \\
Quality mode & +1.60~\% [$-$0.83~\%, +4.32~\%] & 99.9 & 99.9 \\
\bottomrule
\end{tabular}
\end{table}

\begin{table}[!htbp]
\centering
\caption{Speedup over exact on Qwen3.8-Flash-Next; the last column also gives the median request time.}\label{tab:16}
\small
\setlength{\tabcolsep}{4pt}
\begin{tabular}{lllll}
\toprule
Arm & TTFT 9K & TTFT 16K & TTFT 32K & 64K request \\
\midrule
Exact & 1.00$\times$ & 1.00$\times$ & 1.00$\times$ & 1.00$\times$ (82.6 s) \\
RidgeFill & 1.50$\times$ & 1.54$\times$ & 1.60$\times$ & 1.74$\times$ (47.5 s) \\
Quality mode & 1.25$\times$ & 1.22$\times$ & 1.27$\times$ & 1.49$\times$ (55.6 s) \\
\bottomrule
\end{tabular}
\end{table}

\begin{figure}[!htbp]
\centering
\includegraphics[width=\linewidth]{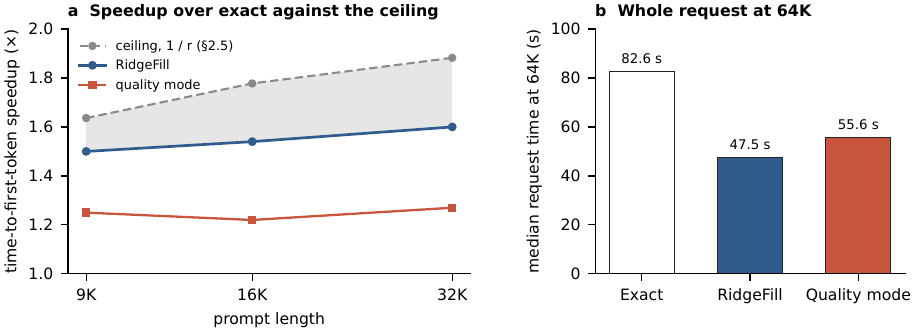}
\caption{Serving on Qwen3.8-Flash-Next. (a) Time-to-first-token speedup over exact, each a ratio of medians of five timed requests paired within a server start, against the ceiling 1/r of \S\ref{sec:2.5}. (b) Median time of a whole request over 25 single-needle samples at 64K, measured on separate server starts. No intervals were computed for the timings.}\label{fig:11}
\end{figure}

On the filled-region RULER samples, RidgeFill fell short of exact on two of 221. Both were 32K niah\_multivalue samples that returned three of their four values; quality mode missed one of the two.

Quality mode lowers the gap to +1.60~\%, but its paired difference from plain RidgeFill is $-$1.02~\% [$-$2.87~\%, +0.80~\%], so the improvement is not established. It also gives up about half of RidgeFill's gain over exact, at 1.22 to 1.27$\times$ against 1.50 to 1.60$\times$. The terminal-state correction is one addition per layer at the end of the prefix, so the protected rows are the likely cost; the final runs do not separate the two. Protected rows are scattered through the prompt, so nearly every 2,048-token chunk reads most of the late layers' expert weights, two thirds of them over PCIe, for a few dozen to a few hundred rows (Appendix~\ref{app:H}). Quality mode is therefore optional.

The measured speedups are 85 to 92~\% of the ceiling from \S\ref{sec:2.5} (Figure~\ref{fig:11}a): 1.50$\times$ against 1.64$\times$ at 9K, 1.54$\times$ against 1.78$\times$ at 16K, and 1.60$\times$ against 1.88$\times$ at 32K. The ceiling counts the layers the fill skips and the tail it doesn't. It leaves out the maps' own cost and the different costs of attention, Gated DeltaNet and expert layers; as on the 27B (\S\ref{sec:4.4}), we have not profiled the split.

This deployment does not test the premise that prefill is compute-bound with bandwidth to spare. Filled rows run no expert block in layers 24 to 47, and plain RidgeFill keeps no exact prefix rows, so only the tail reads the late layers' experts. If each chunk of the prompt reads only the offloaded experts it uses, expert traffic over PCIe shrinks by the same ratio as compute, to about 0.61 of exact at 9K and 0.53 at 32K (the $r$ of \S\ref{sec:2.5}). A compute-bound prefill and a PCIe-bound one therefore predict the same ceiling, and reaching 85 to 92~\% of it can't tell them apart. Expert traffic was not measured. The premise rests on the 27B, where the model's weights are all held in GPU memory and streaming the maps from host memory cost 0.5 s at 16K and nothing measurable at 120K (\S\ref{sec:4.4}).

\section{Reference study}\label{sec:6}
The reference study runs the fill on three 8 to 9B models in a layer-by-layer implementation, where every arm can be compared with exact and with controls of equal cost. Four results come out of it:

\begin{itemize}[leftmargin=*,itemsep=2pt,topsep=4pt]
\item On retrieval from the filled region, the fitted fill beats the identity fill by 39.7 points on Llama-3.1-8B and 49.6 on Qwen3-8B, the two dense models, and by 3.3 points, with an interval through zero, on the hybrid Qwen3.5-9B. Llama's margin is over 8K to 64K; at 8K to 32K, the lengths Qwen3-8B covers, it is 49.1.
\item On novel-token loss, depth decides. On Llama the fitted fill is worse than the identity fill at 1/4 and 3/8 depth and far better at half depth and 5/8, while the Qwen models stay within a few hundredths of a nat of exact at every depth.
\item Attention-selected exact rows restore retrieval on Llama and Qwen3.5 and hurt it on Qwen3-8B.
\item The splices put the error in the keys on Llama, in the values and the recurrent state on Qwen3.5, and in keys and values together on Qwen3-8B.
\end{itemize}

\subsection{Split curve}\label{sec:6.1}
\begin{figure}[!htbp]
\centering
\includegraphics[width=\linewidth]{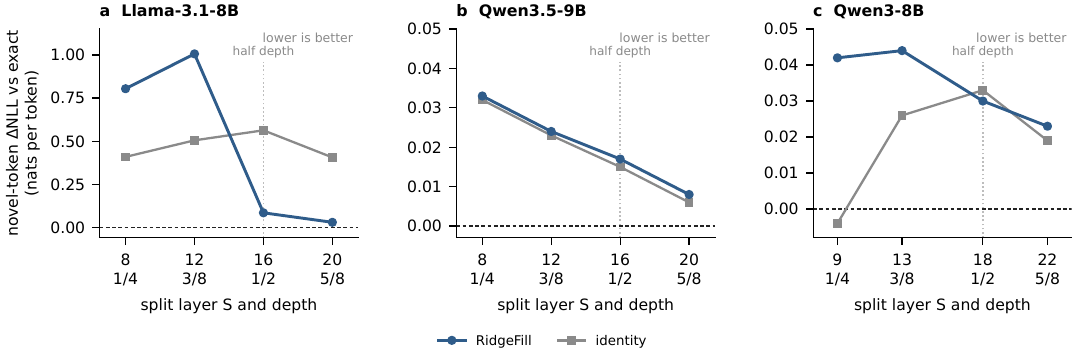}
\caption{Novel-token loss across split depths. RidgeFill and identity use the same split and a 4,096-token tail. Lower is better; the dashed line at zero is exact. Each panel has its own vertical scale. Llama improves sharply at half depth ($S = 16$), while both Qwen models stay within a few hundredths of a nat of exact. The horizontal axis shows the split layer and its depth; Qwen3-8B uses 9, 13, 18 and 22 of 36. The paired RidgeFill $-$ identity intervals are in Table~\ref{tab:17}; the figure has no per-arm error bars.}\label{fig:12}
\end{figure}

Table~\ref{tab:17} gives novel-token $\Delta$NLL against exact at four split depths, all with the 4,096-token tail (Figure~\ref{fig:12}); POP native uses its own split and a one-token tail. Both Qwen models stay within a few hundredths of a nat of exact at every depth, with the fitted and identity fills close together, so on those models the fill's value shows on retrieval and key-token loss (\S\ref{sec:6.2}).

Llama flips with depth. At 1/4 and 3/8 the fitted fill is worse than the identity fill. At half depth and 5/8 it is far better, +0.085 and +0.030 against +0.563 and +0.406, and lower in point estimate than POP's own configuration (+0.373). Section~\ref{sec:6.6} traces the shallow-split failure to the keys, and \S\ref{sec:6.7} reproduces the deep-split result on fresh documents.

\begin{table}[!htbp]
\centering
\caption{Split depth. Novel-token $\Delta$NLL against exact, nats per token. The interval is the paired RidgeFill $-$ identity contrast.}\label{tab:17}
\footnotesize
\setlength{\tabcolsep}{4pt}
\begin{tabular}{lllllll}
\toprule
Model & Depth & S & RidgeFill & Identity & No fill & \begin{tabular}[b]{@{}l@{}}95~\% CI, RidgeFill \\ $-$ identity\end{tabular} \\
\midrule
Llama-3.1-8B & 1/\allowbreak{}4 & 8 & +0.804 & +0.409 & +1.983 & [+0.128, +0.766] \\
 & 3/\allowbreak{}8 & 12 & +1.006 & +0.505 & +1.591 & [+0.307, +0.748] \\
 & 1/\allowbreak{}2 & 16 & +0.085 & +0.563 & +1.471 & [$-$0.548, $-$0.430] \\
 & 5/\allowbreak{}8 & 20 & +0.030 & +0.406 & +1.225 & [$-$0.422, $-$0.343] \\
 & POP native & 22 & n/\allowbreak{}a & +0.373 & n/\allowbreak{}a & n/\allowbreak{}a \\
Qwen3.5-9B & 1/\allowbreak{}4 & 8 & +0.033 & +0.032 & +0.056 & [$-$0.004, +0.005] \\
 & 3/\allowbreak{}8 & 12 & +0.024 & +0.023 & +0.046 & [$-$0.003, +0.005] \\
 & 1/\allowbreak{}2 & 16 & +0.017 & +0.015 & +0.045 & [0.000, +0.004] \\
 & 5/\allowbreak{}8 & 20 & +0.008 & +0.006 & +0.026 & [+0.001, +0.004] \\
 & POP native & 22 & n/\allowbreak{}a & +0.016 & n/\allowbreak{}a & n/\allowbreak{}a \\
Qwen3-8B & 1/\allowbreak{}4 & 9 & +0.042 & $-$0.004 & +0.655 & [+0.038, +0.054] \\
 & 3/\allowbreak{}8 & 13 & +0.044 & +0.026 & +0.243 & [+0.012, +0.025] \\
 & 1/\allowbreak{}2 & 18 & +0.030 & +0.033 & +0.122 & [$-$0.007, +0.001] \\
 & 5/\allowbreak{}8 & 22 & +0.023 & +0.019 & +0.155 & [+0.001, +0.008] \\
 & POP native & 24 & n/\allowbreak{}a & +0.044 & n/\allowbreak{}a & n/\allowbreak{}a \\
\bottomrule
\end{tabular}
\end{table}

\subsection{Arms at the primary split}\label{sec:6.2}
\begin{figure}[!htbp]
\centering
\includegraphics[width=\linewidth]{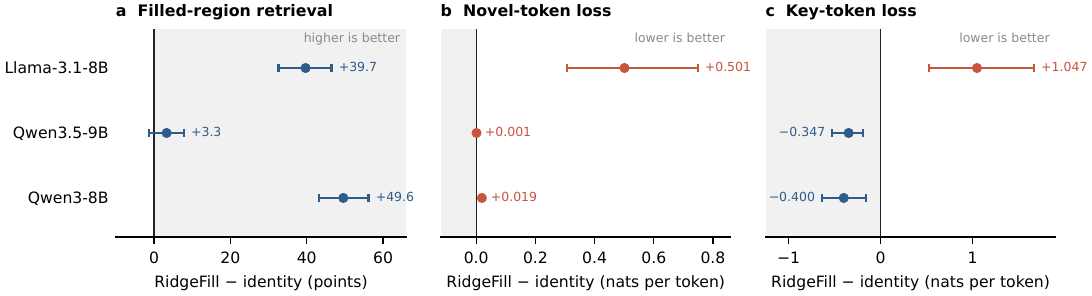}
\caption{RidgeFill $-$ identity at the primary split, with paired 95~\% bootstrap intervals. The shaded side favors RidgeFill: positive for retrieval and negative for loss. Blue points favor RidgeFill and red points favor the identity fill. RidgeFill improves retrieval strongly on the dense models, but at this aggressive Llama split it increases both novel-token and key-token loss. Key tokens are the scored tokens whose loss rises by more than 2 nats when the prefix is removed (\S\ref{sec:3.1}). These comparisons do not include exact prefix rows.}\label{fig:13}
\end{figure}

Table~\ref{tab:18} gives the paired contrasts between RidgeFill and the identity fill at 3/8 depth, the primary split, on every measure (Figure~\ref{fig:13}). Appendix~\ref{app:A} lists every arm on each model, Appendix~\ref{app:B} further paired contrasts, and Appendix~\ref{app:C} filled-region RULER by task and length.

\begin{table}[!htbp]
\centering
\caption{Paired RidgeFill $-$ identity contrasts at the primary split. Positive is better for retrieval, agreement and LongBench; negative is better for loss. RULER uses 216 samples at 8K to 64K on Llama, 219 at 8K to 64K on Qwen3.5 and 138 at 8K to 32K on Qwen3-8B. On Llama's 136 samples at 8K to 32K, the RULER contrast is +49.1 [+39.3, +57.5].}\label{tab:18}
\small
\setlength{\tabcolsep}{4pt}
\begin{tabularx}{\linewidth}{>{\raggedright\arraybackslash\hsize=1.000\hsize}Xlll}
\toprule
Metric & Llama-3.1-8B & Qwen3.5-9B & Qwen3-8B \\
\midrule
RULER filled (points) & +39.7 [+32.6, +46.5] & +3.3 [$-$1.3, +7.8] & +49.6 [+43.3, +56.2] \\
$\Delta$NLL novel (nats) & +0.501 [+0.307, +0.748] & +0.001 [$-$0.003, +0.005] & +0.019 [+0.012, +0.025] \\
$\Delta$NLL key (nats) & +1.047 [+0.521, +1.666] & $-$0.347 [$-$0.523, $-$0.196] & $-$0.400 [$-$0.634, $-$0.159] \\
Decode agreement (points) & +6.3 [+4.6, +8.5] & +0.5 [+0.1, +1.0] & +1.2 [+0.6, +2.0] \\
LongBench average & +2.4 [0.0, +4.7] & n/\allowbreak{}a & n/\allowbreak{}a \\
\bottomrule
\end{tabularx}
\end{table}

The fitted fill tracks exact's decoding more closely than the identity fill does on every model. KL is 0.124 against 0.334 on Llama, 0.012 against 0.021 on Qwen3.5 and 0.013 against 0.032 on Qwen3-8B, and greedy agreement rises on all three. The two Qwen models, which normalize their keys, also keep more of what the prefix contributes: under the fitted fill their key-token loss is 0.35 and 0.40 nats below the identity fill's. Retrieval moves most on the dense models, from 9.7 to 49.4 on Llama and from 44.0 to 93.7 on Qwen3-8B.

Llama pays for it at this split. Its novel-token loss is 0.50 nats and its key-token loss 1.05 nats above the identity fill's, the shallow-split failure of \S\ref{sec:6.1}.

Spending the fill's compute on a longer exact tail instead reaches RULER 13.2, 26.3 and 25.4 on the three models, at 8K to 32K. On novel-token loss, RidgeFill minus that control is $-$0.413 nats on Llama [$-$0.690, $-$0.075], +0.013 on Qwen3.5 and $-$0.086 on Qwen3-8B.

Llama's eight LongBench tasks agree in direction: RidgeFill averages 46.4 against the identity fill's 44.0, and the interval's lower bound is zero. The other arms average 49.7 (exact), 21.3 (no fill), 42.8 (POP native), 48.1 (identity + rows) and 49.1 (RidgeFill + rows).

\subsection{Against SwiftKV}\label{sec:6.3}
Released SwiftKV and RidgeFill have similar novel-token loss at SwiftKV's own split 16 on Llama-3.1-8B (Tables~\ref{tab:19} and~\ref{tab:20}): SwiftKV as released $-$ RidgeFill is +0.010 nats [$-$0.004, +0.025]. The released checkpoint runs with its own one-token tail and RidgeFill with its 4,096-token tail, so this compares each method as it ships. On key tokens the released checkpoint does worse, +0.229 nats, though its interval includes zero [$-$0.020, +0.485].

Given the same 4,096-token tail, SwiftKV's trained projections beat the ridge map on both measures: $-$0.024 nats [$-$0.037, $-$0.015] on novel tokens and $-$0.224 [$-$0.415, $-$0.057] on key tokens. The tail matters most on key tokens, where SwiftKV's loss falls from +0.568 to +0.115 once it gets our tail. SwiftKV's projections were distilled on about 680 million tokens at sequence lengths up to 8,192; the ridge map is a closed-form fit.

\begin{table}[!htbp]
\centering
\caption{SwiftKV comparison at split 16. $\Delta$NLL against exact, nats per token; key tokens over 80 documents.}\label{tab:19}
\small
\setlength{\tabcolsep}{4pt}
\begin{tabular}{lll}
\toprule
Arm (Llama-3.1-8B, split 16) & \begin{tabular}[b]{@{}l@{}}$\Delta$NLL \\ novel\end{tabular} & $\Delta$NLL key \\
\midrule
SwiftKV as released (one-token tail) & +0.095 & +0.568 \\
SwiftKV maps + exact tail & +0.061 & +0.115 \\
RidgeFill (tail 4,096) & +0.085 & +0.339 \\
\bottomrule
\end{tabular}
\end{table}

\begin{table}[!htbp]
\centering
\caption{Paired SwiftKV $-$ RidgeFill contrasts.}\label{tab:20}
\small
\setlength{\tabcolsep}{4pt}
\begin{tabularx}{\linewidth}{>{\raggedright\arraybackslash\hsize=1.000\hsize}Xll}
\toprule
Contrast & Novel tokens [95~\% CI] & Key tokens [95~\% CI] \\
\midrule
SwiftKV as released $-$ RidgeFill & +0.010 [$-$0.004, +0.025] & +0.229 [$-$0.020, +0.485] \\
SwiftKV maps + exact tail $-$ RidgeFill & $-$0.024 [$-$0.037, $-$0.015] & $-$0.224 [$-$0.415, $-$0.057] \\
\bottomrule
\end{tabularx}
\end{table}

\subsection{Against POP}\label{sec:6.4}
POP is one operating point: the identity fill from layer $L - \lfloor L/3 \rfloor$ with a one-token tail. We compare RidgeFill with it at two splits.

At 5/8 depth, two layers below POP's split on both Llama and Qwen3-8B, RidgeFill skips more layers than POP and is non-inferior to it on filled-region retrieval at a 3-point margin (Table~\ref{tab:21}). On both models the lower bound is the binomial floor of \S\ref{sec:3.4}. RidgeFill's novel-token loss is lower in point estimate, +0.030 against +0.373 on Llama and +0.023 against +0.044 on Qwen3-8B. The only interval for this contrast comes from Llama's 30 fresh documents: RidgeFill at 5/8 $-$ POP native is $-$0.268 nats [$-$0.336, $-$0.200] (\S\ref{sec:6.7}). These 5/8-depth comparisons were declared after the primary-split results, as follow-ups (\S\ref{sec:8}). At these lengths RidgeFill is not cheaper. With its 4,096-token tail and the maps, it costs more multiply-accumulates than POP across 8K to 32K on both models, and becomes cheaper only above about 46K tokens on Llama and 59K on Qwen3-8B (\S\ref{sec:2.5}, Figure~\ref{fig:7}).

The primary split gives a mixed picture. Qwen3-8B's split 13 matches POP's split 24 on novel-token loss, RidgeFill $-$ POP +0.001 nats [$-$0.008, +0.010], with 24~\% fewer multiply-accumulates at 32K, but it retrieves less, 93.7 against 98.7. On Qwen3.5, POP is better by 0.008 nats (RidgeFill $-$ POP [+0.003, +0.014]). On Llama it is better by 0.63 nats [+0.401, +0.921], until RidgeFill moves to half depth (\S\ref{sec:6.1}).

No needle-retrieval measurement of POP's configuration has been published, as far as we know; its paper reports LongBench QA. Table~\ref{tab:21} gives one: 94.1 on Llama, 5.7 points below exact.

\begin{table}[!htbp]
\centering
\caption{POP comparison on filled-region retrieval. On Llama, exact and POP native score 99.6 and 93.4 in Appendix~\ref{app:A} and 99.8 and 94.1 here. These are two runs of the same configurations on different hardware, about one sample apart in 136; each is paired with the contrasts in its own table.}\label{tab:21}
\small
\setlength{\tabcolsep}{4pt}
\begin{tabularx}{\linewidth}{>{\raggedright\arraybackslash\hsize=0.853\hsize}X>{\raggedright\arraybackslash\hsize=1.073\hsize}X>{\raggedright\arraybackslash\hsize=1.073\hsize}X}
\toprule
RULER filled, 8K--32K & Llama-3.1-8B (n = 136) & Qwen3-8B (n = 138) \\
\midrule
Exact & 99.8 & 98.7 \\
POP native & 94.1 (split 22) & 98.7 (split 24) \\
Identity at 5/\allowbreak{}8 & 95.0 & 98.9 \\
RidgeFill at 5/\allowbreak{}8 & 98.7 & 99.1 \\
RidgeFill at 5/\allowbreak{}8 $-$ POP native & +4.6; lower bound $-$2.18 (margin $-$3) & +0.4; lower bound $-$2.15 (margin $-$3) \\
RidgeFill $-$ identity at 5/\allowbreak{}8 & +3.7 [+2.0, +6.2] & +0.2 [0.0, +0.9] \\
\bottomrule
\end{tabularx}
\end{table}

\subsection{Exact prefix rows}\label{sec:6.5}
Attention-selected exact rows (\S\ref{sec:2.4}) restore filled-region retrieval on Llama and Qwen3.5, where RidgeFill plus rows is non-inferior to exact at a 2-point margin (Table~\ref{tab:22}). Against controls of the same size, random rows and a longer tail, the selection wins by 16 to 30 points on those two models, so the gain comes from which rows are chosen, not from the extra compute. Hiding the question from the selector costs 16 and 14 points, consistent with the bound of \S\ref{sec:2.4}: a row's influence is limited by the attention it receives from the queries actually asked.

Qwen3-8B is the exception. The rows lower its retrieval from 93.7 to 89.9, while random rows and a longer tail of the same size both reach 93.8. Random rows also mix exact and filled rows, so the mixing alone does not explain the drop. Against RidgeFill alone, the selected rows make 24 samples worse and 10 better; random rows make 8 worse and 10 better. Every loss is on a multi-needle task (multiquery, multivalue or multikey), mostly at 16K and 32K, and the single-needle task loses nothing. We have no tested explanation. One hypothesis fits the data: on Qwen3-8B, making only part of the cache exact also hurts in the splices, where exact values alone drop retrieval to 82.4 (\S\ref{sec:6.6}), and the selected rows fall on the needles, which random rows rarely do.

On Llama at 3/8 the rows also remove the loss failure of \S\ref{sec:6.1}. Novel-token $\Delta$NLL falls from +1.006 to +0.067, and on the worst document, 128K tokens of source code, from +8.189 to +0.180. Without the rows, RidgeFill at 3/8 is worse than no fill on 32 of the 36 documents of 64K tokens or longer and on none of the 90 shorter ones. Appendix~\ref{app:D} asks a second question of a cache filled once.

\begin{table}[!htbp]
\centering
\caption{Exact-row contrasts on filled-region retrieval. Where a lower bound is given, it is the one-sided bound against the 2-point margin; at these sample sizes it equals the binomial floor (\S\ref{sec:3.4}). The first and last rows use 8K to 64K on Llama and Qwen3.5 (216 and 219 samples). The controls ran only at 8K to 32K, so the middle three rows use 136 and 139 samples. Qwen3-8B uses its 138 samples at 8K to 32K throughout. On Qwen3-8B the random-row and longer-tail controls differ on 13 of the 138 samples but have the same total, so their contrasts coincide.}\label{tab:22}
\footnotesize
\setlength{\tabcolsep}{4pt}
\begin{tabularx}{\linewidth}{>{\raggedright\arraybackslash\hsize=1.197\hsize}X>{\raggedright\arraybackslash\hsize=0.952\hsize}X>{\raggedright\arraybackslash\hsize=0.925\hsize}X>{\raggedright\arraybackslash\hsize=0.925\hsize}X}
\toprule
RULER filled contrast & Llama-3.1-8B & Qwen3.5-9B & Qwen3-8B \\
\midrule
RidgeFill + rows $-$ exact & +0.2; lower bound $-$1.38 (n = 216) & 0.0; lower bound $-$1.36 (n = 219) & $-$8.9 [$-$13.6, $-$5.4] (n = 138) \\
RidgeFill + rows $-$ RidgeFill + random rows & +29.6 [+23.3, +36.9] & +18.3 [+13.1, +24.5] & $-$4.0 [$-$8.5, +0.5] \\
RidgeFill + rows $-$ RidgeFill + longer tail & +24.8 [+19.1, +31.6] & +16.5 [+11.5, +22.3] & $-$4.0 [$-$8.5, +0.5] \\
Question-blind rows $-$ rows & $-$16.0 [$-$21.5, $-$11.8] & $-$14.0 [$-$19.2, $-$9.5] & n/\allowbreak{}a \\
RidgeFill + rows $-$ identity + rows & +0.8 [$-$0.1, +2.9] & +0.8 [$-$0.1, +2.7] & +26.4 [+19.9, +32.6] \\
\bottomrule
\end{tabularx}
\end{table}

\subsection{Which component carries the error}\label{sec:6.6}
\begin{figure}[!htbp]
\centering
\includegraphics[width=\linewidth]{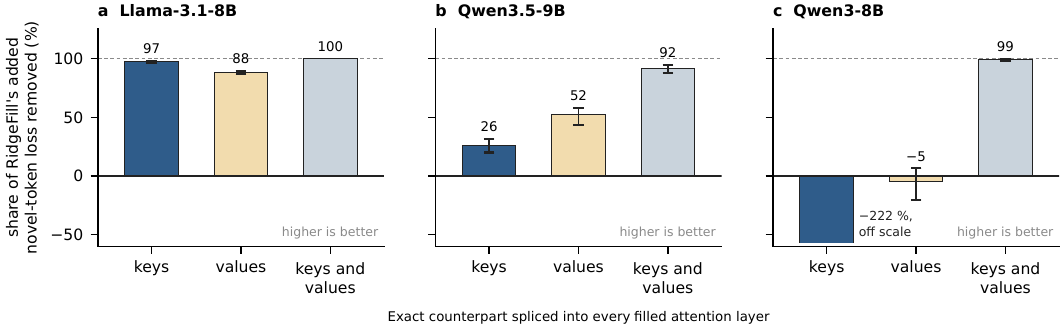}
\caption{Share of RidgeFill's added novel-token loss removed by each splice at the primary split, with 95~\% intervals where they exist. On Qwen3.5 the Gated DeltaNet layers stay filled in every splice. Qwen3-8B's exact-keys bar, $-$222~\%, is off the scale.}\label{fig:14}
\end{figure}

The splices replace the filled keys, the filled values, or both with their exact counterparts, in every filled attention layer at once (Table~\ref{tab:23}, Figure~\ref{fig:14}). Exact keys are inserted before the rotary embedding and rotated at their own positions. ``Loss removed'' is the share of RidgeFill's novel-token $\Delta$NLL that the splice takes away, and ``gap closed'' is the share of its filled-region RULER gap to exact. The key and value scales multiply the filled keys or values by a per-head ratio measured on the exact tail (\S\ref{sec:2.3}).

\begin{table}[!htbp]
\centering
\caption{Splices at the primary split. RULER at 8K to 32K: 136 samples on Llama and 138 on Qwen3-8B. On Qwen3-8B, exact and RidgeFill differ on only 20 samples, too few for an interval on the gap closed. On Qwen3.5 the splices act on the five upper attention layers; the fifteen upper Gated DeltaNet layers stay filled in every row, including ``exact keys and values''. Value-scale rows have no interval, and no value-scale run exists for Qwen3-8B.}\label{tab:23}
\footnotesize
\setlength{\tabcolsep}{4pt}
\begin{tabular}{llllll}
\toprule
Model & Splice & \begin{tabular}[b]{@{}l@{}}$\Delta$NLL \\ novel\end{tabular} & Loss removed [95~\% CI] & \begin{tabular}[b]{@{}l@{}}RULER \\ filled\end{tabular} & Gap closed \\
\midrule
Llama-3.1-8B & RidgeFill & +1.006 & n/\allowbreak{}a & 60.7 & n/\allowbreak{}a \\
 & + exact keys & +0.031 & 97.0~\% [96.0, 97.7] & 99.1 & 98.6~\% \\
 & + exact values & +0.120 & 88.1~\% [86.6, 89.5] & 79.6 & 48.6~\% \\
 & + exact keys and values & 0.000 & 100.0~\% & n/\allowbreak{}a & n/\allowbreak{}a \\
 & + key scale (tail) & +0.502 & 50.1~\% [45.5, 54.5] & n/\allowbreak{}a & n/\allowbreak{}a \\
 & + value scale (tail) & +1.103 & n/\allowbreak{}a & n/\allowbreak{}a & n/\allowbreak{}a \\
Qwen3.5-9B & RidgeFill & +0.024 & n/\allowbreak{}a & not run & not run \\
 & + exact keys & +0.018 & 25.8~\% [19.9, 31.6] & not run & not run \\
 & + exact values & +0.012 & 52.0~\% [43.4, 57.6] & not run & not run \\
 & + exact keys and values & +0.002 & 91.5~\% [87.8, 94.7] & not run & not run \\
 & + key scale (tail) & +0.024 & 0.4~\% [$-$0.9, 1.6] & not run & not run \\
 & + value scale (tail) & +0.020 & n/\allowbreak{}a & not run & not run \\
Qwen3-8B & RidgeFill & +0.044 & n/\allowbreak{}a & 93.7 & n/\allowbreak{}a \\
 & + exact keys & +0.143 & $-$222.4~\% [$-$271.4, $-$180.7] & 94.2 & 10.7~\% \\
 & + exact values & +0.046 & $-$4.7~\% [$-$20.3, 6.4] & 82.4 & $-$221.4~\% \\
 & + exact keys and values & +0.001 & 98.8~\% [97.9, 99.8] & n/\allowbreak{}a & n/\allowbreak{}a \\
 & + key scale (tail) & +0.039 & 11.9~\% [9.2, 17.3] & n/\allowbreak{}a & n/\allowbreak{}a \\
\bottomrule
\end{tabular}
\end{table}

Llama's error is in the keys. Exact keys remove 97~\% of the loss and close 99~\% of the retrieval gap, and rescaling the filled keys by a ratio measured on the tail removes half of it, while the same rescale of the values makes the loss worse. Exact values also remove most of the loss, 88~\%, but close only half of the retrieval gap.

Qwen3.5's values carry more of the error than its keys, 52~\% against 26~\%. Together they remove 91.5~\%, and the rest is the Gated DeltaNet state, which the splices leave filled.

Qwen3-8B's keys and values only work together. Exact keys alone triple the loss, while keys and values together remove 99~\% of it, so the filled keys and values are wrong together but consistent with each other.

Key normalization fits the key-scale pattern. With QK-norm, the norm fixes each key's size, so a key rescale has little left to correct: 0.4~\% on Qwen3.5 and 11.9~\% on Qwen3-8B. Without it, a key's size enters the attention score directly, and that is where Llama's error is. Shares from separate splices need not add up, because the components interact (Appendix~\ref{app:H}).

\subsection{Replication}\label{sec:6.7}
\begin{figure}[!htbp]
\centering
\includegraphics[width=\linewidth]{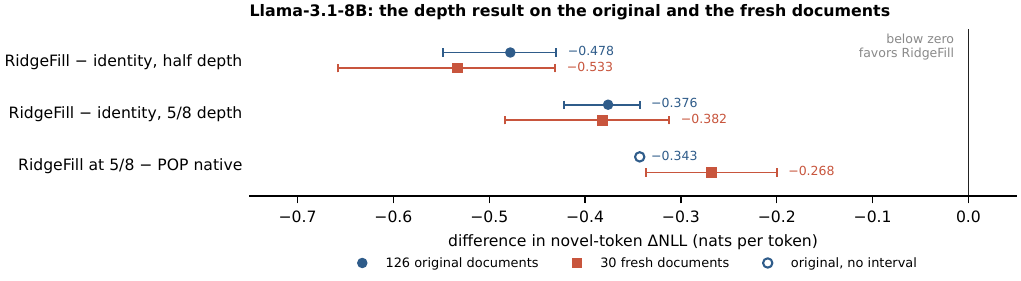}
\caption{Novel-token contrasts on Llama-3.1-8B, on the 126 original documents and on 30 fresh documents built after the splits were chosen. Below zero favors RidgeFill. No 126-document interval exists for RidgeFill at 5/8 $-$ POP native, so only its point estimate is shown.}\label{fig:15}
\end{figure}

Thirty fresh long documents reproduce the depth result on Llama-3.1-8B (Figure~\ref{fig:15}). They were built one per domain and length, from the same sources as the original 126 (papers through the arXiv-summarization corpus of Cohan et al. (2018), and source code through a GitHub code corpus in place of the four repositories), after all maps were fit and after the half and 5/8 splits had been chosen from the 126-document results. They share no run of 50 or more words with the 126 documents, RULER, LongBench or the calibration text. Two predictions were registered before any run on them: RidgeFill beats identity at half depth and at 5/8, and RidgeFill at 5/8 beats POP (\S\ref{sec:9}).

Both held. RidgeFill $-$ identity is $-$0.533 nats [$-$0.658, $-$0.431] at half depth and $-$0.382 [$-$0.484, $-$0.312] at 5/8, against $-$0.478 and $-$0.376 on the original documents. RidgeFill at 5/8 $-$ POP native is $-$0.268 [$-$0.336, $-$0.200]. On key tokens, only 49 tokens in 16 documents, the point estimates are $-$0.458 and $-$0.299, with intervals that include zero.

A follow-up checked retrieval at half depth on the same 136 filled-region samples at 8K to 32K. It was logged before any result but carried no registered prediction. RidgeFill $-$ identity is +12.5 points [+8.6, +17.5], 96.1 against 83.6.

\section{Related work}\label{sec:7}
Methods that skip late layers for prompt tokens differ in what happens to a skipped token's cache and in whether anything is trained (Table~\ref{tab:1}). This section takes the cells of that table in turn, then the closest neighbors outside it.

\textbf{Fill, pretrained in.} YOCO (Sun et al., 2024) and the causal encoder-decoder of DeepSeek-V4.1-Flash (DeepSeek-AI, 2026, Sec. 2.2) serve the upper layers' caches from a projection of a middle layer's state, and train the whole model with that structure in place. RidgeFill retrofits the same design onto models that were not trained for it.

\textbf{Fill, trained retrofit.} SwiftKV (Qiao et al., 2025) prefills later layers' KV caches from an earlier layer's output and recovers quality by distilling the skipped layers' Q/K/V projections from the original model, about 680 million tokens for Llama-3.1-8B. KV Prediction (Horton et al., 2024) runs a small auxiliary model over the prompt and maps its cache to the base model's with a trained predictor, keeping the base model frozen. It reports a 25 to 50~\% cut in prefill compute with little loss in accuracy, and quality falling off quickly beyond 50~\%. On the day DeepSeek-V4.1-Flash was released, kishida retrofitted its design onto Qwen3-8B as the Q3-8B-KVA-Projector (kishida, 2026), with trained per-layer key and value heads that read from layer 18.\footnote{The name KVA comes from kishida's projector and is kept in our engine code.} RidgeFill differs from all three in two ways: its fill is fitted in closed form and predicts each layer's input, so the layer's own projections still run, and it keeps an exact tail.

\textbf{Fill, identity.} POP (He et al., 2026) prunes the last third of the layers during prefill. Each pruned layer writes its keys and values from its own projections of the unchanged hidden state, and the last prompt token runs through the full model. POP reports up to 1.37$\times$ faster prefill at 2K tokens and a sharp decline past 50~\% pruning. Its long-context evaluation uses two LongBench QA tasks and no needle tests. At the same split and tail, POP's fill is the identity arm of \S\ref{sec:6}.

Both SwiftKV and POP report quality falling off once more than half the layers are skipped. RidgeFill's primary split skips about 5/8 of them. On the two Qwen models its novel-token loss there stays within 0.05 nats of exact, but on Llama it rises to +1.006, the same falling-off; at half depth Llama recovers to +0.085 (\S\ref{sec:6.1}).

\textbf{Drop, trained.} TokenGate (Patel et al., 2026) places a learned gate at an early layer, layer 1 in its main experiments, and keeps only the highest-scoring tokens. The layers above the gate neither process the pruned tokens nor write cache entries for them. TokenGate fine-tunes the backbone with LoRA jointly with the gate, separately for each task, and reports 1.4 to 11.3$\times$ faster time to first token and 1.5 to 3.4$\times$ lower peak memory on Qwen3-1.7B and Qwen3.5-2B. SPEED (Oh et al., 2026) keeps prompt tokens' cache entries only in the lower layers, apart from a start-of-sequence anchor, and runs decoding at full depth. On Llama-3.1-8B, instruction-tuned for it, with 75~\% of the layers for prompt tokens, it reports 33~\% faster time to first token at 128K and a benchmark average of 51.2 against 51.4. Dropping a token removes its cache entry, and the memory with it; filling keeps both. Neither method is compared at a matched budget: TokenGate trains per task, SPEED trains once, and both drop tokens.

\textbf{Drop, training-free.} FastKV (Jo et al., 2026) keeps the full context up to a chosen layer and propagates only the selected tokens past it. GemFilter (Shi et al., 2024) uses an early layer's attention to pick the tokens that matter and reruns the model on those alone. LazyLLM (Fu et al., 2024) prunes tokens progressively across layers and can bring a pruned token back at a later step. Speculative Prefill (Liu et al., 2025) belongs here too, despite its name: it uses a small model to pick which prompt tokens to keep and drops the rest. All four select tokens without training.

\textbf{Linear replacement of layers.} ReplaceMe (Shopkhoev et al., 2025) removes a run of transformer blocks and replaces them with one linear map estimated from a small calibration set, with no training. It is one of the closest training-free uses of a fitted linear map. ReplaceMe changes the model for every token, decode included, while RidgeFill leaves the weights and the decode path untouched and approximates only the prompt's late-layer caches. Heo et al. (2026) fit per-head ridge maps in closed form to transfer KV caches between models of one family, so that a larger model can skip its own prefill.

\textbf{Query-aware selection.} The attention-selected rows of \S\ref{sec:2.4} follow SnapKV (Li et al., 2024a), which scores prompt tokens by the attention they receive from an observation window at the end of the prompt. Quest (Tang et al., 2024) is query-aware in the same sense at decode time, choosing cache pages by their estimated attention to the current query. Those methods use the scores to evict or skip cache entries. Here the scores choose which prefix rows run exactly, and the question-blind control (\S\ref{sec:6.5}) measures how much of the selection's value depends on the question being in the window.

\textbf{Sparse attention and prompt compression.} MInference (Jiang et al., 2024) and FlexPrefill (Lai et al., 2025) reduce the attention computation itself, and LLMLingua-2 (Pan et al., 2024) shortens the prompt. Both approaches are orthogonal to skipping layers, and Flash-Next (\S\ref{sec:5}) already runs RidgeFill on top of its native sparse attention.

\textbf{Speculative decoding.} Speculative decoding (Leviathan et al., 2023; Chen et al., 2023) speeds up decoding by having a cheap draft propose tokens that the full model then verifies in one pass. Medusa (Cai et al., 2024) and EAGLE (Li et al., 2024b) draft with extra heads on the model itself, and LayerSkip (Elhoushi et al., 2024) drafts with the model's own early layers. DeepSeek-V3 (DeepSeek-AI, 2024) trains a multi-token-prediction module that can serve as a drafter, and Flash-Next ships one. Section~\ref{sec:1} frames RidgeFill as the same trade run in the other direction (Figure~\ref{fig:1}).

\section{Limitations}\label{sec:8}
\textbf{Shallow splits on a model without QK-norm.} QK-norm normalizes each query and key before attention, and Llama-3.1-8B has none. On Llama at the primary split, 3/8 depth, RidgeFill's novel-token loss is 0.50 nats above the identity fill's. On long documents it is worse than no fill at all: on 32 of the 36 documents of 64K tokens or longer, and on none of the 90 shorter ones. The worst is 128K tokens of source code, where RidgeFill's $\Delta$NLL against exact is +8.19 nats and no fill's is +0.74. Its filled-region retrieval is still 39.7 points above the identity fill's, 49.4 against 9.7. Splicing exact keys and values into the filled cache puts the error in the keys, and half of it in their scale (\S\ref{sec:6.6}). At half depth the order reverses: the fitted fill's loss is 0.478 nats below the identity fill's on the 126 documents and 0.533 below on 30 fresh ones (\S\ref{sec:6.7}). On a model without QK-norm we recommend a split at half depth or deeper, or attention-selected rows, which bring the source-code document down to +0.180 (\S\ref{sec:6.5}). Those rows depend on the question, which has its own cost (below).

\textbf{Key tokens.} Key tokens are the ones whose loss rises by more than 2 nats when the context is cut to its last 4,096 tokens (\S\ref{sec:3.1}), so they measure what the prefix contributes. On them, RidgeFill alone at the primary split is 1.2 to 3.3 nats above exact across the three reference models, and attention-selected rows bring that down to 0.4 to 0.7. A needle score does not show this cost. With the rows, Llama and Qwen3.5 retrieve 99.5 and 99.9 from the filled region, level with exact, and their key-token loss is still 0.41 and 0.43 nats above it. Check key-token loss, and not only retrieval, before choosing a split.

\textbf{Row selection.} The paper uses three row selectors, one per deployment, each tuned on its own development data, and we do not claim that any of them is the right selector for a new model. Two of the three depend on the question. Attention-selected rows lose 16 and 14 points of filled-region retrieval on Llama and Qwen3.5 when the question is hidden from the selector (\S\ref{sec:6.5}). The 27B's probe ranks rows by the attention they receive from the last 256 prompt rows, which normally hold the question, and it was never run with the question hidden. On Qwen3-8B, attention-selected rows lower retrieval from 93.7 to 89.9 while random rows of the same count reach 93.8, and we have not established why.

\textbf{Caches reused across questions.} A cache filled once and then asked several questions should not rely on a selector that reads the first question. Appendix~\ref{app:D} asks each reference model a second question of a cache filled, and its rows chosen, for the first. With the rows, Llama and Qwen3.5 still retrieve every second needle in the filled region, but the test is weak for this purpose. The second question asks for another key from the same template, and the rows chosen for the first already cover a median of 85~\% and 86~\% of the second needle's tokens. It does not show that rows serve an unrelated question. For such a cache, the option the evidence supports is a deeper split with no rows: on Llama at half depth, RidgeFill alone retrieves 96.1 at 8K to 32K (\S\ref{sec:6.7}). A selector that ignores the question, such as the protected tokens of Flash-Next's quality mode, is the other candidate, but quality mode's paired difference from plain RidgeFill, $-$1.02~\% [$-$2.87~\%, +0.80~\%], does not establish a gain (\S\ref{sec:5.3}).

\textbf{Calibration length.} The longest calibration document, which every set of maps uses, is about 30K tokens (30,465 tokens on the Qwen3.5-9B and 27B tokenizers, 29,009 on Qwen3-8B's and 28,352 on Llama's), so no map was fitted on a position much beyond that. The median calibration document is 11.2K to 11.6K tokens, depending on the tokenizer (Appendix~\ref{app:G}). The test documents run to 128K. On Llama at the primary split, RidgeFill's filled-region retrieval falls from 79.8 at 8K to 30.3 at 64K. We have not tested whether longer calibration documents would close that gap.

\textbf{The 27B.} No identity-fill arm was run on this model, so its results compare RidgeFill with stock only. Of the RULER tasks, common-word extraction drops the most, by 10.0 points at 16K and 13.5 at 32K on 20 items each, with intervals that include zero; no other task moves by more than 5 points (\S\ref{sec:4.5}). RULER and the agreement study stop at 32K. Above that, the only quality evidence is LongBench v2's long group of 108 items, where RidgeFill is 0.9 points below stock with an interval from $-$7.4 to +6.5 (\S\ref{sec:4.6}), too wide to rule out a loss of several points. At 32K, on the 20 documents where the two arms disagree most, ten in each direction, and five near the median, the full-precision BF16 model is closer to stock than to RidgeFill on 22 of the 25 (\S\ref{sec:4.7}). RidgeFill's much lower loss on copy-heavy documents is therefore a departure from the model, not a gain. The probe's row count, inherited from the base rules, and its 256-row window were set on development data and not swept. The value rescale was not measured on its own in the final runs. The 27B's calibration text shares license headers, and nothing else, with 12 of LongBench v2's Code Repository items: runs of 78 to 85 words with documents its maps were fitted on, and up to 162 words with documents only its token-frequency table counts (Appendix~\ref{app:G}).

\textbf{Flash-Next.} The perplexity interval runs from $-$0.02~\% to +4.93~\% (\S\ref{sec:5.3}). It does not separate the 2.65~\% rise from zero, and it allows a rise of nearly 5~\%. The plain RidgeFill server start did not check that switching the fill off reproduces exact. The quality-mode start did, bitwise on 6 of 6 documents, and both starts used the same container image, replaced file and arguments (\S\ref{sec:5.2}). About two thirds of the routed experts are held in host memory and read over PCIe during prefill, so this deployment cannot separate the compute the fill saves from the PCIe traffic it saves (\S\ref{sec:5.3}). The fork's image is public, but its patches are not (Appendix~\ref{app:F}).

\textbf{Speed, memory and compute.} Every prefix row still gets a cache entry, so the fill saves no cache memory, and the maps add VRAM. On the 27B they add 1.6 to 2.9 GiB when held in GPU memory and 0.3 to 1.7 GiB when streamed from host memory (\S\ref{sec:4.4}). Methods that drop tokens do save that memory. TokenGate, for example, reports 1.4 to 11.3$\times$ faster time to first token and 1.5 to 3.4$\times$ lower peak memory on Qwen3-1.7B and Qwen3.5-2B (\S\ref{sec:7}). Our speedups are 1.19 to 1.31$\times$ on the 27B, across resident and streamed maps, and 1.50 to 1.60$\times$ on Flash-Next. We do not compare with any token-dropping method at a matched budget. On both serving models the measured speedup is below what the skipped compute allows, and we have not profiled why (\S\ref{sec:4.4}, \S\ref{sec:5.3}). In the reference study, RidgeFill at 5/8 depth costs more multiply-accumulates than POP below about 46K tokens on Llama and 59K on Qwen3-8B, because of its tail and maps, so it costs more across the whole 8K to 32K needle range (\S\ref{sec:2.5}, \S\ref{sec:6.4}).

\textbf{Follow-up analyses.} The 5/8-depth comparisons with POP (\S\ref{sec:6.4}) were chosen after the primary-split results were in and registered only before their own runs, so they are follow-ups. The 30 fresh documents (\S\ref{sec:6.7}) were registered to test their loss results on new text, and they confirmed them; the retrieval results have no such test. The half-depth retrieval check of \S\ref{sec:6.7} was registered before it ran but stated no prediction.

\textbf{Scope.} The evidence comes from three reference models of 8 to 9B parameters and two serving models. Both serving models are hybrids of Gated DeltaNet and attention, both run quantized weights, and they are the only models whose speed was measured; the reference study reports multiply-accumulate counts only. Qwen3-8B runs to 32K tokens only, since it needs YaRN beyond that. Decode agreement on the reference models uses 20 documents, and LongBench v1 runs on Llama only. Flash-Next's 64K RULER is one task, niah\_single\_1, on 25 samples, and its time-to-first-token medians come from prefixes of one document. The ceiling r of \S\ref{sec:2.5} assumes every layer costs the same, which neither hybrid meets, so for speed the measured numbers are the ones to use.

\section{Reproducibility and recipe}\label{sec:9}
\textbf{Code, maps and data.} The llama.cpp fork is public at \url{https://github.com/Approximate-Science/llama.cpp-ridgefill}, and the 27B runs used commit b0439f7bb on its main branch. The maps are on Hugging Face under the organization TheCacheIsALie: \texttt{RidgeFill-\allowbreak{}Projector-\allowbreak{}Qwen3.\allowbreak{}8-\allowbreak{}27B-\allowbreak{}i8} (revision 69e3ea99), \texttt{RidgeFill-\allowbreak{}Projector-\allowbreak{}Qwen3.\allowbreak{}8-\allowbreak{}Flash-\allowbreak{}Next-\allowbreak{}i8} (revision adb763a7), \texttt{Llama-\allowbreak{}3.\allowbreak{}1-\allowbreak{}8B-\allowbreak{}Instruct-\allowbreak{}RidgeFill-\allowbreak{}Projector} (revision d2ea5faa), \texttt{RidgeFill-\allowbreak{}Projector-\allowbreak{}Qwen3-\allowbreak{}8B} (revision 10b9fef7) and \texttt{RidgeFill-\allowbreak{}Projector-\allowbreak{}Qwen3.\allowbreak{}5-\allowbreak{}9B} (revision 03363301). The technical note that preceded this paper (Johnston and Smith, 2026) reports Flash-Next numbers measured in a different engine under a different protocol; this paper's measurements supersede them.

Everything else is in one data repository, \url{https://github.com/Approximate-Science/ridgefill-data} (tag v1.8, commit f3b343fe), kept separate from the map repositories and the engine forks. It holds the reference implementation and fit scripts, the Flash-Next run kit, the determinism fix as a diff with the hashes of both files, the per-sample RULER and per-document loss results for every arm of every deployment, the 27B and Flash-Next run records, the scripts that recompute every interval in this paper, the run ledger's chain index and the entries the results rest on (see Registration below), and the Lean project. The three reference-model repositories hold the maps at all four splits of \S\ref{sec:6.1} as safetensors, and the data repository lists every map file by its sha256. In the data repository, code is released under Apache-2.0 and results under CC BY 4.0. The calibration and evaluation texts come from sources with different licenses, so we release their hashes, not the texts.

\textbf{Pairing and determinism.} In the reference study, each arm is paired with the exact run of the same session; the rows arms' per-document differences come from a second loss run, paired with that run's own exact. On the 27B, the fork with the fill switched off is byte-identical to upstream (\S\ref{sec:4.3}). On Flash-Next, exact's two perplexity passes agree bitwise on all 90 documents, and with the fill switched off the quality-mode start reproduces exact bitwise on 6 of 6 documents (\S\ref{sec:5.2}). Both Flash-Next checks ran with the determinism fix in place.

\textbf{Registration.} The reference-study and Flash-Next runs were declared in the project's run ledger, a hash-chained log, before they ran. Each entry records the claim, the predictions with their decision rules, and the hashes of the run's configuration files. Each entry also carries the hash of the entry before it, so no entry can be edited or reordered without breaking the chain. No external registry holds the ledger. The data repository publishes its chain index, which holds the hash of every entry, and the full text of 27 of the 42 entries the results rest on. The other 15 are published as hashes only. The 27B runs of \S\ref{sec:4} are not in the published ledger: their development runs were logged on a separate branch of it, and the final runs on the A100 have no entry. Their builds, model, maps and inputs are fixed by the hashes in Appendix~\ref{app:F} and the data repository. The fresh-document confirmation (\S\ref{sec:6.7}) was registered after the set was frozen and before any run on it. The half-depth retrieval check was registered before any result existed but stated no prediction. The 5/8-depth comparisons with POP were registered before their own runs but chosen after the primary-split results, so we treat them as follow-ups (\S\ref{sec:8}).

\textbf{Applying RidgeFill to a new model.} These settings come from five models, and a new architecture or workload may need others.

\begin{enumerate}[leftmargin=*,itemsep=2pt,topsep=4pt]
\item Fit the maps. Run the unmodified model over calibration text, record the hidden state at the split and the input to every later layer, and solve one ridge regression per layer (\S\ref{sec:2.2}), with the ridge penalty's setting $\lambda$ chosen on held-out documents (Appendix~\ref{app:G}). One capture pass serves only the splits chosen before it runs: a new split needs a new pass, and a new $\lambda$ needs only new solves.
\item Choose the split by whether the model normalizes its queries and keys (QK-norm). On the two reference models with QK-norm, novel-token loss stays within 0.05 nats of exact at 3/8 depth. Llama-3.1-8B, which has no QK-norm, needs half depth or deeper (\S\ref{sec:6.1}). Attention-selected exact rows also remove Llama's shallow-split failure, but they lowered retrieval on Qwen3-8B (\S\ref{sec:6.5}) and depend on the question (\S\ref{sec:8}). For comparison, the 27B runs at 1/4 depth with 28 to 31~\% of the prefix exact, and Flash-Next at half depth with no exact prefix rows.
\item Use an exact tail of 2,048 to 4,096 tokens, the range of the final runs. Shorter tails were tried only in exploratory work. The tail runs every layer, but it reads the filled prefix cache, so its outputs can still differ from exact.
\item Check that the engine with the fill switched off reproduces exact bitwise, so that every difference you measure is the fill's.
\item Before shipping, measure two numbers against exact: filled-region needle retrieval and key-token $\Delta$NLL (\S\ref{sec:3.1}). Novel-token loss alone can hide the difference: on Qwen3-8B the fitted and identity fills differ by 0.019 nats on it and by 49.6 points on retrieval (\S\ref{sec:3}). A needle score near 100 can hide key-token loss (\S\ref{sec:8}). If you add exact rows, compare them with the same number of random rows, and with the question hidden if the cache will serve more than one question.
\end{enumerate}

Appendix~\ref{app:F} gives the builds and file hashes, Appendix~\ref{app:G} the calibration text and its overlap with the test sets, and Appendix~\ref{app:H} the exact-rows cost model, the lexical-net rules, the derivation of the $\rho$ scale and the formal statements.

\FloatBarrier
\section*{Acknowledgements}
kishida's Q3-8B-KVA-Projector, a retrofit of DeepSeek-V4.1-Flash's design posted the day that model came out, is where this work started and the source of the name KVA. StillDeadcode and the other contributors to our vLLM fork wrote the serving engine and its libr4d kernels, on top of the vLLM project (Kwon et al., 2023).

\FloatBarrier
\section*{Use of AI tools}
Most of the code, the experiment runs, the analysis scripts and the first drafts of the text were produced with AI assistants working under the authors' direction: Claude, GPT-based coding agents (Codex), open-weight models, and Gemini for reviews. The authors designed the method and the experiments, reviewed the code and the results, and take full responsibility for the contents of this paper. No AI tool is an author. References were checked against their arXiv, DOI or publisher records, and the data repository's recompute scripts regenerate the reported results from the released data and check 1,381 values against them.

\FloatBarrier
\section*{References}
\begin{list}{}{\setlength{\leftmargin}{1.5em}\setlength{\itemindent}{-1.5em}\setlength{\itemsep}{1pt}\setlength{\parsep}{0pt}}\small\raggedright
\item Bai, Y. et al. (2024). LongBench: A Bilingual, Multitask Benchmark for Long Context Understanding. ACL 2024, pp. 3119--3137. \href{https://arxiv.org/abs/2308.14508}{arXiv:2308.14508}.
\item Bai, Y. et al. (2025). LongBench v2: Towards Deeper Understanding and Reasoning on Realistic Long-context Multitasks. ACL 2025, pp. 3639--3664. \href{https://arxiv.org/abs/2412.15204}{arXiv:2412.15204}.
\item Cai, T. et al. (2024). Medusa: Simple LLM Inference Acceleration Framework with Multiple Decoding Heads. ICML 2024, PMLR 235, pp. 5209--5235. \href{https://arxiv.org/abs/2401.10774}{arXiv:2401.10774}.
\item Chalkidis, I. et al. (2022). LexGLUE: A Benchmark Dataset for Legal Language Understanding in English. ACL 2022, pp. 4310--4330. \href{https://arxiv.org/abs/2110.00976}{arXiv:2110.00976}.
\item Chen, C., Borgeaud, S., Irving, G., Lespiau, J.-B., Sifre, L., Jumper, J. (2023). Accelerating Large Language Model Decoding with Speculative Sampling. \href{https://arxiv.org/abs/2302.01318}{arXiv:2302.01318}.
\item Cohan, A. et al. (2018). A Discourse-Aware Attention Model for Abstractive Summarization of Long Documents. NAACL 2018, pp. 615--621. \href{https://arxiv.org/abs/1804.05685}{arXiv:1804.05685}.
\item DeepSeek-AI (2024). DeepSeek-V3 Technical Report. \href{https://arxiv.org/abs/2412.19437}{arXiv:2412.19437}.
\item DeepSeek-AI (2026). DeepSeek-V4.1-Flash: Pushing the Limits of KV Cache Compression. \href{https://arxiv.org/abs/2609.19969}{arXiv:2609.19969}.
\item Ding, N. et al. (2023). Enhancing Chat Language Models by Scaling High-quality Instructional Conversations. EMNLP 2023, pp. 3029--3051. \href{https://arxiv.org/abs/2305.14233}{arXiv:2305.14233}.
\item Efron, B. (1987). Better Bootstrap Confidence Intervals. Journal of the American Statistical Association 82(397), 171--185. doi:10.1080/01621459.1987.10478410.
\item Elhoushi, M. et al. (2024). LayerSkip: Enabling Early Exit Inference and Self-Speculative Decoding. ACL 2024, pp. 12622--12642. \href{https://arxiv.org/abs/2404.16710}{arXiv:2404.16710}.
\item Frantar, E., Ashkboos, S., Hoefler, T., Alistarh, D. (2023). GPTQ: Accurate Post-Training Quantization for Generative Pre-trained Transformers. ICLR 2023. \href{https://arxiv.org/abs/2210.17323}{arXiv:2210.17323}.
\item Fu, Q., Cho, M., Merth, T., Mehta, S., Rastegari, M., Najibi, M. (2024). LazyLLM: Dynamic Token Pruning for Efficient Long Context LLM Inference. \href{https://arxiv.org/abs/2407.14057}{arXiv:2407.14057}.
\item Grattafiori, A. et al. (2024). The Llama 3 Herd of Models. \href{https://arxiv.org/abs/2407.21783}{arXiv:2407.21783}.
\item He, J., Fu, Z., Wang, J., Li, Q. (2026). POP: Prefill-Only Pruning for Efficient Large Model Inference. Findings of ACL 2026, pp. 5864--5877. \href{https://arxiv.org/abs/2602.03295}{arXiv:2602.03295}.
\item Heo, T. et al. (2026). Cross-Model KV Cache Transfer in LLM Families: A Closed-Form Linear Mapping for Prefill Reuse. \href{https://arxiv.org/abs/2608.03893}{arXiv:2608.03893}.
\item Horton, M. et al. (2024). KV Prediction for Improved Time to First Token. \href{https://arxiv.org/abs/2410.08391}{arXiv:2410.08391}.
\item Hsieh, C.-P. et al. (2024). RULER: What's the Real Context Size of Your Long-Context Language Models? COLM 2024. \href{https://arxiv.org/abs/2404.06654}{arXiv:2404.06654}.
\item Jiang, H. et al. (2024). MInference 1.0: Accelerating Pre-filling for Long-Context LLMs via Dynamic Sparse Attention. NeurIPS 2024. \href{https://arxiv.org/abs/2407.02490}{arXiv:2407.02490}.
\item Jo, D., Song, J., Kim, Y., Kim, J.-J. (2026). FastKV: Decoupling of Context Reduction and KV Cache Compression for Prefill-Decoding Acceleration. Findings of ACL 2026, pp. 32167--32186. \href{https://arxiv.org/abs/2502.01068}{arXiv:2502.01068}.
\item Johnston, D., Smith, R. P. (2026). RidgeFill: a training-free late-layer cache fill for faster long-prompt prefill (technical note), v1.0. Zenodo, doi:10.5281/zenodo.23179168.
\item kishida (2026). Q3-8B-KVA-Projector: Late Layer KV Approximation Projector for Qwen3-8B. Hugging Face model repository, \url{https://huggingface.co/kishida/Q3-8B-KVA-Projector}, revision e5a2f56f.
\item Kwon, W. et al. (2023). Efficient Memory Management for Large Language Model Serving with PagedAttention. SOSP 2023, pp. 611--626. \href{https://arxiv.org/abs/2309.06180}{arXiv:2309.06180}.
\item Lai, X., Lu, J., Luo, Y., Ma, Y., Zhou, X. (2025). FlexPrefill: A Context-Aware Sparse Attention Mechanism for Efficient Long-Sequence Inference. ICLR 2025. \href{https://arxiv.org/abs/2502.20766}{arXiv:2502.20766}.
\item Leviathan, Y., Kalman, M., Matias, Y. (2023). Fast Inference from Transformers via Speculative Decoding. ICML 2023. \href{https://arxiv.org/abs/2211.17192}{arXiv:2211.17192}.
\item Li, Y. et al. (2024a). SnapKV: LLM Knows What You are Looking for Before Generation. NeurIPS 2024. \href{https://arxiv.org/abs/2404.14469}{arXiv:2404.14469}.
\item Li, Y., Wei, F., Zhang, C., Zhang, H. (2024b). EAGLE: Speculative Sampling Requires Rethinking Feature Uncertainty. ICML 2024, PMLR 235, pp. 28935--28948. \href{https://arxiv.org/abs/2401.15077}{arXiv:2401.15077}.
\item Liu, J., Chen, B., Zhang, C. (2025). Speculative Prefill: Turbocharging TTFT with Lightweight and Training-Free Token Importance Estimation. ICML 2025. \href{https://arxiv.org/abs/2502.02789}{arXiv:2502.02789}.
\item The mathlib Community (2020). The Lean Mathematical Library. CPP 2020, pp. 367--381. doi:10.1145/3372885.3373824.
\item de Moura, L., Ullrich, S. (2021). The Lean 4 Theorem Prover and Programming Language. CADE-28, LNCS 12699, pp. 625--635. doi:10.1007/978-3-030-79876-5\_37.
\item Oh, J., Jeon, H., Ji, H., Kong, K., Lee, J.-Y. (2026). Shallow Prefill, Deep Decoding: Efficient Long-Context Inference via Layer-Asymmetric KV Visibility. \href{https://arxiv.org/abs/2605.06105}{arXiv:2605.06105}.
\item Pan, Z. et al. (2024). LLMLingua-2: Data Distillation for Efficient and Faithful Task-Agnostic Prompt Compression. Findings of ACL 2024, pp. 963--981. \href{https://arxiv.org/abs/2403.12968}{arXiv:2403.12968}.
\item Patel, L., Ranjan, R., Stoica, I., Zaharia, M., Guestrin, C. (2026). TokenGate: Adapting LLMs for Fast and Memory-Efficient Prefill with Learned Token Pruning. \href{https://arxiv.org/abs/2610.26754}{arXiv:2610.26754}.
\item Peng, B., Quesnelle, J., Fan, H., Shippole, E. (2024). YaRN: Efficient Context Window Extension of Large Language Models. ICLR 2024. \href{https://arxiv.org/abs/2309.00071}{arXiv:2309.00071}.
\item Qiao, A., Yao, Z., Rajbhandari, S., He, Y. (2025). SwiftKV: Fast Prefill-Optimized Inference with Knowledge-Preserving Model Transformation. EMNLP 2025, pp. 25734--25753. \href{https://arxiv.org/abs/2410.03960}{arXiv:2410.03960}.
\item Qwen Team (2026a). Qwen3.5: Towards Native Multimodal Agents. Blog post, February 2026, \url{https://qwen.ai/blog?id=qwen3.5}. Model: Qwen3.5-9B, \url{https://huggingface.co/Qwen/Qwen3.5-9B}, revision c20223623576.
\item Qwen Team (2026b). Qwen3.8-Flash-Next: A New Architecture, Towards Ultimate Cost-Efficiency. Blog post, August 2026, \url{https://qwen.ai/blog?id=qwen3.8-flash-next}. Model: \url{https://huggingface.co/Qwen/Qwen3.8-Flash-Next}.
\item Qwen Team (2026c). Qwen3.8-Max: A New Bar for Coding and Cowork. Blog post, August 2026, \url{https://qwen.ai/blog?id=qwen3.8}. Model: Qwen3.8-27B, \url{https://huggingface.co/Qwen/Qwen3.8-27B}.
\item Rae, J. W., Potapenko, A., Jayakumar, S. M., Lillicrap, T. P. (2020). Compressive Transformers for Long-Range Sequence Modelling. ICLR 2020. \href{https://arxiv.org/abs/1911.05507}{arXiv:1911.05507}.
\item Shi, Z., Ming, Y., Nguyen, X.-P., Liang, Y., Joty, S. (2024). Discovering the Gems in Early Layers: Accelerating Long-Context LLMs with 1000x Input Token Reduction. \href{https://arxiv.org/abs/2409.17422}{arXiv:2409.17422}.
\item Shopkhoev, D. et al. (2025). ReplaceMe: Network Simplification via Depth Pruning and Transformer Block Linearization. NeurIPS 2025. \href{https://arxiv.org/abs/2505.02819}{arXiv:2505.02819}.
\item Smith, R. P. (2026a). Qwen3.8-Flash-Next-MXFP4-FP8-GPTQ. Hugging Face model repository, \url{https://huggingface.co/tcclaviger/Qwen3.8-Flash-Next-MXFP4-FP8-GPTQ}, revision 6fe3e746.
\item Smith, R. P. (2026b). vllm, tag 29.08.10: the vLLM fork's engine image (vLLM 0.29.0.dev0 at upstream commit 2bdbbc80 plus the fork's patches). Docker Hub image, \url{https://hub.docker.com/r/tcclaviger/vllm}, digest sha256:809657ddb1533db10a0c08267c3ee6fde180b839e379713414efa121415914a3.
\item Snowflake (2024). Llama-3.1-SwiftKV-8B-Instruct. Hugging Face model repository, \url{https://huggingface.co/Snowflake/Llama-3.1-SwiftKV-8B-Instruct}, revision 22a50b8c.
\item Sun, Y. et al. (2024). You Only Cache Once: Decoder-Decoder Architectures for Language Models. NeurIPS 2024. \href{https://arxiv.org/abs/2405.05254}{arXiv:2405.05254}.
\item Tang, J., Zhao, Y., Zhu, K., Xiao, G., Kasikci, B., Han, S. (2024). Quest: Query-Aware Sparsity for Efficient Long-Context LLM Inference. ICML 2024, PMLR 235, pp. 47901--47911. \href{https://arxiv.org/abs/2406.10774}{arXiv:2406.10774}.
\item unsloth (2026). Qwen3.8-27B-GGUF. Hugging Face model repository, \url{https://huggingface.co/unsloth/Qwen3.8-27B-GGUF}, revision 4ca72078.
\item Yang, A. et al. (2025a). Qwen3 Technical Report. \href{https://arxiv.org/abs/2505.09388}{arXiv:2505.09388}.
\item Yang, S., Kautz, J., Hatamizadeh, A. (2025b). Gated Delta Networks: Improving Mamba2 with Delta Rule. ICLR 2025. \href{https://arxiv.org/abs/2412.06464}{arXiv:2412.06464}.
\item Zhu, D. et al. (2025). Hyper-Connections. ICLR 2025. \href{https://arxiv.org/abs/2409.19606}{arXiv:2409.19606}.
\end{list}

\clearpage
\appendix
\counterwithin{table}{section}
\counterwithin{figure}{section}

\section{All arms at the primary split}\label{app:A}
Every arm uses the primary split and a 4,096-token tail, except POP native (its own split, one-token tail). RULER filled is filled-region accuracy (\%); $\Delta$NLL is against exact in nats per token; decode agreement and KL are against exact's greedy continuation (\S\ref{sec:3.1}); Turn 2 is the second question of Appendix~\ref{app:D}. The controls, POP native and no fill ran RULER only at 8K to 32K, so each table gives that column for every arm and the 8K to 64K column where it exists. The per-document $\Delta$NLL table (342 model--document pairs on the evaluation documents and 30 on Llama-3.1-8B's fresh documents, with every arm and depth that ran) is \texttt{derived/\allowbreak{}per-\allowbreak{}document-\allowbreak{}nll.\allowbreak{}csv} in the data repository. Its key is model, set and document, because 25 of the fresh documents reuse evaluation-document names.

\subsection{Llama-3.1-8B}\label{sec:A.1}
\begin{table}[H]
\centering
\caption{Primary split, Llama-3.1-8B. RidgeFill scores 79.8 at 8K alone and 30.3 at 64K alone. Identity + rows has a LongBench average of 48.1.}\label{tab:A.1}
\scriptsize
\setlength{\tabcolsep}{4pt}
\begin{tabular}{llllllll}
\toprule
Arm & \begin{tabular}[b]{@{}l@{}}RULER \\ filled, \\ 8K--32K (n \\ = 136)\end{tabular} & \begin{tabular}[b]{@{}l@{}}RULER \\ filled, \\ 8K--64K (n \\ = 216)\end{tabular} & \begin{tabular}[b]{@{}l@{}}$\Delta$NLL \\ novel\end{tabular} & $\Delta$NLL key & \begin{tabular}[b]{@{}l@{}}Decode \\ agree\end{tabular} & Decode KL & Turn 2 \\
\midrule
Exact & 99.6 & 99.3 & 0.000 & 0.000 & n/\allowbreak{}a & n/\allowbreak{}a & 100.0 \\
No fill & 0.0 & n/\allowbreak{}a & +1.591 & +5.401 & 71.5~\% & 1.262 & n/\allowbreak{}a \\
Identity & 11.6 & 9.7 & +0.505 & +2.223 & 85.1~\% & 0.334 & n/\allowbreak{}a \\
POP native & 93.4 & n/\allowbreak{}a & +0.373 & +0.275 & n/\allowbreak{}a & n/\allowbreak{}a & n/\allowbreak{}a \\
RidgeFill & 60.7 & 49.4 & +1.006 & +3.271 & 91.4~\% & 0.124 & 57.0 \\
RidgeFill + rows & 99.6 & 99.5 & +0.067 & +0.409 & n/\allowbreak{}a & n/\allowbreak{}a & 100.0 \\
Identity + rows & 98.3 & 98.7 & +0.268 & +1.100 & n/\allowbreak{}a & n/\allowbreak{}a & 100.0 \\
RidgeFill + random rows & 70.0 & n/\allowbreak{}a & +0.911 & +2.860 & n/\allowbreak{}a & n/\allowbreak{}a & n/\allowbreak{}a \\
RidgeFill + longer tail & 74.8 & n/\allowbreak{}a & +0.811 & +2.415 & n/\allowbreak{}a & n/\allowbreak{}a & n/\allowbreak{}a \\
No fill + longer tail & 13.2 & n/\allowbreak{}a & +1.419 & +3.139 & n/\allowbreak{}a & n/\allowbreak{}a & n/\allowbreak{}a \\
Identity + question-blind rows & 51.3 & n/\allowbreak{}a & n/\allowbreak{}a & n/\allowbreak{}a & n/\allowbreak{}a & n/\allowbreak{}a & n/\allowbreak{}a \\
RidgeFill + question-blind rows & 83.6 & n/\allowbreak{}a & n/\allowbreak{}a & n/\allowbreak{}a & n/\allowbreak{}a & n/\allowbreak{}a & n/\allowbreak{}a \\
\bottomrule
\end{tabular}
\end{table}

\subsection{Qwen3.5-9B}\label{sec:A.2}
\begin{table}[H]
\centering
\caption{Primary split, Qwen3.5-9B.}\label{tab:A.2}
\scriptsize
\setlength{\tabcolsep}{4pt}
\begin{tabular}{llllllll}
\toprule
Arm & \begin{tabular}[b]{@{}l@{}}RULER \\ filled, \\ 8K--32K (n \\ = 139)\end{tabular} & \begin{tabular}[b]{@{}l@{}}RULER \\ filled, \\ 8K--64K (n \\ = 219)\end{tabular} & \begin{tabular}[b]{@{}l@{}}$\Delta$NLL \\ novel\end{tabular} & $\Delta$NLL key & \begin{tabular}[b]{@{}l@{}}Decode \\ agree\end{tabular} & Decode KL & Turn 2 \\
\midrule
Exact & 99.8 & 99.9 & 0.000 & 0.000 & n/\allowbreak{}a & n/\allowbreak{}a & 100.0 \\
No fill & 0.0 & n/\allowbreak{}a & +0.046 & +2.760 & 96.8~\% & 0.045 & n/\allowbreak{}a \\
Identity & 80.4 & 78.1 & +0.023 & +1.548 & 97.7~\% & 0.021 & n/\allowbreak{}a \\
POP native & 100.0 & n/\allowbreak{}a & +0.016 & +0.235 & n/\allowbreak{}a & n/\allowbreak{}a & n/\allowbreak{}a \\
RidgeFill & 83.6 & 81.4 & +0.024 & +1.201 & 98.2~\% & 0.012 & 86.0 \\
RidgeFill + rows & 99.8 & 99.9 & +0.008 & +0.431 & n/\allowbreak{}a & n/\allowbreak{}a & 100.0 \\
Identity + rows & 100.0 & 99.1 & +0.007 & +0.485 & n/\allowbreak{}a & n/\allowbreak{}a & 99.0 \\
RidgeFill + random rows & 81.5 & n/\allowbreak{}a & +0.017 & +0.899 & n/\allowbreak{}a & n/\allowbreak{}a & n/\allowbreak{}a \\
RidgeFill + longer tail & 83.3 & n/\allowbreak{}a & +0.013 & +0.674 & n/\allowbreak{}a & n/\allowbreak{}a & n/\allowbreak{}a \\
No fill + longer tail & 26.3 & n/\allowbreak{}a & +0.011 & +0.759 & n/\allowbreak{}a & n/\allowbreak{}a & n/\allowbreak{}a \\
Identity + question-blind rows & 77.5 & n/\allowbreak{}a & n/\allowbreak{}a & n/\allowbreak{}a & n/\allowbreak{}a & n/\allowbreak{}a & n/\allowbreak{}a \\
RidgeFill + question-blind rows & 85.8 & n/\allowbreak{}a & n/\allowbreak{}a & n/\allowbreak{}a & n/\allowbreak{}a & n/\allowbreak{}a & n/\allowbreak{}a \\
\bottomrule
\end{tabular}
\end{table}

\subsection{Qwen3-8B}\label{sec:A.3}
\begin{table}[H]
\centering
\caption{Primary split, Qwen3-8B. RULER filled uses 138 samples at 8K to 32K. ``Eight RULER tasks'' is the score on the four needle tasks plus cwe, fwe, qa\_1 and vt over every sample at 8K to 32K, filled or not: 600 samples. *These arms ran only the four needle tasks, so their score is over those 300 samples. For identity + rows, moving the question first changes filled-region RULER by $-$2.4 points [$-$9.1, +4.4]. Random rows and the longer tail differ on 13 of the 138 filled-region samples and 27 of the 300 needle samples, but their totals are equal.}\label{tab:A.3}
\scriptsize
\setlength{\tabcolsep}{4pt}
\begin{tabular}{llllllll}
\toprule
Arm & \begin{tabular}[b]{@{}l@{}}RULER \\ filled (n \\ = 138)\end{tabular} & \begin{tabular}[b]{@{}l@{}}Eight \\ RULER \\ tasks\end{tabular} & \begin{tabular}[b]{@{}l@{}}$\Delta$NLL \\ novel\end{tabular} & $\Delta$NLL key & \begin{tabular}[b]{@{}l@{}}Decode \\ agree\end{tabular} & Decode KL & Turn 2 \\
\midrule
Exact & 98.7 & 91.8 & 0.000 & 0.000 & n/\allowbreak{}a & n/\allowbreak{}a & 100.0 \\
No fill & 0.0 & 19.0 & +0.243 & +3.072 & 88.4~\% & 0.172 & n/\allowbreak{}a \\
Identity & 44.0 & 39.0 & +0.026 & +1.833 & 95.7~\% & 0.032 & n/\allowbreak{}a \\
POP native & 98.7 & 86.3 & +0.044 & +0.255 & n/\allowbreak{}a & n/\allowbreak{}a & n/\allowbreak{}a \\
RidgeFill & 93.7 & 76.4 & +0.044 & +1.434 & 96.9~\% & 0.013 & 97.3 \\
RidgeFill + rows & 89.9 & 81.7 & +0.009 & +0.735 & n/\allowbreak{}a & n/\allowbreak{}a & 89.3 \\
Identity + rows & 63.4 & 65.1 & $-$0.005 & +0.901 & n/\allowbreak{}a & n/\allowbreak{}a & 54.7 \\
RidgeFill + random rows & 93.8 & 94.8* & +0.016 & +0.974 & n/\allowbreak{}a & n/\allowbreak{}a & n/\allowbreak{}a \\
RidgeFill + longer tail & 93.8 & 94.8* & +0.031 & +0.955 & n/\allowbreak{}a & n/\allowbreak{}a & n/\allowbreak{}a \\
No fill + longer tail & 25.4 & 51.7* & +0.130 & +1.645 & n/\allowbreak{}a & n/\allowbreak{}a & n/\allowbreak{}a \\
Identity + rows, question first & 61.1 & 70.6* & n/\allowbreak{}a & n/\allowbreak{}a & n/\allowbreak{}a & n/\allowbreak{}a & n/\allowbreak{}a \\
RidgeFill + rows, question first & 79.6 & 84.9* & n/\allowbreak{}a & n/\allowbreak{}a & n/\allowbreak{}a & n/\allowbreak{}a & n/\allowbreak{}a \\
\bottomrule
\end{tabular}
\end{table}

\section{Additional contrasts}\label{app:B}
Paired contrasts not shown in \S\ref{sec:6}, at the primary split unless the row says otherwise. RULER is filled-region accuracy in points; $\Delta$NLL is in nats per token.

\begin{table}[H]
\centering
\caption{Additional paired contrasts.}\label{tab:B.1}
\small
\setlength{\tabcolsep}{4pt}
\begin{tabularx}{\linewidth}{>{\raggedright\arraybackslash\hsize=1.000\hsize}Xllll}
\toprule
Contrast & Metric & n & $\Delta$ & 95~\% CI \\
\midrule
Llama: RidgeFill $-$ no fill & RULER filled & 136 & +60.7 & [+53.2, +67.6] \\
Llama: RidgeFill $-$ no fill & $\Delta$NLL novel & 126 & $-$0.586 & [$-$0.844, $-$0.266] \\
Llama: RidgeFill $-$ POP native & $\Delta$NLL novel & 126 & +0.633 & [+0.401, +0.921] \\
Llama: RidgeFill $-$ identity at 1/\allowbreak{}2 & RULER filled & 136 & +12.5 & [+8.6, +17.5] \\
Llama: RidgeFill $-$ identity, 64K only & RULER filled & 80 & +23.8 & [+13.6, +32.8] \\
Llama: RidgeFill + rows $-$ exact, 64K only & RULER filled & 80 & +0.6 & [0.0, +3.1] \\
Llama: identity + rows $-$ exact & RULER filled & 216 & $-$0.6 & [$-$2.5, +0.5] \\
Llama: identity + rows question-blind $-$ identity + rows & RULER filled & 136 & $-$47.1 & [$-$52.6, $-$41.2] \\
Llama: RidgeFill + rows $-$ identity + rows & Turn 2 & 100 & 0.0 & all $\Delta$ = 0 \\
Qwen3.5: RidgeFill $-$ no fill & $\Delta$NLL novel & 126 & $-$0.022 & [$-$0.032, $-$0.014] \\
Qwen3.5: RidgeFill $-$ POP native & $\Delta$NLL novel & 126 & +0.008 & [+0.003, +0.014] \\
Qwen3.5: identity + rows $-$ exact & RULER filled & 219 & $-$0.8 & [$-$2.7, +0.1] \\
Qwen3.5: identity + rows question-blind $-$ identity + rows & RULER filled & 139 & $-$22.5 & [$-$28.4, $-$17.3] \\
Qwen3.5: RidgeFill + rows $-$ identity + rows & Turn 2 & 100 & +1.0 & [0.0, +6.0] \\
Qwen3-8B: RidgeFill $-$ no fill & $\Delta$NLL novel & 90 & $-$0.198 & [$-$0.238, $-$0.167] \\
Qwen3-8B: identity + rows $-$ exact & RULER filled & 138 & $-$35.3 & [$-$41.3, $-$29.5] \\
Qwen3-8B: RidgeFill + rows $-$ identity + rows & Turn 2 & 75 & +34.7 & [+22.7, +46.7] \\
\bottomrule
\end{tabularx}
\end{table}

\section{RULER by task and length}\label{app:C}
Filled-region samples at the primary split and tail. niah\_multiquery at 8K has no filled-region sample on any model, and the no-fill arm did not run at 64K, so where a row spans 8K to 64K its no-fill cell covers 8K to 32K only.

\begin{table}[H]
\centering
\caption{RULER by task and length, Llama-3.1-8B.}\label{tab:C.1}
\scriptsize
\setlength{\tabcolsep}{4pt}
\begin{tabular}{lllllllll}
\toprule
Task & Length & n & Exact & RidgeFill & Identity & No fill & \begin{tabular}[b]{@{}l@{}}RidgeFill \\ + rows\end{tabular} & \begin{tabular}[b]{@{}l@{}}Identity + \\ rows\end{tabular} \\
\midrule
multikey\_1 & 8K & 10 & 100.0 & 80.0 & 10.0 & 0.0 & 100.0 & 100.0 \\
multikey\_1 & 16K & 19 & 100.0 & 63.2 & 5.3 & 0.0 & 100.0 & 100.0 \\
multikey\_1 & 32K & 20 & 100.0 & 50.0 & 20.0 & 0.0 & 100.0 & 100.0 \\
multikey\_1 & 64K & 24 & 100.0 & 20.8 & 0.0 & n/\allowbreak{}a & 100.0 & 100.0 \\
multiquery & 16K & 6 & 100.0 & 45.8 & 0.0 & 0.0 & 100.0 & 100.0 \\
multiquery & 32K & 12 & 97.9 & 47.9 & 0.0 & 0.0 & 97.9 & 97.9 \\
multiquery & 64K & 15 & 96.7 & 6.7 & 1.7 & n/\allowbreak{}a & 96.7 & 98.3 \\
multivalue & 8K & 2 & 100.0 & 87.5 & 25.0 & 0.0 & 100.0 & 100.0 \\
multivalue & 16K & 9 & 100.0 & 61.1 & 8.3 & 0.0 & 97.2 & 88.9 \\
multivalue & 32K & 15 & 98.3 & 18.3 & 16.7 & 0.0 & 100.0 & 93.3 \\
multivalue & 64K & 18 & 97.2 & 1.4 & 5.6 & n/\allowbreak{}a & 100.0 & 98.6 \\
single\_1 & 8K & 9 & 100.0 & 77.8 & 22.2 & 0.0 & 100.0 & 100.0 \\
single\_1 & 16K & 15 & 100.0 & 80.0 & 20.0 & 0.0 & 100.0 & 100.0 \\
single\_1 & 32K & 19 & 100.0 & 78.9 & 5.3 & 0.0 & 100.0 & 100.0 \\
single\_1 & 64K & 23 & 100.0 & 78.3 & 17.4 & n/\allowbreak{}a & 100.0 & 100.0 \\
\bottomrule
\end{tabular}
\end{table}

\begin{table}[H]
\centering
\caption{RULER by task and length, Qwen3.5-9B. single\_1 is pooled over 8K to 64K.}\label{tab:C.2}
\scriptsize
\setlength{\tabcolsep}{4pt}
\begin{tabular}{lllllllll}
\toprule
Task & Length & n & Exact & RidgeFill & Identity & No fill & \begin{tabular}[b]{@{}l@{}}RidgeFill \\ + rows\end{tabular} & \begin{tabular}[b]{@{}l@{}}Identity + \\ rows\end{tabular} \\
\midrule
multikey\_1 & 8K & 10 & 100.0 & 70.0 & 80.0 & 0.0 & 100.0 & 100.0 \\
multikey\_1 & 16K & 19 & 100.0 & 63.2 & 68.4 & 0.0 & 100.0 & 100.0 \\
multikey\_1 & 32K & 21 & 100.0 & 81.0 & 66.7 & 0.0 & 100.0 & 100.0 \\
multikey\_1 & 64K & 24 & 100.0 & 62.5 & 66.7 & n/\allowbreak{}a & 100.0 & 95.8 \\
multiquery & 16K & 6 & 100.0 & 33.3 & 58.3 & 0.0 & 100.0 & 100.0 \\
multiquery & 32K & 12 & 97.9 & 87.5 & 83.3 & 0.0 & 97.9 & 100.0 \\
multiquery & 64K & 15 & 100.0 & 68.3 & 60.0 & n/\allowbreak{}a & 100.0 & 93.3 \\
multivalue & 8K & 2 & 100.0 & 100.0 & 100.0 & 0.0 & 100.0 & 100.0 \\
multivalue & 16K & 9 & 100.0 & 91.7 & 72.2 & 0.0 & 100.0 & 100.0 \\
multivalue & 32K & 16 & 100.0 & 84.4 & 67.2 & 0.0 & 100.0 & 100.0 \\
multivalue & 64K & 18 & 100.0 & 76.4 & 62.5 & n/\allowbreak{}a & 100.0 & 100.0 \\
single\_1 & 8K--64K & 67 & 100.0 & 100.0 & 100.0 & 0.0 & 100.0 & 100.0 \\
\bottomrule
\end{tabular}
\end{table}

\begin{table}[H]
\centering
\caption{RULER by task and length, Qwen3-8B.}\label{tab:C.3}
\scriptsize
\setlength{\tabcolsep}{4pt}
\begin{tabular}{lllllllll}
\toprule
Task & Length & n & Exact & RidgeFill & Identity & No fill & \begin{tabular}[b]{@{}l@{}}RidgeFill \\ + rows\end{tabular} & \begin{tabular}[b]{@{}l@{}}Identity + \\ rows\end{tabular} \\
\midrule
multikey\_1 & 8K & 10 & 100.0 & 90.0 & 50.0 & 0.0 & 100.0 & 80.0 \\
multikey\_1 & 16K & 19 & 100.0 & 94.7 & 31.6 & 0.0 & 84.2 & 63.2 \\
multikey\_1 & 32K & 21 & 95.2 & 90.5 & 23.8 & 0.0 & 85.7 & 38.1 \\
multiquery & 16K & 6 & 100.0 & 91.7 & 8.3 & 0.0 & 87.5 & 54.2 \\
multiquery & 32K & 12 & 95.8 & 81.2 & 16.7 & 0.0 & 70.8 & 31.2 \\
multivalue & 8K & 2 & 100.0 & 100.0 & 37.5 & 0.0 & 87.5 & 75.0 \\
multivalue & 16K & 9 & 100.0 & 94.4 & 19.4 & 0.0 & 83.3 & 36.1 \\
multivalue & 32K & 16 & 98.4 & 90.6 & 23.4 & 0.0 & 87.5 & 29.7 \\
single\_1 & 8K & 9 & 100.0 & 100.0 & 88.9 & 0.0 & 100.0 & 100.0 \\
single\_1 & 16K & 15 & 100.0 & 100.0 & 80.0 & 0.0 & 100.0 & 100.0 \\
single\_1 & 32K & 19 & 100.0 & 100.0 & 84.2 & 0.0 & 100.0 & 100.0 \\
\bottomrule
\end{tabular}
\end{table}

\section{Second question on a once-filled cache}\label{app:D}
The second question is asked of the cache filled, and the rows chosen, for the first. The prompts are the niah\_multikey\_1 samples, which carry four keyed needles. The second question asks for one of the other three keys, chosen by a per-sample seed that is the same on every model. It is appended after the first answer and runs exactly on the carried cache; rows are not chosen again. An answer counts if the first seven-digit number in the reply is right, which is stricter than RULER's substring match. Under RULER's scorer, two cells of Table~\ref{tab:D.1} change, both for Llama's RidgeFill alone: it goes from 57.0 to 62.0, and from 46.9 to 53.1 with the second needle in the filled region.

\begin{table}[H]
\centering
\caption{Second-question accuracy. In 19 of the 100 samples (17 of the 75 on Qwen3-8B) the second needle lies in the exact tail; the last two rows leave those out. On Qwen3-8B, RidgeFill + rows $-$ exact is $-$10.7 points [$-$18.7, $-$5.3].}\label{tab:D.1}
\small
\setlength{\tabcolsep}{4pt}
\begin{tabular}{llll}
\toprule
Arm & Llama-3.1-8B & Qwen3.5-9B & Qwen3-8B \\
\midrule
Samples & 100 & 100 & 75 \\
Exact & 100.0 & 100.0 & 100.0 \\
RidgeFill alone & 57.0 & 86.0 & 97.3 \\
RidgeFill + rows & 100.0 & 100.0 & 89.3 \\
Identity + rows & 100.0 & 99.0 & 54.7 \\
RidgeFill alone, second needle in the filled region & 46.9 & 82.7 & 96.6 \\
RidgeFill + rows, second needle in the filled region & 100.0 & 100.0 & 86.2 \\
\bottomrule
\end{tabular}
\end{table}

The test is weaker than it looks. All four needles share one template, so the rows chosen for the first question already cover a median of 85~\%, 86~\% and 54~\% of the second needle's tokens when that needle is in the filled region. It tests a sibling needle that the selector already favored, not an unrelated question (\S\ref{sec:8}).

\section{RULER prompts in the serving engines}\label{app:E}
Both engines put the question and the answer format in a single user turn, prefill no answer prefix, and run with thinking off. They word the instruction differently.

\begin{itemize}[leftmargin=*,itemsep=2pt,topsep=4pt]
\item On Qwen3.8-27B (llama.cpp), the prompt keeps RULER's question line and appends \texttt{Answer with the requested answer(s) in the form:\allowbreak{} [answer prefix] \textless{}answer(s)\textgreater{}.\allowbreak{}}
\item On Qwen3.8-Flash-Next (vLLM), the prompt replaces the question line with \texttt{Answer the following question:\allowbreak{} [question]} and adds \texttt{In your answer, respond with the following format:\allowbreak{} [answer prefix]:\allowbreak{} \textless{}number\textgreater{}, \ldots{}}, with one \texttt{\textless{}number\textgreater{}} per expected value. The run's records confirm nothing was prefilled: the model wrote the whole answer sentence itself.
\end{itemize}

\section{Builds and file hashes}\label{app:F}
\textbf{Reference study.} The three reference models are the post-trained releases (Table~\ref{tab:F.1}). Llama-3.1-8B is Llama-3.1-8B-Instruct, and Qwen3.5-9B, which Hugging Face lists as an image-and-text model, runs on its text path only. The runs used four commits of the reference implementation (Table~\ref{tab:F.2}). The data repository's \texttt{code/\allowbreak{}} folder holds the implementation, and \texttt{code/\allowbreak{}run-\allowbreak{}commit-\allowbreak{}diffs/\allowbreak{}} holds each run commit as a diff against it. The per-document $\Delta$NLL of every reference arm and depth is one CSV, \texttt{derived/\allowbreak{}per-\allowbreak{}document-\allowbreak{}nll.\allowbreak{}csv}. It covers 342 model--document pairs on the evaluation documents (126 on Llama-3.1-8B and on Qwen3.5-9B, 90 on Qwen3-8B) and, besides those, Llama-3.1-8B's 30 fresh documents.

\begin{table}[H]
\centering
\caption{Reference model checkpoints.}\label{tab:F.1}
\small
\setlength{\tabcolsep}{4pt}
\begin{tabularx}{\linewidth}{lll>{\raggedright\arraybackslash\hsize=1.000\hsize}X}
\toprule
Model & Checkpoint & Revision & Note \\
\midrule
Llama-3.1-8B & unsloth/\allowbreak{}Llama-3.1-8B-Instruct & 4699cc75b550 &  \\
Qwen3.5-9B & Qwen/\allowbreak{}Qwen3.5-9B & c20223623576 & post-trained from Qwen3.5-9B-Base; text path only \\
Qwen3-8B & Qwen/\allowbreak{}Qwen3-8B & b968826d9c46 &  \\
\bottomrule
\end{tabularx}
\end{table}

\begin{table}[H]
\centering
\caption{Commits of the reference implementation and the runs each one produced.}\label{tab:F.2}
\small
\setlength{\tabcolsep}{4pt}
\begin{tabularx}{\linewidth}{l>{\raggedright\arraybackslash\hsize=1.000\hsize}X}
\toprule
Commit & Runs \\
\midrule
42028dfc & the original loss runs, decode agreement and second-question runs on all three models; Qwen3-8B's RULER \\
253f845a & Llama-3.1-8B's RULER; the second loss runs on Llama-3.1-8B and Qwen3.5-9B \\
21ccd244 & Qwen3.5-9B's RULER; Qwen3-8B's second loss run; Qwen3-8B's 5/\allowbreak{}8-depth comparison with POP \\
019ae056 & Llama-3.1-8B's LongBench v1; the fresh documents; Llama-3.1-8B's 5/\allowbreak{}8-depth comparison with POP \\
\bottomrule
\end{tabularx}
\end{table}

\textbf{Map format.} The 27B and Flash-Next maps are int8, with one bf16 scale per 128 weights of each row and a bf16 bias. The 27B release holds 48 maps, for layers 16 to 63, 1.2 GiB in all. It is in the \texttt{ridgefill-\allowbreak{}projector-\allowbreak{}qwen3.\allowbreak{}8-\allowbreak{}27b-\allowbreak{}i8/\allowbreak{}} folder of its Hugging Face repository (revision 69e3ea99), next to the converted GGUF, which is unchanged since revision 07eb31cd. The Flash-Next release holds 24 maps, for layers 24 to 47, 0.60 GiB. With the terminal-state correction for the 18 Gated DeltaNet layers above the split and the token-frequency table that the protected rows use (\S\ref{sec:2.4}), it comes to 0.65 GiB across 27 files. The repository's \texttt{source/\allowbreak{}} folder also holds the bf16 fit that the int8 maps were quantized from; the Flash-Next run kit copies its \texttt{final} map, which no reported number reads, into the engine's files. The reference study's primary-split maps for Llama-3.1-8B (split 12) and Qwen3-8B (split 13) are published as GGUF files for the fork, in bf16 and in Q8\_0. The reference maps at all four splits are also published in bf16 as safetensors, one file per split: in a \texttt{safetensors/\allowbreak{}} folder of the Llama-3.1-8B and Qwen3-8B repositories, and at the top level of the Qwen3.5-9B repository.

\textbf{Serving builds.} Table~\ref{tab:F.3} lists the files the final serving runs loaded. On the 27B, the fork and its commit are in \S\ref{sec:9}, the flags and hardware in \S\ref{sec:4.1}, and the fork's RIDGEFILL.md gives the full server command of the measured configuration. The 27B maps file is the int8 release converted to llama.cpp's Q8\_0 format by the fork's \texttt{convert\_\allowbreak{}ridgefill\_\allowbreak{}to\_\allowbreak{}gguf.\allowbreak{}py}, and the public file has the same hash. The Flash-Next runs used the run kit at commit d1164746, in its \texttt{c5/\allowbreak{}} directory, which is \texttt{code/\allowbreak{}c5/\allowbreak{}} in the data repository, with that commit as a diff in \texttt{code/\allowbreak{}run-\allowbreak{}commit-\allowbreak{}diffs/\allowbreak{}}. Flash-Next serves our MXFP4/FP8 quantization, with MXFP4 experts and FP8 attention, made by an activation-aware quantizer with GPTQ error feedback (Frantar et al., 2023). Its engine is vLLM 0.29.0.dev0 at upstream commit 2bdbbc80 plus our fork's patches, shipped as a public image; the patches themselves are not public. The determinism fix replaces one file in the image. The runs load the released int8 maps, dequantized to bf16 by the kit's \texttt{c5/\allowbreak{}convert.\allowbreak{}py}. The timing prompts are the first 9,216, 16,384 and 32,768 tokens of chat\_32768\_0, a chat document from UltraChat (Ding et al., 2023) fixed before the runs.

\begin{table}[H]
\centering
\caption{Files loaded by the final serving runs.}\label{tab:F.3}
\footnotesize
\setlength{\tabcolsep}{4pt}
\begin{tabularx}{\linewidth}{l>{\raggedright\arraybackslash\hsize=1.000\hsize}X>{\raggedright\arraybackslash\hsize=1.000\hsize}X>{\raggedright\arraybackslash\hsize=1.000\hsize}X}
\toprule
Deployment & File & Source and revision & sha256 \\
\midrule
27B & model, Qwen3.8-27B-UD-Q4\_K\_XL.gguf & unsloth/\allowbreak{}Qwen3.8-27B-GGUF, {\ttfamily 4ca72078\allowbreak{}8d1e01f1\allowbreak{}bff70c03\allowbreak{}3e0d0028\allowbreak{}fd02e502} & {\ttfamily 3f227079\allowbreak{}003add25\allowbreak{}11437e5b\allowbreak{}1e94812e\allowbreak{}36338522\allowbreak{}5bf6a9b4\allowbreak{}7b0054a7\allowbreak{}2bc8b01e} \\
27B & maps, ridgefill-projector-qwen3.8-27b-i8-S16.gguf & TheCacheIsALie/\allowbreak{}RidgeFill-Projector-Qwen3.8-27B-i8, 69e3ea99 (GGUF unchanged since 07eb31cd) & {\ttfamily 8646d41a\allowbreak{}5a28d60e\allowbreak{}f53cc5e1\allowbreak{}fee4b2bf\allowbreak{}b20d0775\allowbreak{}db88d3f8\allowbreak{}9d0731ad\allowbreak{}9f6b629d} \\
Flash-Next & model & tcclaviger/\allowbreak{}Qwen3.8-Flash-Next-MXFP4-FP8-GPTQ, 6fe3e746 & none; multi-file repository pinned by revision \\
Flash-Next & engine image & tcclaviger/\allowbreak{}vllm:29.08.10 & digest {\ttfamily 809657dd\allowbreak{}b1533db1\allowbreak{}0a0c0826\allowbreak{}7c3ee6fd\allowbreak{}e180b839\allowbreak{}e3797134\allowbreak{}14efa121\allowbreak{}415914a3} \\
Flash-Next & determinism fix, replacing vllm/\allowbreak{}models/\allowbreak{}qwen4\_exp/\allowbreak{}amd/\allowbreak{}ops/\allowbreak{}qsa.py & data repository, code/\allowbreak{}engine-fix/\allowbreak{} (the diff and both files' hashes) & {\ttfamily eaf0dd19\allowbreak{}75799f4c\allowbreak{}8897cbea\allowbreak{}62791de4\allowbreak{}4fd28322\allowbreak{}758fb989\allowbreak{}1cbe7ff0\allowbreak{}ef52db99} \\
Flash-Next & maps & TheCacheIsALie/\allowbreak{}RidgeFill-Projector-Qwen3.8-Flash-Next-i8, adb763a7 (map files unchanged since 0b6f2ce0) & manifest ridgefill.json {\ttfamily 795363fa\allowbreak{}153f0e7e\allowbreak{}16454af6\allowbreak{}acabd644\allowbreak{}709e703d\allowbreak{}84de0dd9\allowbreak{}53556522\allowbreak{}6bb51faa} \\
Flash-Next & final map, copied unchanged into the engine's files and never read by a reported number & same repository, source/\allowbreak{}kva-big-s24.safetensors & {\ttfamily c9db6064\allowbreak{}eb275d31\allowbreak{}66df8139\allowbreak{}239959df\allowbreak{}deb64f27\allowbreak{}78707a2f\allowbreak{}42215164\allowbreak{}999cb9cd} \\
\bottomrule
\end{tabularx}
\end{table}

\section{Calibration}\label{app:G}
\textbf{Calibration text.} Every map in this paper is fitted on the same 71 documents, and the Flash-Next maps add 80 more (below). Five more documents are held out to choose $\lambda$.

\textbf{Reference and 27B maps.} Every position of every calibration document is a row, and no document reaches the reference implementation's cap of 32,768 tokens. The tokenizers differ, so the counts differ by model (Table~\ref{tab:G.1}). The 27B maps are fitted on the activations of the served Q4\_K\_XL model, dequantized to bf16 and run in the reference implementation.

\begin{table}[H]
\centering
\caption{Calibration rows and fit times. A fit is one capture pass, then an fp64 solve on the CPU and the $\lambda$ search for each split.}\label{tab:G.1}
\small
\setlength{\tabcolsep}{4pt}
\begin{tabularx}{\linewidth}{lll>{\raggedright\arraybackslash\hsize=1.021\hsize}X>{\raggedright\arraybackslash\hsize=0.979\hsize}X}
\toprule
Model & Rows & \begin{tabular}[b]{@{}l@{}}Median \\ document \\ (tokens)\end{tabular} & Longest document (tokens) & Fit time \\
\midrule
Llama-3.1-8B & 834,541 & 11,219 & 28,352, counting the beginning-of-sequence token & about 24 min, one RTX 4090 \\
Qwen3.5-9B & 839,652 & 11,563 & 30,465 & about 44 min, one A100, including the download \\
Qwen3-8B & 845,043 & 11,482 & 29,009 & 31 min, one R9700 \\
Qwen3.8-27B & 839,652 & 11,563 & 30,465 & 26 min, one RTX 5090 \\
\bottomrule
\end{tabularx}
\end{table}

For each split, the ridge penalty's setting $\lambda$ (\S\ref{sec:2.2}) is chosen from nine values between $10^{-4}$ and 1 by the cache-write cosine of \S\ref{sec:2.2} on the five held-out documents. It is 0.03 at every split on Llama-3.1-8B, Qwen3.5-9B and the 27B, and 0.01 on Qwen3-8B except at split 18, where it is 0.003.

\textbf{Flash-Next maps.} The Flash-Next maps are fitted on the served MXFP4/FP8 model's own activations, captured through our vLLM fork at release 29.07.16. Both capture sessions record that release's image, digest 2703275702ab6316ee91502df75486b421d0d661a99b04000c08859d91a42585, as the running image (data repository, provenance/fits/flashnext/capture-sessions/). The corpus is 151 documents: the 71 above, plus 80 more, the longest about 20,000 tokens. Chat-formatted copies of 102 of them are added. Each document runs whole, as one prompt, and every eighth token position becomes a row, with its full preceding context. The capture code writes only prefill steps of at least 256 tokens, so a document's short final step is skipped. This gives 425,789 rows. The fit weights the chat rows so that they carry half the total weight, as if there were 497,846 rows. As on the reference models, no row lies beyond token 30,465. $\lambda$ = 0.03 was chosen from six values between 0.001 and 0.3 by the layer-input cosine of \S\ref{sec:2.2}, over ten captures: the five held-out documents and their chat copies. The terminal-state correction is fitted on 13 further prompts (\S\ref{sec:2.3}). The capture and fit took about 2 hours on two R9700s.

\textbf{Overlap with the test sets.} Three checks compare the calibration text with the test sets. The first works in tokens, on the text as each model tokenizes it, against the 126 evaluation documents and the 30 fresh ones. No calibration document has the same hash as any of them, and no document used to fit a map shares a run of 50 or more tokens with them; the longest shared run is 47 tokens. Shorter runs are common. At eight tokens most documents share some phrase, and every shared run of 32 tokens or more is license or template boilerplate or a number sequence. One document used only to choose $\lambda$, and never fitted on, shares an 80-token license header with one fresh source-code document, and shorter pieces of it with two others. In words, that header is a 40-word run, under the 50-word rule of \S\ref{sec:6.7}.

The second check works in words, against the reference study's evaluation set and the raw pool of LongBench v1 documents. The evaluation set shares no run of 50 words with the calibration text. Of the 99 documents first assembled for calibration, 28 shared such a run with the raw pool, though never with an evaluated item, and so did one prompt for the terminal-state correction. All 28 are excluded from every map fit. Two uses remain. Flash-Next's terminal-state correction includes that one prompt, and the token-frequency table counts the 28 documents; both Flash-Next's protected rows and the 27B's base-rule count use that table (\S\ref{sec:2.4}). Neither check covered LongBench v2.

The third check covers it. It uses the second check's 50-word rule and also compares whole-text hashes. It compares all 304 calibration texts with LongBench v2's 503 items and with every 27B evaluation prompt. The 304 are every fit, $\lambda$, chat-copy and terminal-state document, plus the 28 excluded ones. No calibration text has the same hash as any evaluation text or file. Thirty-eight document--item pairs share a run of 50 or more words, and one of them is content: a novel chapter, in a different edition, that a document used only for the Flash-Next maps shares with one Single-Document QA item, over a run of 70 words. Flash-Next is not scored on LongBench v2, and the 27B's maps never saw that document. The other pairs are license headers in Code Repository items. Four fit documents, all among those the 27B's maps were fitted on, share runs of 78 to 85 words with five items. Six excluded documents, which only the token-frequency table counts, share up to 162 words with eleven items, four of them among those five, so 12 items in all; the longest run is a complete MIT license. No $\lambda$ document or terminal-state prompt shares a 50-word run with any evaluation text, and no 27B RULER or timing prompt shares one with any calibration text. The script and its result are in the data repository.

\section{Derivations and formal statements}\label{app:H}
\textbf{Exact rows and the cost model.} An exact prefix row does the same compute in the late layers as a tail row, but it takes $\kappa$ times as long there, because it shares each read of the layer weights with far fewer rows. On the 27B, the exact rows are scattered through the prompt, so the second sweep runs them in smaller, less efficient batches than the contiguous tail, and each of those micro-batches reads the late layers' weights in full (\S\ref{sec:4.4}).

On Flash-Next the sharing happens inside each expert. Flash-Next prefills in chunks of 2,048 tokens, and each late layer sends every token to 10 of its 512 routed experts. A tail chunk touches nearly every expert, so each expert's weights are read once for about 40 tokens. In a prefix chunk under quality mode, only the protected rows run the expert blocks. On the Flash-Next run they were 3 to 10~\% of each prompt's prefix, so about 60 to 200 rows of a chunk if they spread evenly. If routing spreads evenly, those rows still touch about 70 to 98~\% of the experts, so nearly the same weights are read for a tenth of the rows or fewer, about 10 to 24 times a tail row's expert traffic per row. The routed experts held in host memory, about two thirds of them (16,042 of 24,576 per GPU), are read over PCIe; the rest, the shared expert and the attention weights stay in VRAM. The even-routing assumption is ours, and neither deployment was profiled, so both accounts are explanations that a profile would have to confirm.

So the prefill time grows in a straight line with the number $E$ of exact prefix rows:

\[\tau(E) = \tau_0 + E\,\sigma ,\]

where $\tau_0$ is the time with no exact rows and $\sigma$, the time one exact row adds, is $\kappa$ times a tail row's time in the late layers. If prefill time follows this model, two measured times fix the line: $\tau_0$, and $\tau(E_1)$ for some number $E_1$ of exact rows, which give $\sigma = (\tau(E_1) - \tau_0)/E_1$. RidgeFill with $E$ exact rows is then no slower than exact prefill, whose time is $\tau_\text{exact}$, when

\[E \le (\tau_\text{exact} - \tau_0)/\sigma .\]

In words, the time RidgeFill saves with no exact rows, divided by the time each exact row costs, is the number of exact rows it can afford. $\kappa$ has to be measured, and the final runs did not measure it: the 27B timings cover only stock and the full configuration, and no count of exact rows was recorded for the Flash-Next timing prompts. Quality mode adds exact rows and is slower than plain RidgeFill on Flash-Next (\S\ref{sec:5.3}). It also adds the terminal-state correction, but that is one addition per layer, so the rows are the likely cost; the final runs do not separate the two.

\textbf{Lexical-net normalization.} Words are runs of Unicode letters and number characters; punctuation, underscores and combining marks separate them. ASCII letters are lower-cased, and every other character is kept as it is. A final s is dropped from words of four or more characters unless the word ends in ss. A sentence ends at a line break; at a period, exclamation mark or question mark followed by whitespace or the end of the text; at the CJK full stop, exclamation mark or question mark; and at control, user-defined, unknown and unused tokens.

\textbf{Why the terminal-state correction is scaled by $\rho$.} A Gated DeltaNet layer keeps a running state. At each token it multiplies the state by a gate between 0 and 1, which fades older tokens, and adds the token's own update. The terminal-state correction adds a fixed offset $\Delta M_\ell$ to the state of layer $\ell$ at the end of the prefix, scaled by $\rho_\ell$, the filled rows' share of that state. Under a simple model of the layer, that scale removes the fill's shift exactly.

Take one layer. The delta rule also erases part of the old state before each write, in the direction of the token's key; for simplicity, drop that erase term. The state after token $t$ is then $M_t = g_t M_{t-1} + u_t$. The state $M_t$ and the update $u_t$ are matrices of the same size, one per head, and the gate $g_t$ is one number between 0 and 1 per head and token. Unrolling over the $n$ prefix rows gives the state at the end of the prefix:

\[M_n = \sum_{j=1}^{n} w_j u_j, \qquad w_j = \prod_{k=j+1}^{n} g_k .\]

Row $j$'s weight $w_j$ is the product of every gate after it, and the last prefix row's weight is 1.

The fill changes the updates of filled rows. Suppose, as an idealization, that it shifts each one by the same amount $\delta$, a matrix the size of the state. The state at the end of the prefix is then off by $\delta \sum_{j\,\text{filled}} w_j$. A fully filled prefix would be off by $\delta \sum_j w_j$, and that is what $\Delta M_\ell$ estimates. Multiplying $\Delta M_\ell$ by

\[\rho_\ell = \frac{\sum_{j\,\text{filled}} w_j}{\sum_j w_j},\]

the filled rows' share of the weight at the end of the prefix, gives $\delta \sum_{j\,\text{filled}} w_j$, which removes the shift exactly. Any other scale leaves part of it, and the unscaled correction overshoots by the exact prefix rows' share. Because the correction is added at the end of the prefix, the tail's gates fade it exactly as they fade the error, so it cancels in the final state too.

Behind the exact tail, the filled rows' share of the final state is small on layers that fade quickly (Figure~\ref{fig:4}). Suppose every gate is at most $\gamma$ \textless{} 1 and the last $m$ rows of the prompt are exact. Each filled row then has at least $m$ gates after it, so the filled rows' total weight in the final state is at most $\gamma^m + \gamma^{m+1} + \dots = \gamma^m/(1-\gamma)$. The total weight of all rows is at least 1, the last row's weight, so the filled rows' share is at most $\gamma^m/(1-\gamma)$. For $\gamma$ = 0.4 and m = 16 this is $7.2 \times 10^{-7}$. Only layers whose gates stay near 1 need the correction.

The bound also holds with the erase term: two runs that receive the same last $m$ inputs end with states that differ by at most $\gamma^m$ times their difference before those inputs. StCorrection.lean and GdnContraction.lean state both results (below).

\textbf{Formal statements.} The statements in Table~\ref{tab:H.1} are proved in Lean 4.34.1 (de Moura and Ullrich, 2021) against Mathlib v4.34.1 (The mathlib Community, 2020), in the KvaLean project (\texttt{code/\allowbreak{}lean/\allowbreak{}kva\_\allowbreak{}lean} in the data repository). The project has 219 theorems, none left unproved (no \texttt{sorry}), and a full \texttt{lake build} passes. Every module except a shared vector-norm helper, VecNorm.lean, ends with \texttt{\#print axioms}, which lists the axioms its proofs depend on, and each lists only the standard three: propext, Classical.choice and Quot.sound. All of the statements concern idealized models. They predict the sign and pattern of the splice and control results, not their size. The project also holds SoftmaxShift.lean, which the paper does not use.

\begin{table}[H]
\centering
\caption{The formal statements the paper relies on.}\label{tab:H.1}
\footnotesize
\setlength{\tabcolsep}{4pt}
\begin{tabularx}{\linewidth}{l>{\raggedright\arraybackslash\hsize=1.505\hsize}X>{\raggedright\arraybackslash\hsize=0.495\hsize}X}
\toprule
File & What it proves & Used in \\
\midrule
RidgeSums.lean & Summing over documents one at a time gives the same ridge solution as one solve over all rows, and the solution is unique for any ridge penalty $\alpha$ \textgreater{} 0. & \S\ref{sec:2.2} \\
StCorrection.lean & The $\rho$ scale, its uniqueness and the bound above, for scalar gates and a linear recurrence. & \S\ref{sec:2.3}, Appendix~\ref{app:H} \\
GdnContraction.lean & A bound on how fast the full delta rule forgets earlier inputs. & Appendix~\ref{app:H} \\
AttnSensitivity.lean & The value and key bounds, for single-head softmax attention without a causal mask. & \S\ref{sec:2.4} \\
ImpactBound.lean & Under a first-order impact model with errors of bounded size, keeping exact the rows of largest sensitivity gives the smallest bound. & \S\ref{sec:2.4} \\
RandomControl.lean & Under an additive gain model, a random subset of the rows gains on average its share of the total, so a random control of the same size measures only the selector's skill. & \S\ref{sec:3.3} \\
QkNormInvariance.lean & RMS normalization without its small constant $\varepsilon$ makes a key's scale invisible to attention; with $\varepsilon$ the effect is bounded. & \S\ref{sec:6.6} \\
CrossTerms.lean & For a quadratic loss, the change from two splices together minus the sum of their separate changes is exactly the cross term between them. & \S\ref{sec:6.6} \\
Speedup.lean & For the ceiling r of \S\ref{sec:2.5}, the speedup 1/\allowbreak{}r is at most L/\allowbreak{}S, and at least 1 when a filled row costs no more than an exact one. & \S\ref{sec:2.5} \\
SpeedupExactRows.lean & The exact-rows cost model above. & Appendix~\ref{app:H} \\
AffineFill.lean & An affine map equals a target on every input only if the target preserves every line; RMS normalization, with or without a learned gain, does not, so no affine fill equals it on every input. & \S\ref{sec:1} \\
ContextFill.lean & A fill that sees only a row's own state cannot equal a target whose output at that row also depends on the other rows; causal single-head softmax attention with a residual and a nonzero value map does, for any query and key maps, so no per-row fill equals it on every prompt, nor does one after an injective per-row map. & \S\ref{sec:1} \\
\bottomrule
\end{tabularx}
\end{table}

\section{Development measurements}\label{app:I}
These measurements come from the method's development, before the evaluation design of \S\ref{sec:3} was fixed. They support statements in \S\ref{sec:1}, \S\ref{sec:2.4} and \S\ref{sec:3}. They are not evidence for \S\ref{sec:4} to \S\ref{sec:6}, and their sets are small. The data repository holds each one's records under \texttt{results/\allowbreak{}development/\allowbreak{}}, marked with the status DEVELOPMENT, and \texttt{recompute/\allowbreak{}dev\_\allowbreak{}history.\allowbreak{}py} recomputes every number below. Intervals are 95~\% percentile bootstraps over the units named.

I.1 and the first two paragraphs of I.2 used an earlier development checkpoint: a community-modified build of Qwen3.8-Flash-Next, quantized to 4 bits (IQ4\_XS) and served in our own development engine. It is not the checkpoint of \S\ref{sec:5}; the data repository names it. The rest used the models of \S\ref{sec:4} and \S\ref{sec:6}.

\subsection{Corrections from approximation theory}\label{sec:I.1}
Each correction was fitted on 6 held-out documents and scored on 23 development prompts at split 16 with a 2,048-token tail. The measure is the relative error of the attention output, averaged over the late attention layers, against the plain fill (Table~\ref{tab:I.1}). $\Sigma$ is the covariance of the fill's error in a key.

\begin{table}[H]
\centering
\caption{Corrections to the filled keys and values on the development checkpoint. Positive is worse. Intervals resample the 23 prompts.}\label{tab:I.1}
\small
\setlength{\tabcolsep}{4pt}
\begin{tabular}{lll}
\toprule
Correction & \begin{tabular}[b]{@{}l@{}}Change in error,~\% [95~\% \\ CI]\end{tabular} & \begin{tabular}[b]{@{}l@{}}Prompts \\ improved\end{tabular} \\
\midrule
Jensen term, $+\tfrac{1}{2} q^\top \Sigma q$ on each logit & +62.9 [+55.1, +70.5] & 0 of 23 \\
Jensen term, sign reversed & +2.5 [$-$1.5, +6.1] & 11 of 23 \\
Jensen term, $\Sigma$ scaled per row & +80.2 [+60.2, +114.1] & 0 of 23 \\
Jensen term with the key--value cross-covariance & +70.1 [+61.7, +78.2] & 0 of 23 \\
Key--value cross-covariance alone & +1.8 [+0.9, +2.6] & 6 of 23 \\
Value map fitted to the attention output & +51.2 [+43.5, +57.7] & 1 of 23 \\
Key temperature per query head & $-$6.5 [$-$8.2, $-$5.0] & 23 of 23 \\
Key-mean shift & $-$8.6 [$-$11.1, $-$6.5] & 23 of 23 \\
\bottomrule
\end{tabular}
\end{table}

The key-mean shift, the largest gain, was then run end to end on the same 23 prompts, with the terminal-state correction in both arms. It changed the loss on novel tokens by +0.0010 nats [$-$0.0048, +0.0063]. The key temperature was not run end to end, because it added nothing on top of the shift.

Nyquist's argument needs an error that changes slowly from one row to the next. The fill's error does not. On three held-out documents, the correlation between a row's error and the previous row's, averaged over layers, is 0.17 to 0.21, and eight rows apart it is at most 0.06. Exact rows spread through the prompt in chunks did worse than the same number of exact rows at the end of the prompt in all 9 settings tried: 10, 25 and 50~\% of the prompt, in chunks of 128, 256 and 512 rows. At 25~\% in chunks of 256, the attention-output error was 0.26 against 0.22.

\subsection{Needles in the filled region}\label{sec:I.2}
On the development checkpoint at split 16, 60 needle prompts hid a seven-digit number at 8K, 16K and 32K tokens and five depths. With the terminal-state correction and a 2,048-token tail, 52 needles fell in the filled region; with a 4,096-token tail, 36 did. The fill retrieved 33 of 52 and 25 of 36, and every needle in the exact tail (8 of 8, 24 of 24). Of the 17 wrong numbers, 13 differ from the needle only in the last digit, for example 4729999 for 4729995. The exact model returned no wrong number, but 16 of its 60 answers held no number at all, because its reasoning used up the answer budget, so this run gives counts and no accuracy against exact.

With a corrected answer format, the exact model retrieved 60 of 60 and the fill 26 of the 36 needles in its filled region (4,096-token tail). Computing exactly the attention keys and values of the rows that hold digit tokens, 2.2 to 6.2~\% of the prompt, retrieved 36 of 36. Lengthening the exact tail instead, by at least as many rows (rounded up to whole blocks of 1,568, three to ten times as many), retrieved 28 of 36.

The reference models repeat the pattern and show where it stops (Table~\ref{tab:I.2}). The settings are split 12 with a 1,024-token tail and split 16 with 1,024 and 4,096 tokens; the unit is a prompt in one setting whose needle lies in the filled region. Exact digit rows help on numeric multi-key needles and do little or nothing on needles that are not numbers: five kinds of text on Llama-3.1-8B, names and phrases on Qwen3.5-9B. The fill fails on those too; at split 12, Llama-3.1-8B's fill retrieved 0 of 44 hexadecimal identifiers, against 44 of 44 for exact. Exact rows chosen by attention, the 5~\% of prefix rows that receive the most attention from the prompt's last tokens, help on both kinds.

\begin{table}[H]
\centering
\caption{Change in retrieval from the filled region over the fill alone, in points [95~\% CI], on the reference models.}\label{tab:I.2}
\small
\setlength{\tabcolsep}{4pt}
\begin{tabular}{lllll}
\toprule
Model & Needles & Units & Exact digit rows & \begin{tabular}[b]{@{}l@{}}Exact rows by \\ attention\end{tabular} \\
\midrule
Llama-3.1-8B & numeric, multi-key & 156 & +25 [+18, +32] & +22 [+14, +29] \\
Llama-3.1-8B & text & 660 & +2 [0, +3] & +15 [+12, +18] \\
Qwen3.5-9B & numeric, multi-key & 156 & +26 [+19, +33] & +21 [+13, +28] \\
Qwen3.5-9B & text & 264 & 0 [$-$1, +1] & +16 [+11, +20] \\
\bottomrule
\end{tabular}
\end{table}

\subsection{Fill error by token class}\label{sec:I.3}
On Llama-3.1-8B and Qwen3.5-9B at split 12, over 23 development prompts, we measured each token's relative error in what the late layers read: the attention keys and values and, on Qwen3.5-9B, the Gated DeltaNet inputs. Across nine token classes (word starts, word pieces, numbers, punctuation, spaces and others; the 23 special tokens left out), the largest class mean is 1.08 and 1.13 times the smallest. What the error costs is concentrated. We estimated each row's cost to first order, as the change in loss if its late attention keys and values were exact, from the gradient at the filled cache. Each document's top 5~\% of tokens hold 80~\% and 63~\% of that estimate's total. Its total overstates the measured gain from making every late attention key and value exact by 2.6 and 2.4 times, so we use it to rank rows only. The token-class shares of \S\ref{sec:2.4} use the same estimate on a separate development corpus of 300 documents, cut into 801 windows on each model. There we summed the estimate's absolute value by token class, leaving out each window's first token, and divided each class's share of that sum by its share of the tokens.

\subsection{Checking a filled cache during decoding}\label{sec:I.4}
On Qwen3.8-27B in a development build of the engine, a decode-time check bounds the attention each filled row receives from the current query and flags rows that cross a threshold for exact recomputation. Scoring every filled row at every step is sound and left the generated tokens unchanged; in two runs it added 7.9 to 9.7~\% to decoding time at 16K and 16.9 to 18.2~\% at 32K. Checking each row only every 4, 8 or 16 steps, with a watch list, costs less, but in 14 configurations at least one row that needed repair went unrepaired in 15 to 98~\% of decoding steps.

\subsection{Needle suites and RULER}\label{sec:I.5}
On Qwen3.8-27B in a development build of the engine, an earlier configuration (40~\% exact rows chosen by the probe, special rows exact and the value rescale) passed both of our needle suites, 96 of 96 needles, as stock did. On RULER at 16K it retrieved 8 of 20 on niah\_multikey\_2 and 6 of 20 on niah\_multikey\_3, against 20 of 20 for stock on each. After this run we stopped using our needle suites for any decision.

\end{document}